# Where "Ignoring Delete Lists" Works:
# Local Search Topology in Planning Benchmarks

**Jörg Hoffmann**                                                    HOFFMANN@MPI-SB.MPG.DE

*Max Planck Institute for Computer Science,*
*Stuhlsatzenhausweg 85,*
*66123 Saarbrücken*
*Germany*

## Abstract

Between 1998 and 2004, the planning community has seen vast progress in terms of the sizes of benchmark examples that domain-independent planners can tackle successfully. The key technique behind this progress is the use of heuristic functions based on relaxing the planning task at hand, where the relaxation is to assume that all delete lists are empty. The unprecedented success of such methods, in many commonly used benchmark examples, calls for an understanding of what classes of domains these methods are well suited for.

In the investigation at hand, we derive a formal background to such an understanding. We perform a case study covering a range of 30 commonly used STRIPS and ADL benchmark domains, including all examples used in the first four international planning competitions. We *prove* connections between domain structure and local search topology – heuristic cost surface properties – under an idealized version of the heuristic functions used in modern planners. The idealized heuristic function is called $h^+$, and differs from the practically used functions in that it returns the length of an *optimal* relaxed plan, which is **NP**-hard to compute. We identify several key characteristics of the topology under $h^+$, concerning the existence/non-existence of unrecognized dead ends, as well as the existence/non-existence of constant upper bounds on the difficulty of escaping local minima and benches. These distinctions divide the (set of all) planning domains into a taxonomy of classes of varying $h^+$ topology. As it turns out, many of the 30 investigated domains lie in classes with a relatively easy topology. Most particularly, 12 of the domains lie in classes where FF's search algorithm, provided with $h^+$, is a polynomial solving mechanism.

We also present results relating $h^+$ to its approximation as implemented in FF. The behavior regarding dead ends is provably the same. We summarize the results of an empirical investigation showing that, in many domains, the topological qualities of $h^+$ are largely inherited by the approximation. The overall investigation gives a rare example of a successful analysis of the connections between typical-case problem structure, and search performance. The theoretical investigation also gives hints on how the topological phenomena might be automatically recognizable by domain analysis techniques. We outline some preliminary steps we made into that direction.

## 1. Introduction

Between 1998 and 2004, one of the strongest trends in the planning community has been that towards heuristic planners, more specifically towards the use of heuristic distance (in most cases, goal distance) estimation functions. The best runtime results, progressing far beyond the sizes of benchmark examples that previous domain-independent planners could tackle successfully, have been achieved based upon a technique phrased "ignoring delete lists".

There, the heuristic function is derived by considering a relaxation of the planning task at hand, where the relaxation is to assume that all delete lists (i.e. the negative effects of the available planning operators) are empty. During search, may it be forward or backward, state space or plan space, the heuristic value of a search state in this framework is (an estimate of) the difficulty of extending the state to a solution using the relaxed operators, where "difficulty" is defined as *the number of (relaxed) actions needed*.

The number of real actions needed to extend a search state to a solution is at least as high as the number of relaxed actions needed. So optimal (shortest) relaxed solutions can, in principle, be used to derive admissible heuristic functions. However, as was first proved by Bylander (1994), deciding bounded plan existence, i.e., the existence of a plan with at most some given number of actions, is **NP**-hard even when there are no delete lists.[1] Thus there is not much hope to find optimal relaxed plans (i.e., optimal relaxed solution-extensions of search states) fast. Instead, one can *approximate* the length of an optimal relaxed plan to a search state. Techniques of this kind were first, independently, proposed by McDermott (1996) and by Bonet, Loerincs, and Geffner (1997), who developed the planners Unpop (McDermott, 1996, 1999), and HSP1 (Bonet et al., 1997). Both these planners perform forward state space search guided by an approximation of relaxed goal distance. Unpop approximates that distance by backchaining from the goals, HSP1 approximates that distance by a forward value iteration technique.

In the 1st international planning competition, *IPC-1* (McDermott, 2000), hosted at AIPS-1998, HSP1 compared well with the four other competitors. This inspired the development of HSP-r and HSP2 (Bonet & Geffner, 1999, 2001b, 2001a), GRT (Refanidis & Vlahavas, 1999, 2001), AltAlt (Nguyen & Kambhampati, 2000; Srivastava, Nguyen, Kambhampati, Do, Nambiar, Nie, Nigenda, & Zimmermann, 2001), as well as FF (Hoffmann, 2000; Hoffmann & Nebel, 2001a; Hoffmann, 2001a). HSP-r avoids heuristic re-computations by changing the search direction. HSP2 implements the various HSP versions in a configurable hybrid system. GRT avoids heuristic re-computations by changing the heuristic direction (the direction in which relaxed plans are computed). AltAlt uses a planning graph to extract heuristic values. FF uses a modified technique for approximating optimal relaxed plan length (namely, by computing a not necessarily optimal relaxed plan, which can be done in polynomial time), as well as new pruning and search techniques. FF inspired the integration of heuristic search engines into Mips (Edelkamp & Helmert, 2001) and STAN4 (Fox & Long, 2001), using elaborated variations of FF's relaxed plan length estimation technique.

In the 2nd international planning competition, *IPC-2* (Bacchus, 2001), hosted at AIPS-2000, the heuristic planners dramatically outperformed the other approaches runtime-wise, scaling up to benchmark examples far beyond reach of previous, e.g., Graphplan-based (Blum & Furst, 1995, 1997), systems. This caused the trend towards heuristic planners to still increase. Various researchers extended relaxed plan distance estimation techniques to temporal and numeric settings (Do & Kambhampati, 2001; Hoffmann, 2002, 2003a; Edelkamp, 2003b). Others adapted them for use in partial order plan-space search (Nguyen

---

1. For *parallel* planning, where the bound is on the number of parallel time steps needed, deciding bounded plan existence is easy without delete lists. However, heuristic functions based on this observation have generally not been found to provide useful search guidance in practice (see, for example, Haslum & Geffner, 2000; Bonet & Geffner, 2001b).

& Kambhampati, 2001; Younes & Simmons, 2002), developed variations of them to provide new means of heuristic guidance (Onaindia, Sapena, Sebastia, & Marzal, 2001; Sebastia, Onaindia, & Marzal, 2001), or modified them to take exclusion relations in a planning graph into account (Gerevini & Serina, 2002; Gerevini, Serina, Saetti, & Spinoni, 2003).

In the 3rd international planning competition, *IPC-3* (Long & Fox, 2003), hosted at AIPS-2002, out of 11 domain-independent competing systems, 7 were using relaxed plan distance estimations in one or the other form. The 1st prize winner LPG (Gerevini & Serina, 2002; Gerevini, Saetti, & Serina, 2003) uses, amongst other heuristics, a relaxed planning technique to estimate the difficulty of sub-goal achievement in a planning graph. In the 4th international planning competition, *IPC-4* (Hoffmann & Edelkamp, 2005; Edelkamp, Hoffmann, Englert, Liporace, Thiebaux, & Trüg, 2005), hosted at ICAPS-2004, out of 13 competing sub-optimal systems, 12 were using relaxed plan based heuristics. There were two 1st prize winners in that category: Fast-Downward (Helmert, 2004; Helmert & Richter, 2004) and SGPlan (Chen & Wah, 2003; Chen, Hsu, & Wah, 2004). The latter uses the numeric version of FF as a sub-process. One version of the former combines FF's heuristic estimates with a new heuristic function based on causal graph analysis (Helmert, 2004).

In the investigation at hand, we derive a formal background as to what classes of domains methods of the kind described above are well suited for. We make two simplifying assumptions. First, we consider forward state space search only, as used by, for example, Unpop, HSP, Mips, FF, and Fast-Downward. In a forward state space search, one starts at the initial state and explores the space of reachable states until a goal state is found. The state transitions follow a *sequential* planning framework, where only a single action is applied at a time.[2] Assuming forward search makes the investigation easier since such a search is a very natural and simple framework. Our second simplifying assumption is to idealize matters in that we consider the heuristic value given by the *optimal* relaxed plan length (the length of a shortest sequential relaxed plan) to each search state $s$; we denote that value with $h^+(s)$. Under this assumption, as we will see there are many *provable* connections between domain structure and heuristic quality. Of course, the simplifying assumptions restrict the relevance of the results for practical planners. More on this is said below in this section, and in Section 7. Another, more benign, restriction we make is to consider solvable tasks only. This is a very common restriction in AI Planning, particularly in the competitions, where the main focus is on how good planners are at *finding plans*. More specifically, the main focus of the investigation at hand is to characterize the kinds of domains in which (relaxed-plan based) heuristic planners can find plans fast.

It is common knowledge that the behavior of heuristic search methods (may they be global or local, i.e., with or without backtracking mechanisms) depends crucially on the quality of the underlying heuristic function. This has, for example, been studied in the SAT community, for example by Frank, Cheeseman, and Stutz (1997). In their work, these authors empirically investigate properties of the *local search topology*, i.e., of topological properties like the sizes of local minima etc., in SAT instances under a standard heuristic function. We adapt Frank et al.'s definitions to AI planning. In *difference* to Frank et al., we take a more analytical approach where we *prove* properties that are valid across certain ranges, namely domains, of example problem instances. We investigate a range of 30 commonly used STRIPS and ADL benchmark domains including all examples used in the first four international planning competitions. We identify several key characteristics of the topology of the respective search spaces under $h^+$. The characteristics are the following.

---

2. In principle, a parallel forward search is possible, too. To the best of the author's knowledge, there is no published work about an implementation of this, at the time of writing. The main difficulty is, presumably, the high branching factor.

1. In 24 of the benchmark domains, there are no unrecognized dead ends, i.e., no states from which the goal is unreachable but for which there is a relaxed plan.

2. In 17 of the above 24 benchmark domains, the "maximal exit distance from local minima" is constantly bounded, i.e., one can always escape local minima (regions where all neighbors have a higher heuristic value) within a number of steps that is constant across all instances of the domain, regardless of their size (in fact, in 13 of these domains there are no local minima at all).

3. In 12 of the above 17 benchmark domains, the "maximal exit distance from benches" is constantly bounded, i.e., one can always escape benches (regions where all states have the same heuristic value) within a number of steps that is constant across all instances of the domain, regardless of their size (in 6 domains the bound is 1, in one domain it is even 0).

Beside the "positive" results proving characteristic qualities of the $h^+$ function, the investigation also provides (parameterized) counter-examples in the negative cases. The results divide the investigated domains (more generally, all possible planning domains) into a meaningful taxonomy of classes which differ in terms of their topological behavior with respect to $h^+$. Many of the 30 investigated domains lie in relatively easy classes, i.e., classes where $h^+$ is a – provably – high-quality heuristic. Most particularly, the 12 domains with all the above properties lie in classes where FF's search algorithm is a polynomial solving mechanism, under the idealizing assumption that FF's approximative heuristic function identifies the real $h^+$ distances. FF's search algorithm, called *enforced hill-climbing*, tries to escape local minima or benches by means of a breadth-first search. Breadth-first search is exponential only in the search depth. So if local minima and benches can always be escaped from within a constant number of steps – as is the case in these 12 domains – then the effort spent in the search is polynomially bounded. In this way, our results provide non-trivial insights into typical-case problem structure (in benchmarks), and its possible effects on search performance. Examples of successful theoretical investigations of this kind are extremely rare in the AI literature.

To give the reader a feeling for what we are looking at, Figure 1 shows two visualized state spaces. The shown tasks are instances of two domains from the easiest classes of the taxonomy, Gripper and Logistics. The graph nodes are the states, the edges are the state transitions (action applications), the height is given by the $h^+$ value.[3] In both pictures, the initial state is somewhere in the left top part. The goal states are, of course, the states with minimal – zero – $h^+$ value. The Gripper picture speaks for itself. The Logistics topology is less extreme, but still the state space forms one big valley at the bottom of which there are the goal states.

---

3. The $h^+$ values here, and in an empirical investigation (Hoffmann, 2001b, 2003b) preceding our theoretical analysis, were computed by an iterative deepening forward search in the space of relaxed action sequences.

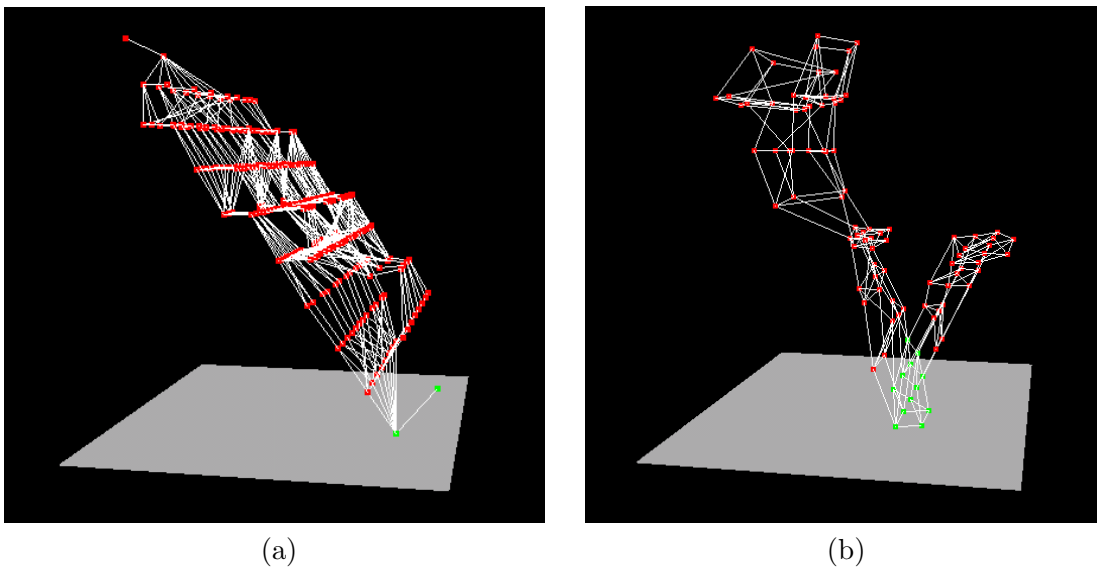

(a)                                  (b)

Figure 1: Visualized state space under $h^+$ of (a) a Gripper and (b) a Logistics instance.

Of course, FF's approximation of $h^+$, which we refer to as $h^{FF}$, does *not* always identify the real $h^+$ values, and so it is a priori not evident what relevance the theoretical results about $h^+$ have for FF's efficiency in practice. Additionally, most forward searching planners do not use enforced hill-climbing, for which the topological results have the most striking impact. Finally, and most importantly, several competitive other planners do not even perform a forward search, or use additional/new techniques in the heuristic function that are explicitly aimed at identifying better information than relaxed plans. Prominent systems of the former kind are HSP-r and LPG, prominent systems of the latter kind are LPG and Fast-Downward.

As for the relevance of the results for the performance of FF, the practical performance of FF coincides quite well with them. More concretely, the behavior of $h^+$ with respect to dead ends is *provably* the same as that of $h^{FF}$. Moreover, a large-scale empirical investigation (contained in Hoffmann, 2003b) has shown that, in many domains, the topology of $h^+$ is largely preserved by $h^{FF}$. We include a section containing a brief summary of these results. The relevance of the topological results for forward search algorithms other than enforced hill-climbing, and the performance of planners using other search paradigms or enhanced heuristics, is discussed in Section 7.

We remark that our topological investigation was not specifically intended to identify properties relevant to enforced hill-climbing. The theoretical investigation was preceded by an empirical investigation (Hoffmann, 2001b, 2003b) where we measured all kinds of topological parameters, including, for example, size and diameter of local minima, benches, and other structures such as so-called "valley" regions. It turned out that the only topology parameters that showed interesting behavior across a significant number of domains were the maximal exit distance parameters considered in the investigation at hand. This, in fact, came as a surprise to us – we invented enforced hill-climbing in FF *before* it became clear that many of the planning benchmarks share topological properties favoring precisely this particular search algorithm.

Observe that the proved results are of a worst-case nature, i.e., a heuristic search using $h^+$ can show good behavior in an example suite of a domain even if that domain lies in a very difficult class of the taxonomy – given the particular example instances in the test suite do not emphasize on the worst cases possible in the domain. Where relevant, we will discuss this issue with regards to the example suites used in the competitions.

The employed proof methods give hints as to how the topological phenomena might be automatically detectable using general domain analysis techniques. In an extra section, we report on a first (not yet very successful) attempt we made to do that.

The proofs for the individual planning domains are, in most cases, not overly difficult, but the full details for all domains are extremely space consuming. The details (except for the 5 IPC-4 domains), i.e., PDDL-like definitions of the domains as well as fully detailed proofs, can be looked up in a long (138 pages) technical report (Hoffmann, 2003c) that also forms an online appendix to the article.[4] The article itself provides proof sketches, which are much better suited to get an overall understanding of the investigation and its results. Since even the proof sketches are sometimes hard to read, they are moved into an appendix; another appendix provides brief descriptions of all domains. The main body of text only gives the results and an outline of the main proof arguments used to obtain them.

The paper is organized as follows. Section 2 provides the necessary background, i.e. a straightforward formal framework for STRIPS and ADL domains, an overview of the investigated domains, and the definitions of local search topology. Section 3 presents some core lemmas underlying many of the proofs in the single domains, and illustrates the lemmas' application in a small example. Section 4 gives all the results with a brief proof outline, and shows the resulting planning domain taxonomy. Section 5 presents the results relating $h^+$ to $h^{FF}$, and Section 6 reports on our first attempt to design domain analysis techniques for automatically detecting the $h^+$ topological phenomena. Section 7 concludes the article with a brief discussion of our contributions and of future work. Appendix A contains the proof sketches for the individual domains, Appendix B contains the domain descriptions.

## 2. Background

Background is necessary on the planning framework, the investigated domains, and local search topology.

### 2.1 Planning Framework

To enable theoretical proofs to properties of planning *domains* rather than single tasks, we have defined a formal framework for STRIPS and ADL domains, formalizing in a straightforward manner the way domains are usually dealt with in the community. We only outline the rather lengthy definitions, and refer the reader to the TR (Hoffmann, 2003c) for details. In what follows, by sets we mean finite sets unless explicitly said otherwise.

---

4. We remark that the TR is *not* a longer version of the paper at hand. The TR's overall structure and presentation angle are very different, and it is only intended as a source of details if needed.

A planning domain is defined in terms of a set of *predicates*, a set of *operators*, and a possibly infinite set of *instances*. All logical constructs in the domain are based on the set of predicates. A *fact* is a predicate applied to a tuple of objects. The operators are ($k$-ary, where $k$ is the number of operator parameters) functions from the (infinite) set of all objects into the (infinite) set of all STRIPS or ADL *actions*. A STRIPS action $a$ is a triple $(pre(a), add(a), del(a))$: $a$'s *precondition*, which is a conjunction of facts; $a$'s *add list*, a fact set; and $a$'s *delete list*, also a fact set. An ADL action $a$ is a pair $(pre(a), E(a))$ where the precondition $pre(a)$ is a first order logical formula without free variables, and $E(a)$ is a set of effects $e$ of the form $(con(e), add(e), del(e))$ where $con(e)$, the *effect condition*, is a formula without free variables, and $add(e)$ (the effect's add list) as well as $del(e)$ (the effect's delete list) are fact sets. If the add list of an action/effect contains a fact $p$, we also say that the action/effect *achieves p*.

An instance of a domain is defined in terms of a set of objects, an *initial state*, and a *goal condition*. The initial state is a set of facts, and the goal condition is a formula without free variables (in the STRIPS case, a conjunction of facts). The facts that are contained in the initial state are assumed to be true, and all facts *not* contained in it are assumed to be false, i.e., as usual we apply the closed-world assumption. An instance of a domain constitutes, together with the domain's operators, a *planning task* $(A, I, G)$ where the action set $A$ is the result of applying the operators to the instance's objects (i.e., to all object tuples of the appropriate lengths), and the initial state $I$ and goal condition $G$ are those of the instance. We identify instances with the respective planning tasks.

A *state s* is a set of facts. A logical formula *holds* in a state if the state is a model of the formula according to the standard definition for first order logic (where a logical atom, a fact, holds iff it is contained in the state). The *result* $Result(s, \langle a \rangle)$ of applying an action sequence consisting of a single STRIPS or ADL action $a$ to a state $s$ is defined as follows. If the action's precondition does not hold in $s$, then $Result(s, \langle a \rangle)$ is undefined. Otherwise, $Result(s, \langle a \rangle)$ is obtained from $s$ by including all of $a$'s add effects, and (thereafter) removing all of $a$'s delete effects – if $a$ is an ADL action, only those add effects $add(e)$ are included (delete effects $del(e)$ are removed) for which the respective effect condition $con(e)$ holds in $s$. The result of applying a sequence $\langle a_1, \ldots, a_n \rangle$ consisting of more than one action to a state $s$ is defined as the iterative application of the single actions in the obvious manner: apply $a_1$ to $s$, then apply $a_2$ to $Result(s, \langle a_1 \rangle)$, and so on.

A *plan*, or *solution*, for a task $(A, I, G)$ is a sequence of actions $P \in A^*$ that, when successively applied to $I$, yields a *goal state*, i.e., a state in which $G$ holds. (We use the standard notation $M^*$, where $M$ is a set, to denote the set of all sequences of elements of $M$.) For many proofs we need the notion of optimality. A plan $P$ for a task $(A, I, G)$ is *optimal* if there is no plan for $(A, I, G)$ that contains fewer actions than $P$.

Note that, as announced in the introduction, the definition, in particular the definition of plan optimality, stays within the forward state space search framework where *plans are simple sequences of actions*. Note also that ignoring the delete lists simplifies a task only if all formulas are negation free. For a fixed domain, tasks can be polynomially normalized to have that property: compute the negation normal form to all formulas (negations only in front of facts), then introduce for each negated fact $\neg B$ a new fact not-$B$ and make sure it is true in a state iff $B$ is false (Gazen & Knoblock, 1997). This is the pre-process done in, for example, FF. In the investigation at hand, we have considered the normalized versions of the domains.[5]

We also consider a few domains, from the IPC-4 collection, that feature *derived predicates*. Such predicates are not affected by the effects of the operators, and their truth value is instead derived from the values of the other, *basic*, predicates, via a set of derivation rules. A derivation rule has the form $\phi(\overline{x}) \Rightarrow P(\overline{x})$ where $P$ is the derived predicate and $\phi$ (a formula) is the rule's *antecedent*, both using the free variables $\overline{x}$. The obvious idea is that, if $\phi(\overline{x})$ holds, then $P(\overline{x})$ can be concluded. In a little more detail, the semantics are defined as follows. In the initial state, and whenever an action was applied, first all derived predicate instances (*derived facts*) are assumed to be false, then all derivation rules are applied until a fixpoint occurs. The derived facts that could not be concluded until then are said to be false (this is called *negation as failure*). Derived predicates can be used just like any other predicate in the operator preconditions, in the effect conditions, and in the goal condition. However, to ensure that there is a unique fixpoint of rule application, the use of derived predicates in derivation rule antecedents is restricted (in the context of IPC-4) to a positive use in the sense that these predicates do not appear negated in the negation normal form of any rule antecedent (Hoffmann & Edelkamp, 2005).

To make ignoring delete lists a simplification, one also needs that the derived facts are used only positively in the operator preconditions, effect conditions, and goal condition (otherwise the derived predicates can, for example, be used to model negated preconditions etc.). Due to the negation as failure semantics of derived predicates, there isn't a simple compilation of negations as in the pure ADL case. The approach we take here, and that is implemented in, for example, the version of FF that treats derived predicates (Thiebaux, Hoffmann, & Nebel, 2003, 2005), is to simply ignore (replace with true) negated derived predicates in (the negation normal form of) operators and the goal (see also below, Section 2.3).

## 2.2 Domains Overview

As said before, our case study covers a total of 30 commonly used STRIPS and ADL benchmark domains. These include all the examples from the first four international competitions, plus 7 more domains used in the literature. Brief descriptions of all domains can be looked up in Appendix B, full formal definitions of the domains (except the 5 IPC-4 domains) are in the TR (Hoffmann, 2003c). Note that, for defining a domain, one must amongst other things decide what exactly the instances are. Naturally, to do so we have abstracted from the known example suites. In most cases the abstraction is obvious, in the less obvious cases the respective subsection of Appendix B includes some explanatory remarks.

Here, we provide a brief overview of the 30 analyzed domains. The domains are categorized into three groups according to their semantics, at a high level of abstraction. The categorization is *not*, in any way, related to the topological characterization we will derive later. We use it only to give the overview some structure.

---

5. Ignoring delete lists in the normalized domains comes down to a relaxation that, basically, allows (the translated) facts to take on both truth values at the same time.

1. **Transportation domains**. These are domains where there are locations, objects that must be transported, and vehicles that are the means of transportation.[6] Operators mostly either move a vehicle, or load (unload) an object onto (from) a vehicle. The domains differ in terms of various constraints. An important one is that, in many domains, vehicles can move instantaneously between any two locations, while in other domains the movable links between locations form arbitrary road maps. There are 13 transportation domains in the collection we look at. *Logistics* – the classical transportation domain, where trucks/airplanes transport objects within/between cities. *Gripper* – a robot with two gripper hands transports a number of balls (one at a time in each hand) from one room to another. *Ferry* – a single ferry transports cars one at a time. *Driverlog* – trucks need drivers on board in order to move, the location links form bi-directional road maps (which can be different for trucks and drivers). *Briefcaseworld* – a briefcase moves, by conditional effects, all objects along that are inside it. *Grid* – a robot transports keys on a grid-like road map where positions can be locked and must be opened with keys of matching shapes. *Miconic-STRIPS* – an elevator transports passengers, using explicit actions to board/deboard passengers. *Miconic-SIMPLE* – like *Miconic-STRIPS*, but passengers board/deboard "themselves" by conditional effects of the action that stops the elevator at a floor. *Miconic-ADL* – like *Miconic-SIMPLE*, but various constraints must be obeyed (for example, VIPs first). *Zenotravel* – airplanes use fuel items that can be replenished one by one using a refuel operator. *Mprime* – on an arbitrary road map, trucks use non-replenishable fuel items, and fuel can be transferred between locations. *Mystery* – like *Mprime*, but without the possibility to transfer fuel. *Airport* – inbound and outbound planes must be moved safely across the road map of an airport.

2. **Construction domains**. These are generally not as closely related as the transportation domains above. What the construction domains have in common, roughly, is that a complex object must be built out of its individual parts. There are 6 such domains in the collection we look at. *Blocksworld-arm* – the classical construction domain, where blocks are picked up/put down or stacked onto/unstacked from each other by means of a robot arm. *Blocksworld-no-arm* – like above, but blocks are moved around directly from a block to a block / from the table to a block / from a block to the table. *Depots* – a combination of *Blocksworld-arm* and *Logistics*, where objects must be transported between locations before they can be stacked onto each other. *Freecell* – an encoding of the solitaire card game that comes with Microsoft Windows (the "complex object" to be constructed is the final position of the cards). *Hanoi* – an encoding of the classical Towers of Hanoi problem. *Assembly* – a complex object must be assembled together out of its parts, which themselves might need to be assembled beforehand.

3. **Other domains**. There are 11 domains in the collection whose semantics do not quite fit into either of the above groups. *Simple-Tsp* – a trivial STRIPS version of the TSP problem, where the move operator can be applied between any two locations. *Movie* – in order to watch a movie, one must buy snacks, set the counter on the video to zero, and rewind the tape. *Tireworld* – a number of flat tires must be replaced, which involves various working steps (like removing a flat tire and putting on a new one). *Fridge* – for a number of fridges, the broken compressors must be replaced, which involves various working steps (like loosening/fastening the screws that hold a compressor). *Schedule* – objects must be processed (painted, for example) on a number of machines. *Satellite* – satellites must take images (of phenomena in space), using appropriate instruments. *Rovers* – rovers must navigate along a road map, take soil/rock samples as well as images, and communicate the resulting data to a lander. *Pipesworld* – oil derivatives must be propagated through a pipeline network. *PSR* – some lines must be re-supplied in a faulty electricity network. *Dining-Philosophers* – the deadlock situation in the Dining-Philosophers problem, translated to ADL from the automata-based "Promela" language (Edelkamp, 2003a), must be found. *Optical-Telegraph* – similar to *Dining-Philosophers*, but considering an encoding of a telegraph communication system.

---

6. The term "transportation domains" was suggested, for example, by Long and Fox (2000) and Helmert (2003). The transportation benchmarks are generally more closely related than the other groups of domains overviewed below, and we will sometimes discuss transportation domains on a rather generic level.

## 2.3 Local Search Topology

Remember that we only consider solvable tasks, since the main focus of the investigation is to characterize the kinds of domains in which heuristic planners can *find plans* fast. Some discussion of unsolvable tasks is in Section 7.

Given a planning task $(A, I, G)$. The *state space* $(S, T)$ is a graph where $S$ are all states that are reachable from the initial state, and $T$ is the set of state transitions, i.e., the set of all pairs $(s, s') \in S \times S$ of states where there is an action that leads to $s'$ when executed in $s$. The *goal distance* $gd(s)$ for a state $s \in S$ is the length of a shortest path in $(S, T)$ from $s$ to a goal state, or $gd(s) = \infty$ if there is no such path. In the latter case, $s$ is a *dead end*; we discuss such states directly below. A *heuristic* is a function $h : S \mapsto \mathbf{N} \cup \{\infty\}$.[7] A heuristic can return $\infty$ to indicate that the state at hand might be a dead end.

Given a STRIPS action $a = (pre(a), add(a), del(a))$, the *relaxation* $a^+$ of $a$ is $(pre(a), add(a), \emptyset)$. Given an ADL action $a = (pre(a), E(a))$, the relaxation $a^+$ of $a$ is $(pre(a), E(a)^+)$ where $E(a)^+$ is the same as $E(a)$ except that all delete lists are empty. For a set $A$ of actions, the *relaxation* $A^+$ of $A$ is $A^+ := \{a^+ \mid a \in A\}$. An action sequence $\langle a_1, \ldots, a_n \rangle$ is a *relaxed plan for* $(A, I, G)$ if $\langle a_1^+, \ldots, a_n^+ \rangle$ is a plan for $(A^+, I, G)$. With that, for any state $s$, $h^+(s) = min\{n \mid P = \langle a_1, \ldots, a_n \rangle \in A^*, P$ is relaxed plan for $(A, s, G)\}$, where the minimum over an empty set is $\infty$.

In the presence of derived predicates, as said above we additionally relax the planning task by ignoring (replacing with true) all negated derived predicates in the negation normal forms of preconditions, effect conditions, and the goal condition. Note that, with this additional simplification, it can happen that $h^+(s) = 0$ although $s$ is not a goal state, because the simplification might relax the goal condition itself. Indeed, this happens in the PSR domain. In all other domains we consider here, derived predicates are either not used at all or used only positively, so there $h^+(s) = 0$ iff $s$ is a goal state.

---

7. While the article focuses mainly on the $h^+$ heuristic, we keep the topology definitions – which do not depend on the specific heuristic used – somewhat more general.

One phenomenon that is clearly relevant for the performance of heuristic state space search is that of dead end states $s$, $gd(s) = \infty$. A heuristic function $h$ can return $h(s) = \infty$. Taking this as an indication that $s$ is a dead end, the obvious idea is to remove $s$ from the search space (this is done in, for example, HSP and FF). This technique is only adequate if $h$ is *completeness preserving* in the sense that $h(s) = \infty \Rightarrow gd(s) = \infty$ for all $s \in S$. With a completeness-preserving heuristic, a dead end state $s$ is called *recognized* if $h(s) = \infty$ and *unrecognized* otherwise. Note that $h^+$ is completeness preserving. If a task can not be solved even when ignoring the delete lists, then the task is unsolvable. From now on we assume that the heuristic we look at is completeness preserving. With respect to dead ends, any planning state space falls into one of the following four classes. The state space is called:

1. *Undirected*, if, for all $(s, s') \in T$, $(s', s) \in T$.

2. *Harmless*, if there exists $(s, s') \in T$ such that $(s', s) \notin T$, and, for all $s \in S$, $gd(s) < \infty$.

3. *Recognized*, if there exists $s \in S$ such that $gd(s) = \infty$, and, for all $s \in S$, if $gd(s) = \infty$ then $h(s) = \infty$.

4. *Unrecognized*, if there exists $s \in S$ such that $gd(s) = \infty$ and $h(s) < \infty$.

In the first class, there can be no dead ends because everything can be undone. In the second class, some things can not be undone, but those single-directed state transitions do not "do any harm", in the sense that there are no dead end states. In the third class, there are dead end states, but all of them are recognized by the heuristic function. The only critical case for heuristic search is class four, where a search algorithm can run into a dead end without noticing it. This is particularly relevant if, potentially, large regions of the state space consist of unrecognized dead end states. To capture this, we define the *depth* of an unrecognized dead end $s$ as the number of states $s'$ such that $s'$ is an unrecognized dead end, and $s'$ is reachable from $s$ by a path that moves only through unrecognized dead ends.

Our investigation determines, for each of the 30 benchmark domains looked at, exactly in which of the above four dead end classes the instances of the domain belong. For the domains where it turns out that there can be unrecognized dead ends, we construct parameterized examples showing that the unrecognized dead ends can be arbitrarily deep. In several domains, individual instances can fall into different classes. In this case we associate the overall domain with the worst-case class, i.e., the class with highest index in the above. For example, in Miconic-ADL, if there are no additional constraints to be obeyed on the transportation of passengers then the state space is harmless as in Miconic-SIMPLE. But if constraints on, for example, the possible direction of travel and the access to floors are given, then unrecognized dead ends can arise. To avoid clumsy language, henceforth, if we say that a state space is harmless/recognized/unrecognized, then we mean that it falls into the respective class, or into a class below it.

We now get into the definitions of general topological phenomena, i.e., of relevant properties of the search space surface. We adapt the definitions given for SAT by Frank et al. (1997). The difference between the SAT framework there, and the planning formalism here, lies in the possibly single-directed state transitions in planning. In the search spaces considered by Frank et al., all state transitions can be traversed in both directions. Single-directed state transitions can have an important impact on the search space topology, enabling, for example, the existence of dead ends.[8]

The base entity in the state space topology are what Frank et al. name *plateaus*. These are regions that are equivalent under reachability aspects, and look the same from the point of view of the heuristic function. For $l \in \mathbf{N} \cup \{\infty\}$, a *plateau P of level l* is a maximal subset of $S$ for which the induced subgraph in $(S, T)$ is strongly connected, and $h(s) = l$ for each $s \in P$.[9] Plateaus differ in terms of the possibilities of leaving their heuristic level, i.e., of reaching an *exit*. For a plateau $P$ of level $l$, an exit is a state $s$ reachable from $P$, such that $h(s) = l$ and there exists a state $s'$, $(s, s') \in T$, with $h(s') < h(s)$. Based on the behavior with respect to exits, we distinguish between five classes of plateaus. We need the notion of *flat paths*. These are paths in $(S, T)$ on that the value of $h$ remains constant.

1. A *recognized dead end* is a plateau $P$ of level $l = \infty$.

2. A *local minimum* is a plateau $P$ of level $0 < l < \infty$ from that no exit is reachable on a flat path.

3. A *bench* is a plateau $P$ of level $0 < l < \infty$, such that at least one exit is reachable from $P$ on a flat path, and at least one state on $P$ is not an exit.

4. A *contour* is a plateau $P$ of level $0 < l < \infty$ that consists entirely of exits.

5. A *global minimum* is a plateau $P$ of level 0.

Each plateau belongs to exactly one of these classes. Intuitively, the roles that the different kinds of plateaus play for heuristic search are the following. Recognized dead ends can be ignored with a completeness-preserving heuristic function. Local minima are difficult because all neighbors look worse, so it is not clear in which direction to move next. Benches are potentially easier, because one can step off then without temporarily worsening the heuristic value. From contours, one can step off immediately.[10]

The main difficulty for a heuristic search is how to deal with the local minima and the benches. In both cases, the search algorithm must (eventually) find a path to an exit in order to get closer to the goal (as far as the heuristic function is informed about what is closer to the goal and what is not). How difficult it is to find an exit can be assessed by a variety of different parameters. The size (number of states) or diameter (maximum distance between any two states) of the local minimum/the bench, and the number of nearby exit states, to name some important ones. In the benchmarks considered, as mentioned in the introduction, we empirically found that there are no (or very few) interesting observations to be made about these parameters (Hoffmann, 2001b, 2003b).

---

8. One can, of course, introduce backtracking mechanisms into a search space, such as always giving the planner the possibility to retract the last step. But that does not affect the relevant topological differences between search spaces – instead of domains with/without dead ends, one gets domains where backtracking is necessary/not necessary.

9. The difference to the undirected case is that we require the states on the plateau to be *strongly* connected – with undirected state transitions this is trivially fulfilled by any set of connected states.

10. The differences to the undirected case lie in that there can be plateaus of level $\infty$, and that we allow exits to not lie on the plateaus themselves. The latter is just a minor technical device to obtain a compact terminology.

What one *can* frequently observe are interesting properties of the *distance* to the nearest exit state. The distance $dist(s, s')$ between any two states $s, s' \in S$ is the usual graph distance, i.e., the length of a shortest path from $s$ to $s'$ in $(S, T)$, or $\infty$ if there is no such path. The *exit distance* $ed(s)$ of a search state $s$ is the distance to the nearest exit, i.e.:

$$ed(s) = min\{d \mid d \text{ is the length of a path in } (S, T) \text{ from } s \text{ to a state } s' \text{ s.t. } h(s') = h(s),$$
$$\text{and there exists a state } s'' \text{ s.t. } (s', s'') \in T, \text{ and } h(s'') < h(s') \},$$

where, as before, the minimum over an empty set is $\infty$. Note that we do not require the path in the definition to be flat, i.e., it may be that, in order to reach $s'$, we temporarily have to increase the $h^+$ value. This is because we want the definition to capture possible "escape" routes from any state in the state space, including states that lie on local minima.

The *maximal local minimum exit distance*, $mlmed(S, T)$, of a state space $(S, T)$ is the maximum over the exit distances of all states on local minima, or 0 if there are no such states. The *maximal bench exit distance*, $mbed(S, T)$, of a state space $(S, T)$ is the maximum over the exit distances of all states on benches, or 0 if there are no such states. We will find that, in many of the considered domains, there are constant upper bounds on $mlmed(S, T)$ and/or $mbed(S, T)$ under $h^+$, i.e., bounds that are valid irrespectively of the (size of the) instance chosen.

The following is an implication that is relevant for the subsequent investigation.

**Proposition 1** *Given a solvable task $(A, I, G)$, with state space $(S, T)$ and a completeness-preserving heuristic $h$, where $h(s) = 0 \Rightarrow gd(s) = 0$ for $s \in S$. If there exists an unrecognized dead end $s \in S$, then $mlmed(S, T) = \infty$.*

**Proof:** Let $s$ be an unrecognized dead end, and let $s'$ be a state reachable from $s$ so that the $h$ value of $s'$ is minimal. Then $s'$ is an unrecognized dead end, too (in particular, $s$ is considered reachable from itself), and since $h(s') = 0 \Rightarrow gd(s') = 0$ we have $h(s') > 0$. We further have, since the $h$ value of $s'$ is minimal among the states reachable from $s$, that $h(s'') \geq h(s')$ for all states $s''$ reachable from $s'$. Thus the plateau on which $s'$ lies is a local minimum – no exits are reachable, in particular not on flat paths. This also shows that $s'$ has infinite exit distance. $\qquad\square$

Proposition 1 says that, in every region of unrecognized dead ends, there is a local minimum, given $h(s) = 0 \Rightarrow gd(s) = 0$.[11] With the above definitions, that unrecognized dead end state yields an infinite local minimum exit distance. It makes sense to define things this way because an (arbitrarily deep) unrecognized dead end is worse than any local minimum: it can not be escaped from at all.

---

11. Remember that the latter can be untrue for $h^+$ only if the domain features derived predicates that appear negated in the negation normal form of the goal condition. And even then, by the argument in the proposition, every region of unrecognized dead ends would contain a global minimum consisting of non-solution states. We could have defined such "fake"-global minima to be local minima, but decided against it in order to not overly complicate the topological definitions, and since that detail does not seem very important. As said before, in all but one of our 30 domains we have $h^+(s) = 0 \Leftrightarrow gd(s) = 0$ anyway.

## 3. Some Core Lemmas

In many of the investigated domains, intuitively similar patterns of problem structure cause the characteristic qualities of $h^+$. Some of this common structure can be generalized and captured in concise definitions and lemmas. The lemmas formulate sufficient criteria implying that (the state space of) a planning task has certain topological properties. Proofs for domains proceed, where possible, by applying the lemmas to arbitrary instances. In several domains where the lemmas can not be applied immediately (due to syntactic details of the domain definitions), similar proof arguments suffice to show the desired topological properties.

We restrict ourselves to STRIPS tasks in the lemmas. Appropriate extensions to ADL and/or to derived predicates are probably possible at least in certain cases, but we have not investigated this in detail – such extensions are likely to be rather complicated notationally, and the simpler STRIPS case suffices to transport the ideas.

| initial state: | | | |
|---|---|---|---|
| $at(V, L_1), at(O_1, L_1), at(O_2, L_2)$ | | | |
| goal: | | | |
| $at(O_1, L_2), at(O_2, L_1)$ | | | |
| actions: | | | |
| name | precondition | add list | delete list |
| $move(l, l')$ | $at(V, l)$ | $at(V, l')$ | $at(V, l)$ |
| $load(o, l)$ | $at(V, l), at(o, l)$ | $in(o, V)$ | $at(o, l)$ |
| $unload(o, l)$ | $at(V, l), in(o, V)$ | $at(o, l)$ | $in(o, V)$ |

Figure 2: A simple STRIPS transportation task.

Throughout the section, we assume we are given a STRIPS task $(A, I, G)$. As an illustrative example for the definitions and lemmas, we will use the simple transportation task defined in Figure 2. In what follows, there are three separate sections, concerned with dead ends, local minima, and benches, respectively.

The definitions and lemmas in the following are not syntactical, in the sense that they make use of informations that can not be computed efficiently (for example, inconsistencies between facts). We do not discuss this, and focus exclusively on the role of the definitions and lemmas as tools for proving $h^+$ topology. The role of the definitions and lemmas as tools for *automatically detecting* $h^+$ topology will be discussed in Section 6.

### 3.1 Dead Ends

We first focus on criteria sufficient for the non-existence of dead ends. Our starting point is a reformulated version of a simple result mentioned by, for example, Koehler and Hoffmann (2000). We need the notion of *inconsistency*. Two facts are inconsistent if there is no reachable state that contains both of them. A set of facts $F$ is inconsistent with another set of facts $F'$ if each fact in $F$ is inconsistent with at least one fact in $F'$.[12] An action $a \in A$ is *invertible* if:

(1) $add(a)$ is inconsistent with $pre(a)$;

(2) $del(a) \subseteq pre(a)$;

(3) there is an action $\overline{a} \in A$ such that

    (a) $pre(\overline{a}) \subseteq (pre(a) \cup add(a)) \setminus del(a)$,

    (b) $add(\overline{a}) = del(a)$, and

    (c) $del(\overline{a}) = add(a)$.

The intentions behind these requirements are the following. (1) and (2) ensure that $a$'s effects all occur, (3a) ensures that $\overline{a}$ is applicable, and (3b) and (3c) ensure that $\overline{a}$ undoes $a$'s effects. As an example, all actions in the illustrative task from Figure 2 are invertible. For example, an $a = move(l, l')$ action is inverted by $\overline{a} = move(l', l)$. To see that, simply insert the definitions: $add(a) = \{at(V, l')\}$ is inconsistent with $pre(a) = \{at(V, l)\}$; $del(a) = \{at(V, l)\} = pre(a)$; $pre(\overline{a}) = \{at(V, l')\} = add(a)$; $add(\overline{a}) = \{at(V, l)\} = del(a)$; $del(\overline{a}) = \{at(V, l')\} = add(a)$. Similarly easily, one sees that $load(o, l)$ and $unload(o, l)$ invert each other. Examples of benchmark domains with invertible actions are Blocksworld (in both variants), Logistics, and Gripper.

**Lemma 1** *[Koehler & Hoffmann, 2000] Given a STRIPS planning task $(A, I, G)$. If all actions $a \in A$ are invertible, then the state space of the task is undirected.*

**Proof:** For any state $s$ and applicable action $a$, $\overline{a}$ is applicable in $Result(s, \langle a \rangle)$ due to condition (3a) of invertibility. Conditions (1) and (2) make sure that $a$'s effects do in fact appear (condition (1) requires that each fact in the add list is inconsistent with at least one fact in the precondition), and conditions (3b) and (3c) make sure that $\overline{a}$ undoes exactly those effects. $\square$

We remark that, in contrast to what one may think at first sight, a task can have an undirected state space even if some actions are not invertible in the above sense. Imagine, for example, an action $a$ where $del(a) = \{p\}$ and $pre(a) = \{p'\}$, and, due to the domain semantics, if $p'$ is true then $p$ is also true. This means that $a$'s delete effect always appears; however, this can not be detected with the simple syntax check, $del(a) \subseteq pre(a)$, used in the definition above.

We next provide a new criterion that is weaker – more broadly applicable – than Lemma 1, and that only implies the non-existence of dead ends. The criterion is based on a weaker version of invertibility, and on two alternative properties whose combination can make an action "safe".

To make an action $a$ not lead into a dead end, it is already sufficient if the inverse action re-achieves *at least* what has been deleted, and does not delete any facts that have been true before. That is, given a state $s$ in which $a$ is applicable, applying $\overline{a}$ in $Result(s, \langle a \rangle)$ leads us back to a state $s'$ that satisfies $s' \supseteq s$. Formally, an action $a \in A$ is *at least invertible* if there is an action $\overline{a} \in A$ such that:

(1) $pre(\overline{a}) \subseteq (pre(a) \cup add(a)) \setminus del(a)$,

(2) $add(\overline{a}) \supseteq del(a)$, and

(3) $del(\overline{a})$ is inconsistent with $pre(a)$.

---

12. It may seem more natural to define inconsistency between fact sets in a symmetrical fashion, demanding that every fact in $F$ be inconsistent with *every* fact in $F'$. In our context here, that definition would be stronger than what we need.

Condition (1) here ensures, as before, that $\overline{a}$ is applicable in $Result(s, \langle a \rangle)$. Condition (2) ensures that $\overline{a}$ re-achieves every fact that was deleted by $a$. Condition (3) ensures that the facts deleted by $\overline{a}$ were not true in $s$ anyway. Note that any invertible action is also at least invertible. Conditions (1) and (2) are obviously given. As for condition (3), if $del(\overline{a}) = add(a)$ (condition (3c) of invertibility), and $add(a)$ is inconsistent with $pre(a)$ (condition (1) of invertibility), then $del(\overline{a})$ is inconsistent with $pre(a)$. So "invertible" is stronger than "at least invertible"; we chose the name "at least" for the latter to illustrate that, with this definition of invertibility, $\overline{a}$ potentially "re-"achieves *more* facts than we had in the original state $s$.

As an example, consider what happens if we modify the $move(l, l')$ action in Figure 2 to include a $visited(l')$ fact in its add list. The resulting action is no longer invertible because $move(l', l)$ does not delete $visited(l')$. If we apply, in state $s$, $move(l, l')$ and $move(l', l)$ in sequence, then now that gets us to a state $s'$ that is identical to $s$ except that it also includes $visited(l)$ and $visited(l')$, which may not have been true before. Move actions of this kind form the Simple-Tsp domain. They are at least invertible in the above sense: $pre(move(l', l)) = \{at(V, l')\} = add(move(l, l'))$; $add(move(l', l)) = \{at(V, l), visited(l)\} \supseteq \{at(V, l)\} = del(move(l, l'))$; $del(move(l, l')) = \{at(V, l')\}$ is inconsistent with $\{at(V, l)\} = pre(move(l, l'))$.

Another property implying that an action can not lead into dead ends is this. If the action must be applied at most once (because its add effects will remain true), and it deletes nothing but its own preconditions, then that action needs not be inverted. Formally, an action $a \in A$ has *static add effects* if:

$$add(a) \cap \bigcup_{a' \in A} del(a') = \emptyset.$$

An action $a \in A$ has *relevant delete effects*, if:

$$del(a) \cap (G \cup \bigcup_{a \neq a' \in A} pre(a')) \neq \emptyset.$$

If $del(a) \cap (G \cup \bigcup_{a \neq a' \in A} pre(a')) = \emptyset$, then we say that $a$ has *no* relevant delete effects, which is the property we will actually be interested in. In the illustrative task from Figure 2, imagine we disallow unloading an object at its initial location, and loading an object at its goal location. Then the remaining unload actions ($unload(O_1, L_2)$ and $unload(O_2, L_1)$) have static add effects – no action can delete the goal position of an object – and no relevant delete effects – the only action that needs an object to be in the vehicle is the respective unload at the goal location. Actions that have such characteristics are, for example, the actions that make passengers get out of the lift in Miconic-STRIPS (a passenger can get into the lift only at his/her origin floor, and get out of the lift only at his/her destination floor). Another example is contained in the Tireworld domain, where there is an action that inflates a flat wheel: there is no "de-flating" action and so the add effects are static; no action nor the goal needs a wheel to be flat so there are no relevant delete effects.

**Lemma 2** *Given a solvable STRIPS planning task* $(A, I, G)$. *If it holds for all actions* $a \in A$ *that either*

*1. a is at least invertible, or*

*2. a has static add effects and no relevant delete effects,*

*then the state space of the task is harmless.*

**Proof:** In short, to any reachable state $s = Result(I, \langle a_1, \ldots, a_n \rangle)$ a plan can be constructed by inverting $\langle a_1, \ldots, a_n \rangle$ (applying the respective inverse actions in the inverse order), and executing an arbitrary plan for $(A, I, G)$ thereafter. In these processes, actions that are not (at least) invertible can be skipped because by prerequisite they have static add effects and no relevant delete effects.

In more detail, the proof argument proceeds as follows. To any reachable state $s = Result(I, \langle a_1, \ldots, a_n \rangle) \in S$, we identify a solution $P$ for $(A, s, G)$. Let $\langle p_1, \ldots, p_m \rangle \in A^*$ be a solution for $(A, I, G)$ (which exists as $(A, I, G)$ is solvable by prerequisite). We construct $P$ with the algorithm shown in Figure 3.

$M := \emptyset$
**for** $i := n \ldots 1$ **do**
    **if** $a_i$ is at least invertible by $\overline{a_i}$ **then**
        **if** $\overline{a_i} \notin M$ apply $\overline{a_i}$ **endif**
    **else** $M := M \cup \{a_i\}$
    **endif**
**endfor**
**for** $i := 1 \ldots m$ **do**
    **if** $p_i \notin M$ **then** apply $p_i$ **endif**
**endfor**

Figure 3: Constructing plans in tasks where all actions are either at least invertible, or have static add effects and no relevant delete effects.

In the algorithm, $M$ serves as a kind of memory set for the actions that could not be inverted. We need to prove that the preconditions of all applied actions are fulfilled in the state where they are applied, and that the goals are true upon termination. Let us start with the first loop. We denote by $s_i := result(I, \langle a_1, \ldots, a_i \rangle)$ the state after executing the $i$th action on the path to $s$, and by $s_i'$ the state before the first loop starts with value $i$. We prove:

$$s_i' \supseteq (s_i \cap (G \cup \bigcup_{a \in A \setminus M_i} pre(a))) \cup \bigcup_{a \in M_i} add(a)$$

$M_i$ here denotes the current state of the set. We proceed by backward induction over $i$. If $i = n$, we got $s_i' = s_i$ and $M_i = \emptyset$, so the equation is trivially true. Now assume the equation is true for $i \geq 1$. We prove that the equation holds for $i - 1$. If $a_i$ is not at least invertible, then no action is applied, $s_{i-1}' = s_i'$, and $M_{i-1} = M_i \cup \{a_i\}$. Concerning the left hand side of the expression on the right hand side of the equation, we observe that $a_i$ does by prerequisite not delete any fact from $G \cup \bigcup_{a \in A \setminus M_{i-1}} pre(a)$ ($M_{i-1}$ contains $a_i$), so all relevant facts from $s_{i-1}$ have already been true in $s_i'$. Concerning the right hand side of the expression on the right, we observe that the facts in $add(a_i)$ are never deleted by prerequisite, so $\bigcup_{a \in M_{i-1}} add(a)$ is contained in $s_i'$. Now assume that $a_i$ is at least invertible by $\overline{a_i}$. We got $M_{i-1} = M_i$. Assume $\overline{a_i}$ is applied, i.e., $\overline{a_i} \notin M_i$. It is applicable because its preconditions are contained in $s_i$, and it is not an element of $M_i$. For the resulting state $s_{i-1}'$, all facts that $a_i$ has deleted from $s_{i-1}$ are added, and only facts are deleted that have not been true in $s_{i-1}$ anyway; also, none of the add effects of actions in $M_i$ is deleted, so the equation is fulfilled. Finally, if $\overline{a_i}$ is not applied, $\overline{a_i} \in M_i$, then $\overline{a_i}$ has static add effects and was applied before, so its add effects are contained in $s_i'$, and $a_i$'s delete effects are empty.

Inserting $i = 0$ in the equation we have just proved, we get

$$s_0 \supseteq (I \cap (G \cup \bigcup_{a \in A \setminus M_0} pre(a))) \cup \bigcup_{a \in M_0} add(a)$$

The second loop starts from $s_0$. So we start a solution plan, excluding the actions in a set $M_0$, from a state including all initial facts that are contained in the goal or in the precondition of any action not in $M_0$. As the state additionally contains all add effects of all actions in $M_0$, and those add effects are not deleted by any action, it is clear that we can simply skip the actions in $M_0$ and achieve the goal. □

As an example to illustrate the proof, consider a reachable state in the Tireworld domain. Every action is invertible, except the action that inflates a wheel. Say, as in the proof, we are in a state $s$ reached by the action sequence $\langle a_1, \ldots, a_n \rangle$. What the algorithm in Figure 3 will do is, undo everything we have done, by applying the respective $\overline{a_i}$ actions, except for the inflating actions $a_i$. The latter will be stored in the set $M$. This gets us to a state that is identical to the initial state, except that we have already inflated some of the flat wheels (those corresponding to the actions in $M$). From that state, the algorithm executes an arbitrary solution, skipping the previously applied inflating actions (in $M$).

## 3.2 Local Minima

We define an important kind of relationship between the role of an action in the real task and its role in the relaxed task. Combining this definition with the notions of at least invertible actions, and (no) relevant delete effects, yields a criterion that is sufficient for the non-existence of local minima under $h^+$ (or, equivalently, for 0 being an upper bound on the maximal local minimum exit distance). The criterion can be directly applied in 7 of the 30 investigated domains, and can be applied with slight modifications in 2 more domains. Many of the more individual proofs make use of similar, albeit somewhat more complicated, proof arguments.

The key property behind the lack of local minima under $h^+$ is, most of the time, that every action that is good for solving the real task is also good for solving the relaxed task. Formally, an action $a \in A$ is *respected by the relaxation* if:

*for any reachable state $s \in S$ such that $a$ starts an optimal plan for $(A, s, G)$, there is an optimal relaxed plan for $(A, s, G)$ that contains $a$.*

Note that one can assume the relaxed plan to start with $a$, since in the relaxation it can only be better to apply an action earlier.

All actions in the illustrative task from Figure 2 are respected by the relaxation. Consider the $move(l, l')$ actions, for example. If, in a state $s$, an optimal plan starts with $move(l, l')$, then there must be a good reason for this. Either a) at $l'$ there is an object that has yet to be transported, or b) an object is in the truck that must be transported to $l'$. In both cases, any relaxed plan must also transport the object, and there is no chance of doing so without moving to $l'$ at some point. Similarly, if an optimal plan starts with a $load(o, l)$ action, then this means that $o$ must be transported somewhere else, and the relaxed plan does not get around loading it. Finally, if an optimal plan starts with an $unload(o, l)$ action, then this means that $l$ is the goal location of $o$, and any relaxed plan will have to include that action.

Similar arguments as the above can be applied in many transportation domains. The argument regarding move actions becomes a little more complicated if there are non-trivial road maps, unlike in the illustrative example where there are only two locations that are reachable in a single step from each other. Say the road map is a (any) directed graph, and we modify the move action from Figure 2 only in that we add a precondition fact demanding the existence of an edge from $l$ to $l'$. Then all move actions are still respected by the relaxation, because ignoring delete lists does not affect the shape of the road map. Any optimal real path from a location $l$ to a location $l'$ coincides with an optimal relaxed path of movements from $l$ to $l'$ (even though the result of executing the path will be different). From there, the claim follows with the same argument as above, namely, that an optimal plans moves from $l$ to $l'$ only if some object provides a reason for doing so.

If a transportation domain features additional constraints on, or side effects of, the move actions, then they may not be respected by the relaxation. We give an example below, after formulating our main lemma regarding local minima under $h^+$.

Note that there can exist local minima even if all actions are respected by the relaxation. Consider the following transportation task, featuring single-directional edges in the road map graph. As argued above, all actions are respected by the relaxation. A vehicle and two objects $o_1$, $o_2$ are initially at $l$; $o_1$ must go to $l_1$ and $o_2$ must go to $l_2$; the edge from $l$ to $l_1$ is single-directed and the edge from $l$ to $l_2$ is single-directed; between $l_1$ and $l_2$, there is a path of $n$ bi-directional (undirected) edges. The optimal relaxed plan for the state $s$ where, from the initial state, $o_1$ and $o_2$ were loaded, has length 4: move from $l$ to $l_1$ and $l_2$, and unload $o_1$ and $o_2$ at $l_1$ and $l_2$, respectively. However, once one moved, in $s$, to either $l_1$ or $l_2$, the optimal relaxed plan length goes up to $n + 2$, since the entire path between $l_1$ an $l_2$ must be traversed. So $s$ lies on a local minimum, given that $n > 2$; note that, by setting $n$ to arbitrarily high values, we get a local minimum with arbitrarily large exit distance.

It turns out that preventing the above example, precisely, making use of the notions of invertibility and of relevant delete effects, as introduced above, suffices to get rid of local minima under $h^+$.

**Lemma 3** *Given a solvable STRIPS task $(A, I, G)$, such that the state space $(S, T)$ does not contain unrecognized dead ends. If each action $a \in A$*

*1. is respected by the relaxation, and*

*2. is at least invertible or has no relevant delete effects,*

*then there are no local minima in $(S, T)$ under evaluation with $h^+$.*

**Proof:** The states with $gd(s) = \infty$ are not on local minima by prerequisite, with $h^+(s) = \infty$. We will prove that, in every reachable state $s$ with $0 < gd(s) \neq \infty$, if an action $a$ starts an optimal plan for $(A, s, G)$, then $h^+(Result(s, \langle a \rangle)) \leq h^+(s)$. This proves the lemma: iterating the argument, we obtain a path from $s$ to a goal state $s'$, where the value of $h^+$ does *not increase* on the path. This means that from $s$ an exit is reachable on a flat path – $h(s') = 0 < h(s)$ so at some point on the path the $h^+$ value becomes lower than $h(s)$, thus $s$ can not lie on a local minimum.

Let $s$ be a reachable state with $0 < gd(s) \neq \infty$. Let $a$ be an action that starts an optimal plan for $(A, s, G)$. We denote $s' := Result(s, \langle a \rangle)$. The action is respected by the relaxation, so there is an optimal relaxed plan $P^+(s)$ for $(A, s, G)$ that starts with $a$.

Case (A), removing $a$ from $P^+(s)$ yields a relaxed plan for $(A, s', G)$. Then $h^+(s') < h^+(s)$ follows, and we are finished. This is the case, in particular, if $a$ has no relevant delete effects: the facts that $a$ deletes are not needed by any other action nor by the goal, so $P^+(s)$ without $a$ achieves the goal starting from $s'$ (where $a$ has already been applied).

Case (B), assume removing $a$ from $P^+(s)$ does *not* yield a relaxed plan for $s'$. Then, with what was said before, $a$ does have relevant delete effects, and must thus be at least invertible. That is, there is an action $\overline{a} \in A$ with $pre(\overline{a}) \subseteq (pre(a) \cup add(a)) \setminus del(a)$ and $add(\overline{a}) \supseteq del(a)$. The action $\overline{a}$ is guaranteed to be applicable in $s'$, and it re-achieves $a$'s delete effects. Denote by $P^+(s')$ the action sequence that results from replacing, in $P^+(s)$, $a$ with $\overline{a}$. Then $P^+(s')$ is a relaxed plan for $(A, s', G)$. This can be seen as follows. Observe that, by definition, $P^+(s)$ without $a$ is a relaxed plan for $Result(s, \langle a^+ \rangle)$ (we abbreviate the notation somewhat to improve readability). The desired property now follows because $Result(s', \langle \overline{a}^+ \rangle)$ is a superset of $Result(s, \langle a^+ \rangle)$: we have $Result(s, \langle a^+ \rangle) = s \cup add(a)$, $s' = (s \cup add(a)) \setminus del(a)$, and $add(\overline{a}) \supseteq del(a)$. So $P^+(s')$ is a relaxed plan for $(A, s', G)$, yielding $h^+(s') \leq h^+(s)$. □

The proof to Lemma 3 demonstrates along which lines, typically, the proof arguments in this investigation proceed. Given a state $s$, consider an action $a$ that starts an optimal plan for $s$, and consider an optimal relaxed plan $P^+$ for $s$ (that contains $a$, ideally). Then, determine how $P^+$ can be modified to obtain a relaxed plan for the state that results from $a$'s execution. This technique forms the basis of literally all proofs except those concerned with dead ends. Note that the second prerequisite of Lemma 3 is fulfilled by planning tasks qualifying for the undirectedness or harmlessness criteria given by Lemmas 1 and 2. Note also that, with what was said above, we have now proved that the state space of the illustrative example in Figure 2 is undirected, and does not contain any local minima under $h^+$.

Domains where all actions are respected by the relaxation are, for example, the STRIPS transportation domains Logistics, Gripper, Ferry, and Miconic-STRIPS. In all these cases, the respective proof arguments are very similar to what we said above. It is instructive to have a look at some examples where an action is *not* respected by the relaxation. In a transportation domain, this can, for example, happen due to fuel usage as a "side effect" of moving. Concretely, in the Mystery above, applying a move action deletes a fuel unit at the start location (the location in that the move starts). If fuel is running low at some locations, a (real) plan may have to move along fuel-rich deviations in the road map. A relaxed plan does not need to do that – it can always move along the shortest connections on the map – because, there, the actions do not delete the fuel units.

Formulated somewhat more generally, relaxed plans can take "short-cuts" that don't work in reality. If these short-cuts are disjoint (in the starting actions) with the real solution paths, then local minima may arise even if all actions are (at least) invertible. In the above discussed transportation case, the short-cuts correspond in a very intuitive manner to what one tends to think about as short-cuts (on a road map, namely). This is not the case in general, i.e., in other kinds of domains. Consider the Blocksworld-arm state depicted in Figure 4.

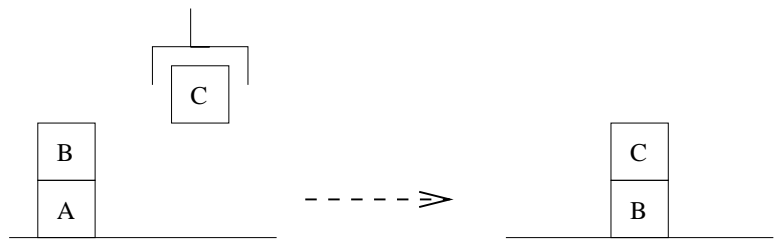

Figure 4: A local minimum state in Blocksworld-arm. The goal is to have B on the table, and C on B.

In the depicted state, denoted $s$, B is on A is on the table, and the arm holds C. The goal is to have B on the table, and C on B.[13] The only optimal plan for $s$ is to put C down on the table, then unstack B from A and put it down on the table, then pickup C and stack it onto B. The only optimal *relaxed* plan for $s$, however, is to stack C onto B immediately, then unstack B from A, then put B down to the table. The "short-cut" here is that the relaxed plan does not have to put C down on the table, because stacking C onto B does not delete the fact that declares B's surface as unoccupied. As a result, $s$ lies on a local minimum under $h^+$.[14] The reason, intuitively, why $h^+$ does not yield any local minima in many domains, is that "vicious" short-cuts like in this example just don't happen.

---

13. Usually, in Blocksworld there are no goals demanding a block to be on the table. In the example, this is done only for the sake of simplicity: one could just introduce one more block D and demand that B be on D for the goal.

### 3.3 Benches

We could not find a nice general sufficient criterion implying upper bounds on the maximal exit distance from local minima – except the special case above where there are no local minima at all and thus 0 is an upper bound on the maximal local minimum exit distance. We did, however, find a simple proof argument determining an upper bound on the maximal exit distance from benches, in tasks that qualify for the application of Lemma 3. The proof argument works, sometimes with slight modifications, in all the 7 domains where Lemma 3 can be directly applied – in all these domains, the maximal bench exit distance is bounded by 1 (bounded by 0, in one case).

The proof argument is based on observing that, in many domains, some of the actions have only delete effects that are irrelevant (for the relaxed plan, at least) once the action was applied on an optimal solution path. Formally, an action $a \in A$ has *relaxed-plan relevant delete effects* if:

*for any reachable state $s \in S$ such that $a$ starts an optimal plan for $(A, s, G)$, there is no optimal relaxed plan $\langle a, a_1, \ldots, a_n \rangle$ for $(A, s, G)$ such that $del(a) \cap (G \cup \bigcup_{i=1}^{n} pre(a_i)) = \emptyset$.*

If, for any reachable state $s \in S$ such that $a$ starts an optimal plan for $(A, s, G)$, there *is* an optimal relaxed plan $\langle a, a_1, \ldots, a_n \rangle$ for $(A, s, G)$ such that $del(a) \cap (G \cup \bigcup_{i=1}^{n} pre(a_i)) = \emptyset$, then we say that $a$ has *no* relaxed-plan relevant delete effects, which is the property we will actually be interested in. With this notation, if $a$ has no relaxed-plan relevant delete effects, and it starts an optimal plan for $s$, then a relaxed plan for $Result(s, \langle a \rangle)$ can be constructed as the sequence $\langle a_1, \ldots, a_n \rangle$, i.e., by skipping $a$ from the relaxed plan for $s$. Thus the $h^+$ value decreases from $s$ to $Result(s, \langle a \rangle)$. Note that $n$ can be set to 0 if $a$ results in a goal state from $s$. Note also that, by definition, any action with no relaxed-plan relevant delete effects is respected by the relaxation; if an action is not respected by the relaxation, then we can not claim anything about $h^+$ anyway. Note finally that, assuming an action $a$ that is respected by the relaxation, if $a$ has no relevant delete effects, i.e., if $a$ does not delete a goal or any precondition of another action, then $a$ also has no relaxed-plan relevant delete effects in the sense of our definition.

Consider again our illustrative example from Figure 2. Say we have a state $s$ in which $load(o, l)$ starts an optimal plan. This means that $o$ has yet to be transported, to a location $l' \neq l$. In particular, it means that $at(o, l)$ is not a goal, and it follows that the action – whose only delete effect is $at(o, l)$ – has no relevant delete effects (no other action has $at(o, l)$ in its precondition). Further, say $unload(o, l)$ starts an optimal plan in $s$. This means that $l$ is the goal location of $o$. After applying the action, the goal for $o$ will be achieved, and no action will need to refer to $o$ again, in particular no action will require $o$ to be inside the vehicle, which is the only delete effect of $unload(o, l)$. So that action neither has relaxed-plan relevant delete effects. In contrast, consider the $move(l, l')$ action, that deletes $at(V, l)$. Say we are in the state $s$ where $O_1$ has been loaded into $V$ from the initial state of the task. Then $move(L_1, L_2)$ starts an optimal plan for $s$, and any relaxed plan for $Result(s, \langle move(L_1, L_2) \rangle)$ has to include the action $move(L_2, L_1)$, moving back from $L_2$ to $L_1$ in order to be able to transport $O_2$. So the delete effect of $move(L_1, L_2)$, namely $at(V, L_1)$, *is* relaxed-plan relevant.

---

14. We have $h^+(s) = 3$. The $h^+$ value after, in $s$, putting C down to the table is 4 (any relaxed plan has to apply two actions for each of the two goals). The $h^+$ value after stacking, in $s$, C onto B is still 3 (the relaxed plan is unstack C B, unstack B A, put down B), but from there the only successor state is to unstack C from B again, going back to $s$.

If, in a task satisfying the prerequisites of Lemma 3, an optimal starting action has no relaxed-plan relevant delete effects, then one can apply case (A) in the proof of Lemma 3, and obtain a smaller $h^+$ value. To bound the maximal exit distance from benches, all we need to do is to identify a maximum number of steps after which that will happen.

**Lemma 4** *Given a solvable STRIPS task $(A, I, G)$ that satisfies the prerequisites of Lemma 3. Let $d$ be a constant so that, for every non dead-end state $s \in S$, there is an optimal plan $\langle a_1, \ldots, a_n \rangle$ where the $d$-th action, $a_d$, has no relaxed-plan relevant delete effects. Then $mbed(S, T) \leq d - 1$.*

**Proof:** Let $s$ be a reachable state with $0 < gd(s) \neq \infty$. Let $\langle a_1, \ldots, a_n \rangle$ be an optimal plan for $(A, s, G)$, where $a_d$ has no relaxed-plan relevant delete effects. Denote, for $0 \leq i \leq n$, $s_i := Result(s, \langle a_1, \ldots, a_i \rangle)$. With the argumentation in Lemma 3, we have $h^+(s_i) \leq h^+(s)$ for all $i$. Consider the state $s_{d-1}$. By prerequisite, there is an optimal relaxed plan for $(A, s_{d-1}, G)$ that has the form $\langle a_d, a'_1, \ldots, a'_m \rangle$, where $del(a_d) \cap (G \cup \bigcup_{i=1}^{m} pre(a'_i)) = \emptyset$. But then, obviously, $\langle a'_1, \ldots, a'_m \rangle$ is a relaxed plan for $s_d$, and so $h^+(s_d) \leq h^+(s_{d-1}) - 1$. The distance from $s$ to $s_{d-1}$ is $d - 1$, and so the lemma follows. □

Lemma 4 can be directly applied in 5 of the 7 domains that qualify for Lemma 3. Its proof argument can, in a somewhat more general version, be applied in the 2 other domains as well – namely, in Ferry and Gripper, where loading an object deletes space in the vehicle – and in one more domain – namely, Miconic-SIMPLE, that uses some simple ADL constructs. In the other domains where we proved an upper bound on the maximal exit distance from benches (and/or an upper bound on the maximal exit distance from local minima), the proof arguments are (a lot, sometimes) more complicated. Reconsidering the illustrative example, as stated above the *load* and *unload* actions have no relaxed-plan relevant delete effects, while the *move* actions do. Now, obviously, since the two locations are accessible from each other with a single move, no optimal plan applies more than one move action in a row, i.e., in any optimal plan the first or second action will be a *load/unload*. With Lemma 4 this tells us that the maximal exit distance from benches is bounded by 1. A very similar argument can be applied in all other transportation domains where every pair of locations is connected via a single move (as in, for example, Logistics). More generally, in (the standard encoding of) a transportation domain with no other constraints (regarding, for example, fuel), and with an undirected road map graph, the exit distance is bounded by the *diameter* of the road map graph, i.e., by the maximum distance of any two locations (nodes) in the graph. The "worst" thing a solution plan might have to do is to traverse the entire road map before loading/unloading an object.[15]

---

15. With directed road map graphs, as explained above, local minima can arise. More technically, Lemma 3 can not be applied, and so Lemma 4 can not be applied either.

## 4. A Planning Domain Taxonomy

We now list our proved results, with brief explanations of how we obtained these results. We then summarize the results in the form of a planning domain taxonomy.

We group the "positive" results – those which prove the non-existence of topological phenomena that are problematic for heuristic search – together in single theorems. The "negative" results are shown separately by sketching counter examples. We consider dead ends, local minima, and benches in that order. Remember that, with respect to dead ends, the only problematic case for heuristic search is when there are unrecognized dead ends, c.f. Section 2.3.

**Theorem 1** *The state space of any solvable instance of*

1. *Blocksworld-arm, Blocksworld-no-arm, Briefcaseworld, Depots, Driverlog, Ferry, Fridge, Gripper, Hanoi, or Logistics is undirected,*

2. *Grid, Miconic-SIMPLE, Miconic-STRIPS, Movie, Pipesworld, PSR, Satellite, Simple-Tsp, Tireworld, or Zenotravel is harmless,*

3. *Dining-Philosophers, Optical-Telegraph, Rovers, or Schedule is recognized under evaluation with $h^+$.*

In Blocksworld-arm, Blocksworld-no-arm, Driverlog, Ferry, Gripper, Hanoi, and Logistics, Lemma 1 can be directly applied. In Briefcaseworld, Depots, and Fridge, due to some subtleties the actions are not invertible in the syntactical sense, but it is easy to show that every action has an inverse counterpart. In Movie, Miconic-STRIPS, Simple-Tsp, and Tireworld, Lemma 2 can be directly applied, in Grid and Miconic-SIMPLE similar proof arguments as used in Lemma 2 suffice. In Pipesworld, PSR, Satellite, and Zenotravel, some easy-to-see more individual domain properties prove the absence of dead ends. In the domains where all dead ends are recognized by $h^+$, the individual domain properties exploited in the proofs are somewhat more involved. For example, in Rovers there is a plan to a state if and only if, for all soil/rock samples and images that need to be taken, there is a rover that can do the job, and that can communicate the gathered data to a lander. The only chance to run into a dead end is to take a soil/rock sample with a rover that can not reach a lander (the soil/rock sample is available only once). But then, there is no relaxed plan to the state either.

In the 6 domains not mentioned in Theorem 1 (Airport, Assembly, Freecell, Miconic-ADL, Mprime, Mystery), it is easy to construct arbitrarily deep unrecognized dead ends (arbitrarily long paths of unrecognized dead ends). For example, in Mystery and Mprime the relaxed plan can still achieve the goal in situations where too much fuel was consumed already; in Airport, two planes that block each other's paths may move "across" each other in the relaxed plan.

The positive results regarding local minima are these.

**Theorem 2** *Under $h^+$, the maximal local minimum exit distance in the state space of any solvable instance of*

1. *Blocksworld-no-arm, Briefcaseworld, Ferry, Fridge, Grid, Gripper, Hanoi, Logistics, Miconic-SIMPLE, Miconic-STRIPS, Movie, Simple-Tsp, or Tireworld is* 0,

2. *Zenotravel is at most* 2, *Satellite is at most* 4, *Schedule is at most* 5, *Dining-Philosophers is at most* 31.

In Ferry, Gripper, Logistics, Miconic-STRIPS, Movie, Simple-Tsp, and Tireworld, Lemma 3 can be applied. In Fridge and Miconic-SIMPLE, the actions do not adhere syntactically to the definitions of invertibility and (no) relevant delete effects, but have similar semantics. So Lemma 3 can not be directly applied, but similar arguments suffice: it is easy to see that all actions are respected by the relaxation, and the proof of Lemma 3 can be individually adapted to take into account the particular properties regarding invertibility and relevant delete effects. (For example, if a passenger gets out of the lift in Miconic-SIMPLE, then the delete effect is that the passenger is no longer inside the lift, which does not matter since the passenger has reached her destination.) In Blocksworld-no-arm, Briefcaseworld, and Grid, rather individual (and sometimes quite involved) arguments prove the absence of local minima under $h^+$. The proof method is, in all cases, to consider some state $s$ and identify a flat path from $s$ to a state with better $h^+$ value. For example, in Grid this is done by moving along a path of locations contained in the relaxed plan for $s$, until a key can be picked up/put down, or a lock can be opened (this is a very simplified description, the actual procedure is quite complicated). In Hanoi, one can prove that the optimal relaxed solution length for any state is equal to the number of discs that are not yet in their final goal position. This suffices because no optimal plan moves a disc away from its final position. Note that, thus, the Hanoi state spaces under $h^+$ are a sequence of benches decreasing exponentially in diameter and size.

In Zenotravel, Satellite, and Schedule, the proofs proceed by identifying a constant number of steps that suffices to execute one action $a$ in the optimal relaxed plan for a state $s$, and, without deleting $a$'s relevant add effects, to re-achieve all relevant facts that were deleted by $a$. In Dining-Philosophers (as well as Optical-Telegraph), due to the subtleties of the PDDL encoding – which was, as said, obtained by an automatic compilation from the automata-based "Promela" language (Edelkamp, 2003a) – $h^+$ is only very loosely connected to goal distance: in the relaxation, an automaton (for example, a philosopher) can always "block itself" with at most 3 actions. The bound for Dining-Philosophers follows from the rather constant and restrictive domain structure, where a constant number of process transitions, namely 6, always suffices to block one more philosopher. The proved bound is derived from this, by considering that 4 planning actions are needed for each process transition, and that certain additional actions may be needed due to the subtleties of the PDDL encoding (where a process can be "in between" two of its internal states). We remark that the bound is valid even for the trivial heuristic function returning the number of yet un-blocked philosophers. In fact, the proof for $h^+$ can be viewed as a corollary of a proof for this heuristic function; we get back to this at the end of this section. We finally remark that the highest exit distance under $h^+$ that we could actually construct in Dining-Philosophers was 15. We conjecture that this is a (tight) upper bound.

In Satellite, Schedule, and Zenotravel, the proved upper bounds are tight. In all of Dining-Philosophers, Satellite, Schedule, and Zenotravel, the bounds are valid for *any* non-dead end state $s$. So, beside a bound on the local minimum exit distance, these results also provide a bound on the bench exit distance, and will be re-used for that below in this section.

In Airport, Assembly, Freecell, Miconic-ADL, Mprime, and Mystery, as stated above there can be unrecognized dead ends, so by Proposition 1 the local minimum exit distance in these domains is unbounded. In all other domains not mentioned in Theorem 2, i.e., in Blocksworld-arm, Depots, Driverlog, Optical-Telegraph, Pipesworld, PSR, and Rovers, one can construct local minima with arbitrarily large exit distances. The most complicated example is Optical-Telegraph, where, in difference to Dining-Philosophers, one can construct situations where the number of process state transitions needed to block one more process is arbitrarily high. Optical-Telegraph is basically a version of Dining-Philosophers with more complicated philosophers, that have more freedom of what to do next. This freedom enables situations where a whole row of philosophers at the table must perform two transitions each in order to block one more philosopher. Details are in Appendix A.2. A simpler example is Blocksworld-arm (as well as Depots, in which Blocksworld-arm situations can be embedded). Consider the following situation. There are $n$ blocks $b_1, \ldots, b_n$ that initially form a stack where $b_i$ is on $b_{i+1}$ and $b_n$ is on the table. The goal is to build the same stack on top of another block $b_{n+1}$, i.e., the goal is a stack $b_1, \ldots, b_n, b_{n+1}$. Reaching, from the initial state, a state with better $h^+$ value involves disassembling the entire stack $b_1, \ldots, b_n$. During the disassembling process, $h^+$ increases. Note that this is basically an extended version of the illustrative example from Figure 4.

As an interesting side remark, note that we have now proved a topological difference between Blocksworld-arm and Blocksworld-no-arm: in the latter, there are no local minima at all under $h^+$, in the former, the exit distance from them can be arbitrarily large. While this is intriguing, it is not quite clear if there is a general message to learn from it. One might interpret it as telling us, in a formal way, that encoding details can have a significant impact on topology, and with that on search performance. FF, for example, is much more efficient in Blocksworld-no-arm than in Blocksworld-arm. It should be noted, however, that the two domains differ also semantically, namely in that plans in Blocksworld-no-arm are half as long as plans in Blocksworld-arm. From a practical point of view, it would be interesting to explore if this Blocksworld observation can be generalized into encoding methods trying to model a domain in a way making it best suited for $h^+$. Some more on this is said in Section 7.

The positive results regarding benches are these.

**Theorem 3** *Under $h^+$, the maximal bench exit distance in the state space of any solvable instance of Simple-Tsp is 0, Ferry is at most 1, Gripper is at most 1, Logistics is at most 1, Miconic-SIMPLE is at most 1, Miconic-STRIPS is at most 1, Movie is at most 1, Zenotravel is at most 2, Satellite is at most 4, Schedule is at most 5, Tireworld is at most 6, and Dining-Philosophers is at most 31.*

In Simple-Tsp, Ferry, Gripper, Logistics, Miconic-STRIPS, Movie, and Tireworld, Lemma 4 can be directly applied. Determining what actions have (no) relaxed-plan relevant delete effects is easy in all the domains; in Tireworld it is somewhat complicated to see when, at the latest, such an action can be applied in an optimal plan. For Miconic-SIMPLE, similar arguments as in Lemma 4 suffice. For Zenotravel, Satellite, Schedule, and Dining-Philosophers, the respective bounds were shown above already.

Note that, in Simple-Tsp, we proved that there are no local minima and that the exit distance is 0. This implies that $h^+$ is, in fact, identical to the real goal distance: the entire state space consists of contours and global minima.

Our topological distinctions divide planning domains into a taxonomy of classes which differ in terms of the behavior of their state spaces with respect to $h^+$. A visualization of the taxonomy, with the results for the 30 investigated domains, is given in Figure 5.

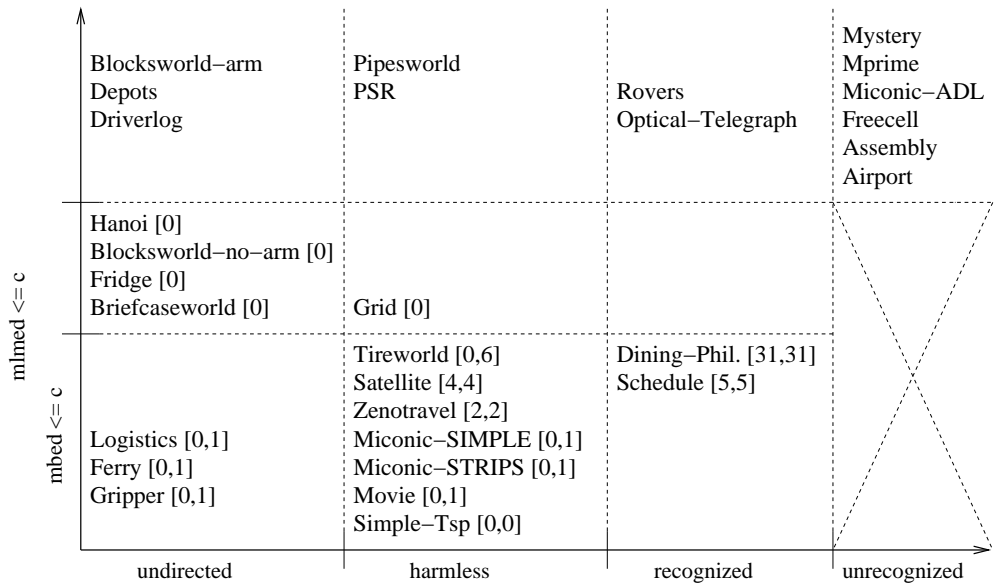

Figure 5: A planning domain taxonomy, overviewing our results.

The taxonomy, as shown in Figure 5, has two dimensions. The $x$-axis corresponds to the four dead end classes. The $y$-axis corresponds to the existence or non-existence of constant upper bounds on the local minimum exit distance, and on the bench exit distance. Note that this visualization makes the simplifying assumption that the domains with bounded bench exit distance are a subset of the ones with bounded local minimum exit distance. This assumption is not justified in general, but holds true in our specific collection of domains. Also, the question whether there is a bound on the difficulty of escaping benches does not seem as relevant when, anyway, it can be arbitrarily difficult to escape local minima.[16] The specific bounds proved for the individual domains are given in parentheses, local minimum exit distance bound preceding bench exit distance bound in the cases where there are both. The bottom right corner of the taxonomy is crossed out because no domain can belong to the respective classes.[17]

_______________________

16. Similarly, when benches can be arbitrarily large it is not as relevant if or if not the local minima are small or non-existent. In that sense the respective results for Briefcaseworld, Fridge, Grid, Blocksworld-no-arm, and Hanoi are only moderately important. Still they constitute interesting properties of these domains.

17. By Proposition 1, the existence of unrecognized dead ends implies the non-existence of constant upper bounds on the local minimum exit distance, given there are no states with $gd(s) \neq 0$ but $h^+(s) = 0$. Such states can exist, but only if the domain features derived predicates that appear negated in the negation

What Figure 5 *suggests* is that $h^+$-approximating heuristic planners are fast because many of the common benchmark domains lie in the "easy" regions of the taxonomy. More concretely, as described in the introduction, when provided with the $h^+$ function, FF's search algorithm enforced hill-climbing is polynomial in the domains located in the lowermost classes of the taxonomy (i.e., in domains with constant bounds on both maximal exit distances). From a more empirical perspective, the distinction lines in the taxonomy coincide quite well with the practical performance of FF. FF excels in 11 of the 12 domains that belong to the lowermost classes of the taxonomy (the more difficult domain is Dining-Philosophers, whose upper bound is exceptionally high). In the 5 "middle" domains (no local minima but potentially large benches) FF performs well, but does not scale up as comfortably as in the easier domains. As for the more complex domains: Blocksworld-arm, Depots, Driverlog, Optical-Telegraph, Pipesworld, and PSR are amongst the most challenging domains for FF. In Mprime and Mystery, FF performs just as bad as most other planners. In Freecell and Miconic-ADL, FF is among the top performing planners, but often runs into unrecognized dead ends in the larger instances (for example, the larger Freecell instances used at AIPS-2000). In Airport, Assembly and Rovers, FF performs pretty well in the respective competition example suites; however, in these domains the competition suites hardly explore the worst-cases of the domain topology (details on this are in Appendix A).

We do not discuss in detail the relation between the taxonomy and the empirical performance of all the other heuristic planners that make use of an $h^+$ approximation in one or the other way. One observation that can definitely be made is that all these planners have no trouble in solving instances from the domains with the most extreme $h^+$ properties. In Simple-Tsp, Ferry, Gripper, Logistics, Miconic-SIMPLE, Miconic-STRIPS, and Movie, to some extent also Zenotravel, all such planners scale up very comfortably. In particular, they scale up much more comfortably in these domains than they typically do in the other domains, at least without additional (for example, goal ordering) techniques.

In the next section, we treat the connection between the taxonomy and FF's performance in a more analytical way, by relating the properties of $h^+$ to properties of FF's approximation of $h^+$, called $h^{FF}$. Before we do so, some remarks on the relation of the taxonomy to complexity theory are in order. The question is whether there is a provable relation, i.e., a relation between the distinction lines in the taxonomy, and the complexity of deciding plan existence in the respective domains. We were able to construct an **NP**-hard domain (a domain where deciding plan existence is **NP**-hard) where $h^+$ does not yield any local minima; the maximal bench exit distance in that domain is, however, unbounded. We tried, but we were not able to come up with an **NP**-hard domain that has constant bounds on both maximal exit distances. It remains an open question whether such a domain exists or not. If the answer is "yes", then the lowermost classes of the taxonomy form a group of domains that are worst-case hard, but typically very easy to solve (at least as far as

---

normal form of the goal condition. But even then, in the presence of unrecognized dead ends there would be "fake"-global minima, i.e., global minima consisting of non-solution states, in fact consisting of unrecognized dead ends.

reflected by the hitherto benchmarks). If the answer is "no", then we have identified a very large polynomial sub-class of planning.[18]

Talking about polynomial sub-classes, an intriguing observation can be made here about the trivial heuristic function returning, for a state $s$, the number of goals that are not true in $s$. Let's call this function $h^G$. With a little thinking, one realizes that, in fact, *all the 12 domains where we proved constant bounds on both maximal exit distances under $h^+$ also have such constant bounds under $h^G$*. On the other hand, for the remaining 18 of the 30 domains (except Miconic-ADL) it is easy to see that there are no constant bounds for $h^G$. In Logistics, for example, clearly the maximum number of steps needed to achieve one more goal is 12: 4 steps each (move, load, move, unload) within a package's origin city, between the origin city and the destination city, and within the destination city. In Dining-Philosophers, for example, the upper bound for $h^+$ was, as said, proved as a corollary of an upper bound for $h^G$. In Blocksworld, for example, clearly it can take arbitrarily many steps to achieve one more goal, namely if a block that must be moved is buried beneath $n$ other blocks that do not need to be moved.

While the above observation appears rather significant at first sight, it is probably not very important, neither in theory nor in practice. For one thing, it is a coincidence that, here, the set of domains with both constant bounds under $h^+$ is the same as the set of domains with both constant bounds under $h^G$. A simple counter example for the general case is a "graph-search" domain, where the task is to find a path between two nodes in a directed graph, using the obvious "at"-predicate and "connected"-predicate based encoding. There, $h^+$ is equal to the real goal distance (since one never needs to move back), while $h^G$ can, clearly, be arbitrarily bad. For another thing, while domains like Logistics have constant exit distance bounds under $h^G$, these bounds are too large to be practically useful. For example, with $h^+$, FF needs to look at most 2 steps forward in each breadth-first search iteration of enforced hill-climbing, in any Logistics instance. With $h^G$, breadth-first searches up to depth 12 would be needed. So, at most, the observation regarding $h^G$ is a noteworthy statement about the current planning benchmarks. It remains an open question whether the (coincidental) correspondence between the bounds for $h^+$, and for $h^G$, in the investigated 30 domains, can be exploited for, e.g., detecting such bounds automatically.

## 5. Relating $h^+$ to $h^{FF}$

Our discussion relating $h^+$ to $h^{FF}$ is structured in two separate sections. The first one briefly discusses provable relations between $h^+$ and $h^{FF}$. The second section summarizes the results of a large-scale empirical investigation aimed at identifying to what extent the topological properties of $h^+$, in the benchmarks, get preserved by $h^{FF}$.

### 5.1 Provable Relations between $h^+$ and $h^{FF}$

One thing that is very easy to observe is that the behavior of $h^+$ and $h^{FF}$ is provably the same with respect to dead ends, i.e., both heuristics return $\infty$ in the same cases. This is simply because both heuristics return $\infty$ in a state $s$ iff there is no relaxed plan for $s$.

---

18. Presumably, to prove the latter, one would need to characterize that class in a purely *syntactic* manner on the level of PDDL definitions, since $h^+$ is derived directly from the PDDL syntax. The author's wild guess it that this is not going to work, and that the answer is "yes".

For $h^+$ this follows by definition. For $h^{FF}$ it follows from the completeness, relative to the relaxation, of the algorithm that computes relaxed plans (Hoffmann & Nebel, 2001a). That algorithm is a relaxed version of Graphplan (Blum & Furst, 1995, 1997). In each state $s$, FF runs Graphplan on the task where $s$ is the initial state, and the delete lists of all actions are empty. Without delete lists, Graphplan is guaranteed to terminate in polynomial time. If Graphplan terminates unsuccessfully, then $h^{FF}(s)$ is set to $\infty$. Otherwise, the number of actions in the returned plan is taken as the heuristic value $h^{FF}(s)$ of the state.[19] Graphplan is a complete algorithm – it terminates successfully if and only if there is a plan – and so $h^{FF}$ is set to $\infty$ iff there is no relaxed plan for $s$. It follows that the dead end classes of the benchmarks are the same under $h^+$ and $h^{FF}$.

The relaxed plans found by Graphplan have (just as in general STRIPS) the property that they are optimal in terms of the number of parallel time steps, but not in terms of the number of actions. So, in general, $h^{FF}$ is not the same as $h^+$ (even if **P** is the same as **NP**). FF uses the following heuristic techniques for action choice in relaxed Graphplan, aiming at minimizing the number of selected actions (Hoffmann & Nebel, 2001a). First, if a fact can be achieved by a NOOP (a dummy action propagating a fact from time step $t$ to time step $t + 1$ in Graphplan's planning graph), then that NOOP is selected. This guarantees that every non-NOOP action is selected at most once (of course, selected NOOP actions are not counted into the relaxed plan). Second, if there is no NOOP available then an action with minimal precondition weight is chosen, where "weight" is defined as the summed-up indices of the first layers of appearance (in the planning graph) of the precondition facts. Third, actions selected at the same parallel time step are assumed to be linearized by order of selection; so an action $a$ selected after $a'$ will be assumed to achieve a fact $p \in add(a) \cup pre(a')$ even if $a$ and $a'$ are selected at the same parallel time step.

There are two very restrictive sub-classes of STRIPS in which $h^{FF}$ is provably the same as $h^+$. The first demands that every fact has at most one achiever.

**Proposition 2** *Let $(A, I, G)$ be a STRIPS planning task so that, for all facts $p$, there is at most one action $a \in A$ with $p \in add(a)$. Then, for all states $s$ in the task, $h^+(s) = h^{FF}(s)$.*

**Proof:** The proposition follows from the observation that, when running relaxed Graphplan, the only choice points are those for action selection; these choice points will always be empty or unary in our case. This implies that all actions selected by Graphplan are contained in any relaxed plan. In more detail, the latter can be proved by an induction over the regression steps in relaxed Graphplan. Let $s$ be a state for which there is a relaxed plan. At the top level of the regression, actions $a$ are selected to support all goals that are not contained in $s$. These goals need to be supported in any relaxed plan, and there are no other actions for doing so. The same holds true for the preconditions of the selected actions: if $p \in pre(a)$ is not in $s$, then a supporter must be present in any relaxed plan, and that supporter will be selected by relaxed Graphplan. Iterating the argument, we get the desired property. The claim then follows because, as proved by Hoffmann and Nebel (2001a), relaxed Graphplan selects every action at most once. $\square$

---

19. Note that this is an estimate of *sequential* relaxed plan length. The length of the planning graph built by Graphplan corresponds to the optimal length of a parallel relaxed plan, an admissible heuristic estimate. However, as indicated before, such heuristic functions have generally not been found to provide useful search guidance in practice (see, for example, Haslum & Geffner, 2000; Bonet & Geffner, 2001b).

Our second sub-class of STRIPS demands that there is at most one goal, and at most one precondition per action.

**Proposition 3** *Let $(A, I, G)$ be a STRIPS planning task so that $|G| \leq 1$ and, for all $a \in A$, $|pre(a)| \leq 1$. Then, for all states $s$ in the task, $h^+(s) = h^{FF}(s)$.*

**Proof:** Under the given restrictions, relaxed planning comes down to finding paths in the graph where the nodes are the facts, and an edge is between $p$ and $p'$ iff there is an action with $pre(a) = p$ and $add(a) = p'$ (empty preconditions can be modelled by a special fact node that is assumed to be always true). A state has a relaxed plan iff it makes a fact node true from which there is a path to the goal node. Relaxed Graphplan identifies a shortest such path. □

The prerequisites of Propositions 2 and 3 are maximally generous, i.e., when relaxing one of the requirements, one loses the $h^+(s) = h^{FF}(s)$ property. To obtain sub-optimal relaxed plans with Graphplan, i.e., to construct cases where $h^+(s) \neq h^{FF}(s)$, it suffices to have one fact with two achievers, and either two goal facts or one action with two preconditions. The following is such an example. There are the facts $g_1$, $g_2$, $p$, and $p'$. The goal is $\{g_1, g_2\}$, the current state is empty. The actions are shown in Figure 6.

| name | | $(pre,$ | $add,$ | $del)$ |
|---|---|---|---|---|
| **op**$g_1$ | = | $(\{p\},$ | $\{g_1\},$ | $\emptyset)$ |
| **op**$g_2$-$p$ | = | $(\{p\},$ | $\{g_2\},$ | $\emptyset)$ |
| **op**$g_2$-$p'$ | = | $(\{p'\},$ | $\{g_2\},$ | $\emptyset)$ |
| **op**$p$ | = | $(\emptyset,$ | $\{p\},$ | $\emptyset)$ |
| **op**$p'$ | = | $(\emptyset,$ | $\{p'\},$ | $\emptyset)$ |

Figure 6: Actions in an example task where $h^{FF} \neq h^+$.

The optimal relaxed plan here is $\langle$**op**$p$, **op**$g_1$, **op**$g_2$-$p\rangle$. However, Graphplan might choose to achieve $g_2$ with **op**$g_2$-$p'$, ending up with the (parallel) relaxed plan $\langle\{$**op**$p$, **op**$p'\}$, $\{$**op**$g_1$, **op**$g_2$-$p'\}\rangle$. Note that each action has only a single precondition, only a single fact has more than one achiever, and there are only two goals. A similar example can be constructed for the case where there is only one goal but one action with two preconditions.

Obviously, the syntax allowed by either of Propositions 2 or 3 is far too restrictive to be adequate for formulating practical domains.[20] We did not investigate whether there are any more interesting situations where $h^+$ and $h^{FF}$ are the same; our intuition is that this is not the case.

A different question is whether there are provable relations between $h^+$ and $h^{FF}$ in (some of) the 30 benchmark domains considered in the $h^+$ investigation. We did not investigate this question in detail – note that such an investigation would involve constructing detailed arguments about all the individual domains, which is clearly beyond the scope of this paper. None of the domains is captured by either of Propositions 2 or 3. A few results that are easy to obtain are the following. In Simple-TSP, Movie, and Miconic-STRIPS, $h^+$ and $h^{FF}$ are the same. This follows from the extremely simple structure of these domains, where finding step-optimal relaxed plans with Graphplan always results in relaxed plans with an optimal number of actions. However, even in the only slightly more complicated domains Ferry, Gripper, Logistics, Miconic-SIMPLE, and Zenotravel, one can easily construct states where Graphplan's relaxed plans may be unnecessarily long. In Miconic-STRIPS this does not happen because there is only a single vehicle (the lift), with no capacity restrictions (on the number of loaded objects, i.e., passengers). With several vehicles and transportable objects, as can occur in Logistics and Zenotravel (as well as Driverlog, Depots, Mprime, Mystery, and Rovers), the difference between $h^+$ and $h^{FF}$ can become arbitrarily large. Just imagine that $n$ objects must be transported from $l$ to $l'$, and $n$ vehicles are available at $l$. For parallel relaxed planning, it makes no difference if a single vehicle transports all objects, or if one different vehicle is selected *per individual object*. In particular, even with FF's action choice heuristics in relaxed Graphplan, $h^{FF}$ may be $2n+1$ just as well as $3n$.[21] In Ferry and Gripper, where there is only a single vehicle (with capacity restrictions), it may be that there is an upper bound on the difference between $h^+$ and $h^{FF}$; we did not check that in detail.

---

20. We remark that the syntax identified by Proposition 3 is a sub-class of a tractable class of STRIPS planning identified by Bylander (1994). In Bylander's class, a constant number $g$ of goal facts is allowed, where $g$ can be greater than 1; the preconditions may be positive or negative.

In spite of the above, the author's personal experience from developing FF is that, at least in relatively simply structured domains with not many different operators/different ways to achieve facts, the relaxed plans found by relaxed Graphplan are typically pretty close to optimal. There are, presumably, the following two reasons for this. First, the employed action choice heuristics. For example, in the Grid domain, a relaxed plan may choose to pick up a key $k$ with the sole purpose of dropping it again when picking up another key $k'$ with a pickup-and-lose action (c.f. Appendix B.12). This does not happen when selecting actions with minimal precondition weight (the pickup-and-lose action has a higher weight than the pickup action unless one already holds $k$ in the considered state). Second, many of the published benchmark *instance suites* are quite restricted. In Logistics, for example, the situation outlined above, $n$ objects and $n$ vehicles waiting at a location $l$, can not happen for trucks because there is only a single truck in each city. As for airplanes, in the published benchmark instances there usually are only few of these, and so $n$ will be small.

## 5.2 Empirical Relations between $h^+$ and $h^{FF}$

In a large-scale empirical investigation (Hoffmann, 2003b), it turned out that $h^{FF}$ typically preserves the quality of $h^+$. The investigation was aimed at verifying, in those domains where $h^+$ has some "positive" topological property (for example, yielding no local minima), to what extent that property is inherited by $h^{FF}$. We considered 20 benchmark domains, namely the same domains as in the paper at hand, except the 10 IPC-3 and IPC-4 domains.

---

21. One could circumvent this particular phenomenon by, when selecting an action in relaxed Graphplan, employing a minimization of the summed up weight of the preconditions of *all* actions selected so far. It is a topic for future work to explore if this has any effect on FF's performance.

Note that, of the latter 10 domains, only three, namely Dining-Philosophers, Satellite, and Zenotravel, have positive topological properties.

The experimental approach was to take samples from state spaces (a technique adapted from work by Frank et al., 1997). More precisely, the method was the following. Of each domain, a random generator was used to produce a large set of example instances. The instances were grouped together according to the values of the *domain parameters*, i.e., the input parameters to the generator (for example, number of floors and number of passengers in Miconic-SIMPLE). Then, for each single instance, 100 states were sampled, i.e., 100 random sequences of actions were executed in the initial state, where the sequence length was chosen randomly in the interval between 0 and 2 times FF's plan length.[22] Of each resulting state $s$, the exit distance $ed(s)$ was computed by a breadth-first search, and another search determined whether $s$ was located on a valley, i.e., whether there was no path from $s$ to a goal state on which the $h^{FF}$ value decreased monotonically.[23] The maximal exit distance of an instance was approximated as the maximum over the exit distances of the sample states. For every group of instances, the mean number of states on valleys, and the mean maximal exit distance, were computed. The results were visualized by plotting these values over the scaling domain parameters. We give some examples for this directly below, after summarizing the overall results.

The results of the experiment strongly suggested that $h^{FF}$ typically preserves the quality of $h^+$, in the considered benchmark domains. Of the 13 domains in which $h^+$ provably yields no local minima, almost no sample states were located on valleys except in 2 domains, namely Grid and Hanoi. More precisely, in the 11 other domains the experiment considered a total of 230 groups of random instances; in one of these groups, 5.0% of the sample states lay on valleys, in another group it were 2.2%, in another eight groups it were below 1.0%, and in the remaining 220 groups not a single valley state was found. As for the maximal exit distance from benches, of all the tested instances of domains in which there is a bound under $h^+$, only a single sample state had an exit distance larger than that bound, namely an exit distance of 2 instead of 1 in the Logistics domain.[24]

| Blocksworld-no-arm | 0.0 | 0.0 | 0.0 | 0.1 | 0.0 |
|---|---|---|---|---|---|
| Gripper | 0.0 | 0.0 | 0.0 | 0.0 | 0.0 |
| Hanoi | 0.0 | 0.0 | 96.0 | 100.0 | 100.0 |
| Tireworld | 0.0 | 0.0 | 0.0 | 0.0 | 0.0 |

Figure 7: Percentage of sample states on valleys. Mean values for a linear increase of the respective domain parameter.

Figure 7 provides the results regarding sample states on valleys, in those considered domains where there are no local minima (and thus no valleys) under $h^+$, and where in-stances are characterized by a single domain parameter (Movie and Simple-Tsp are left out since there $h^{FF}$ is provably the same as $h^+$). In Blocksworld-no-arm, that parameter is the number of blocks (plus randomization of initial and goal states); in Gripper it is the number of balls to be transported; in Hanoi it is the number of discs; in Tireworld it is the number of flat tires. In each domain, from left to right the table entries correspond to a linear increase in the domain parameter ($2 \ldots 11$ blocks, $1 \ldots 100$ balls, $3 \ldots 10$ discs, and $1 \ldots 5$ tires, respectively). Obviously, the only domain that does not "behave" is Hanoi – where $h^+$ isn't a very useful heuristic anyway, yielding very large benches, c.f. Section 4.

---

22. We tried a few other sampling strategies and found that they did not make much difference in terms of the obtained results.

23. Intuitively, each local minimum lies at the bottom of a valley. We used valleys in the experiment since it may be hard to find a local minimum state by sampling.

24. The author's guess is that the results of a similar empirical investigation in Dining-Philosophers, Satellite, and Zenotravel would be similar, i.e., that the sampled maximal exit distances would hardly increase above the upper bounds proved for $h^+$.

| Blocksworld-no-arm | 0.3 | 1.8 | 2.8 | 3.8 | 3.7 |
|---|---|---|---|---|---|
| Gripper | 1.0 | 1.0 | 1.0 | 1.0 | 1.0 |
| Hanoi | 6.0 | 23.0 | 12.0 | 2.0 | 2.0 |
| Tireworld | 6.0 | 6.0 | 6.0 | 6.0 | 2.0 |

Figure 8: Sampled maximal exit distance. Mean values for a linear increase of the respective domain parameter.

Figure 8 shows the results regarding the sampled maximal exit distance in domains characterized by a single domain parameter. In Gripper and Tireworld, the sampled values respect the bound that is valid for $h^+$ (in the largest Tireworld example, sampling did not find a maximum state in the rather large state space). By comparison, the sampled values in Blocksworld-no-arm, where there is no bound for $h^+$, show a clear increase. Again, the behavior of Hanoi is odd.

Figure 9 shows (part of) the results for a domain that is characterized by more than one domain parameter, namely Logistics. In domains with at least two domain parameters, the experimental method was to run one experiment for each pair of them. In each experiment, all parameters except the respective pair was set to some fixed value. The data could then be visualized in 3-dimensional plots like the ones in Figure 9. In the figure, the parameters scaled are the number of cities and the number of objects ("packages") to be transported; the parameter range is $1 \ldots 9$ in both cases. City size and number of airplanes are both fixed to 3. Of each parameter value combination, 10 random instances were generated (and 100 states were sampled per instance). No valley states were found, except with 3 cities and 9 objects, where 2 of the 1000 sample states were located on a valley. With 5 cities and 3 objects, in a single instance one sample state had exit distance 2, rather than the bound 1 valid for $h^+$ – the single such bound violation found in the entire experiment.[25]

As indicated before, the Grid domain was, with Hanoi, the only domain for that the experiment suggested a major difference between the topologies of $h^+$ and $h^{FF}$. Large fractions of the sample states, up to 62.4%, were located on valleys. There was a clear tendency of increase of the percentage, both with increasing grid size and with increasing number of keys to be transported.

All in all, the experiment confirmed that, in all of the Blocksworld-no-arm, Briefcase-world, Ferry, Fridge, Gripper, Logistics, Miconic-SIMPLE, and Tireworld domains, $h^{FF}$ largely preserves the quality of $h^+$ (no local minima and/or a constant bound on the maximal exit distance from benches). Remember that Miconic-STRIPS, Movie, and Simple-Tsp are three more domains where this, provably, applies.

---

25. The decrease in the mean sampled maximal exit distance for very large parameter values suggests that it becomes harder, for sampling, to find the maximum states in the rather large state spaces.

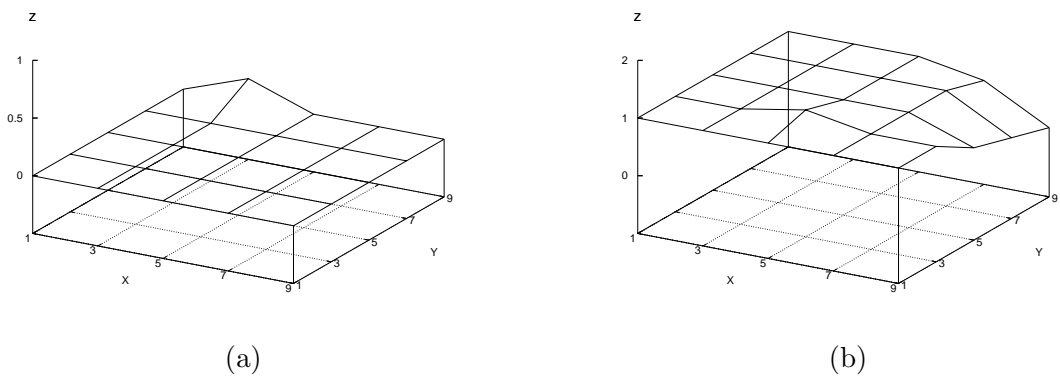

Figure 9: Mean sampled valley percentage (a) and maximal exit distance (b) in Logistics, when scaling cities ($x$-axis) against objects ($y$-axis).

## 6. Towards Automatically Detecting $h^+$ Phenomena

The lemmas presented in Section 3 provide a natural starting point for investigations into domain analysis techniques trying to detect the topological phenomena automatically. Such domain analysis techniques would be useful for configuring hybrid systems, i.e., for the automatic selection of heuristic functions that are likely to be well-suited for solving a given planning task. Further, such techniques would be useful for avoiding the need to re-do the $h^+$ investigation for every single new planning domain. Finally, on the basis of such analysis techniques one may be able to compute good lower bounds on $h^+$, and with that an informative admissible heuristic function. Some more discussion of these points is contained in Section 7.

The question to be addressed is if, to what extent, and how, the application of the lemmas from Section 3 can be automated, i.e., if and how one can automatically check whether their prerequisites are satisfied in a given STRIPS task. In the section at hand, we present a preliminary attempt we made to do that. While the attempt was not very successful, we believe that the investigation has value in showing up what one can achieve with some simple analysis techniques, and what weak points would be needed to be improved upon in order to obtain better results.

Invertible (or at least invertible) actions, and actions with irrelevant delete/static add effects, are syntactically defined in Section 3 and thus easy to "detect". The only difficulty is to find inconsistencies between facts. While this is as hard as planning itself, there are several approximation techniques in the literature (for example, Blum & Furst, 1995, 1997; Fox & Long, 1998; Gerevini & Schubert, 2000, 2001; Rintanen, 2000), which tend to work very well, at least in the current benchmarks. The challenge is to find more syntactical characterizations of actions that are respected by the relaxation, and of actions that have no relaxed-plan relevant delete effects. Now, in many domains where these phenomena occur, such as for example Ferry, Gripper, Logistics, Miconic-STRIPS, Movie, Simple-Tsp, and Tireworld, intuitively when one looks at the domains the causes of the phenomena seem similar. But when getting down to the actual syntax of these domain descriptions, the individual details are very different and it becomes very difficult to get a hold on the common ground. There does not seem to be a simple syntactical definition that captures the behavior of the actions in all these domains; at least we did not find such a syntactical definition. Instead, we tried to reason about the "additive structure" of the domains, and its possible "interactions" with the delete effects. (The intuition being that, in the domains with very simple $h^+$ topology, the interactions aren't very harmful.) We captured the additive structure of a domain/of an instance in a data structure called *fact generation trees*. The next subsection describes this data structure and its basic properties, then a subsection gives our results in an extreme case of $h^+$ topology, then a subsection outlines a somewhat more advanced analysis technique we developed.

## 6.1 Fact Generation Trees

The fact generation tree, short *FGT*, to a planning instance is basically the AND/OR tree that results from a regression search starting at the goals, when ignoring the delete effects of the actions. Tree nodes are labelled with facts and actions alternatingly. Fact nodes are OR nodes – they represent a choice of achieving actions – and action nodes are AND nodes – their preconditions represent sets of facts that must be achieved together. We assume a goal achievement action, as known from, for example, the description of UCPOP (Penberthy & Weld, 1992). That action is the *root* (AND) node of the FGT, and the top level goals form its sons. Obviously, the sons of a fact node are all the actions that achieve the fact, and the sons of each action node are all the precondition facts of the action. (For the sake of simplicity, we stayed in a pure STRIPS framework in this investigation.) Tree structures of this kind were, for example, described and used by Nebel, Dimopoulos, and Koehler (1997) in their work on automatically detecting irrelevant facts and operators. Note that the FGT does not take account of the interactions that may arise when trying to achieve the facts below an AND node together. As an effect of ignoring the delete lists, the FGT treats all these facts completely separately.

We terminate the FGT by applying the following two rules.

1. Say we just inserted an action node $N(a)$ labeled with action $a$. If there is a fact $p \in pre(a)$ so that a fact node labeled with $p$ occurs on the path from the root node to $N(a)$, then $N(a)$ is pruned.

2. Say we just inserted, as a son of an action node $N(a)$, a fact node $N(p)$ labeled with fact $p$. If there is an action $a'$ with $p \in pre(a')$, so that an action node labeled with $a'$ occurs on the path from the root node to $N(a)$, then $N(p)$ is pruned.

Intuitively, the rules disallow the generation of branches in the FGT that would be redundant for a relaxed plan. Formally, we call a relaxed plan non-redundant if no strict subsequence of it is still a relaxed plan (i.e., no action can be omitted). Every non-redundant relaxed plan, for every (not necessarily reachable) state, can be embedded into a connected, rooted, and non-redundant sub-tree of the FGT built in the way described above. We will be more precise after introducing the illustrative example in Figure 10, that we will use throughout this section.

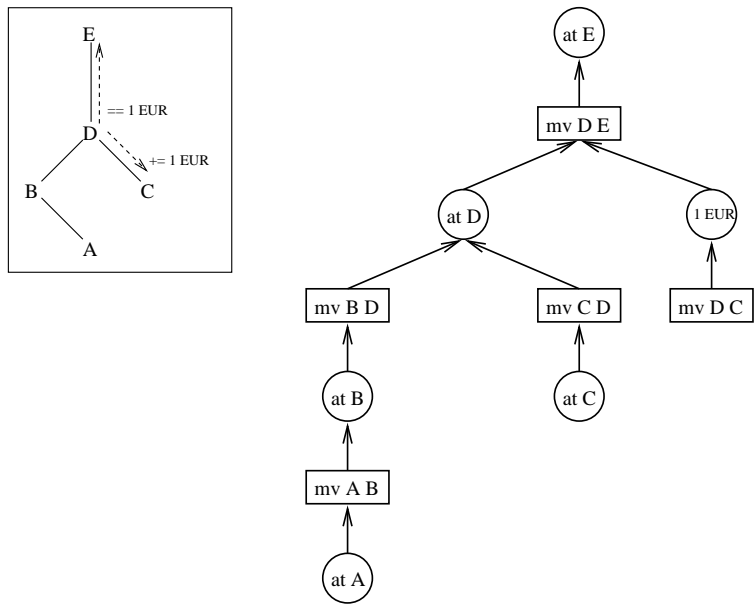

Figure 10: Sketch and FGT of the illustrative example.

In the example, the task is to reach location E. The available actions are moves along (bi-directional) graph edges in the obvious encoding using an "at" predicate, except the move from D to E, which requires as an additional precondition that we be in possession of 1 EUR. We can acquire the 1 EUR as an add effect of the action that moves from D to C. The main part of Figure 10 shows the FGT to the example, the picture in the top left corner illustrates the example by showing its road map graph and an indication of the role of the 1 EUR constructs. The root node, i.e., the artificial goal-achievement action, is not included in the figure. Due to termination rule 1, (for example) moving from E to D is not included as a son of the fact node labeled "at D" (the precondition "at E" is the root node). Due to termination rule 2, "at D" is not a son of the action node labeled "mv D C" ("at D" already occurs as a precondition of "mv D E" above).

Every action in a non-redundant relaxed plan (to some arbitrary state) achieves some unique "needed" fact that is not achieved by any preceding action, and that is needed for the goal or for the precondition of a subsequent action.[26] It is not overly difficult to prove that one can thus embed such a relaxed plan into the FGT by processing the relaxed plan from back to front, associating each action with the corresponding node below a needed fact added by the action, starting at the goal facts. The resulting sub-tree is connected and rooted in the sense that actions are only associated with consecutive AND nodes, starting at the root node. The sub-tree is non-redundant in the sense that, of every OR node, at most one son gets associated with an action. Termination rule 1 is valid since a fact that is needed at the end of a relaxed plan can not also be needed at its start. Termination rule 2 is valid since for every needed fact there is at least one representative node in the corresponding sub-tree. For illustration, consider the different locations in the graph underlying the example in Figure 10. If one is located, for example, at A and does not have the 1 EUR, then the entire FGT except the "mv C D" node corresponds to the sub-tree for a non-redundant relaxed plan. This sub-tree is obtained as follows. The relaxed plan is "mv A B", "mv B D", "mv D C", "mv D E". The needed facts added by these actions are "at B", "at D", "1 EUR", and "at E", respectively. Starting from the goal fact "at E", first "mv D E" gets associated with the respective action node. Then the fact nodes "at D" and "1 EUR" – the preconditions of the action just dealt with – become open, and "mv B D" as well as "mv D C" get associated with the respective node below their respective needed fact. As a consequence of the "mv B D" action, fact node "at B" becomes open, and "mv A B" gets associated with the action node below it. Then the process stops. If, in the current state, one is, for example, located at C with the 1 EUR, then the process selects the sub-tree that consists of the "mv C D" and "mv D E" nodes only.

---

26. This observation was made by, for example, Hoffmann and Nebel (2001b), where it is used to detect actions that do not participate in any non-redundant relaxed plan, and that thus do not need to be considered in the heuristic computations done by planners such as FF or HSP.

Every non-redundant relaxed plan in the instance, in particular every optimal relaxed plan in the instance, corresponds to a sub-tree of the FGT. The FGT being a summary of all possible relaxed plans in that sense, our idea is to examine the FGT for harmful interactions – *conflicts* – with the potential to appear in a relaxed plan. The hope is to be able to draw conclusions from the non-existence/restricted form of conflicts to topological properties of $h^+$. We next outline an extreme case analysis of this kind, namely one that postulates the absence of any conflicts in the FGT. Note here that, in difference to the situation in the illustrative example, in general the FGT can contain action/fact labels in multiple nodes. The worst-case size of the FGT is exponential in the size of the instance description. So, to design practically usable domain analysis techniques, one would need to approximate the FGT, instead of building it completely. This aspect is not treated at all in what follows, where our objective is (only) to find implications between FGT structure and $h^+$ topology in the first place.

## 6.2 Interaction-free Planning Tasks

Think of a conflict as a situation where one part of a (non-redundant) relaxed plan can hinder the execution/success of another part of the relaxed plan. If there are no such conflicts, then every (non-redundant) relaxed plan is executable in reality, implying that $h^+$ is equal to the real goal distance (which of course implies that there are no local minima etc). In the investigated 30 benchmark domains, this is the case (only) in Simple-Tsp, which we use as a motivating example.

We define three kinds of conflicts in the FGT. We call two action nodes, labeled by actions $a$ and $a'$, *allied* if they participate together in a non-redundant sub-tree, i.e., they can occur together in the embedding of a relaxed plan, but are not descendants of each other. (This is the case iff the paths from the root node to $a$ and $a'$ separate in an AND node.) Our first kind of conflicts is given by a pair of allied action nodes labeled $a$ and $a'$, where $a$ deletes a precondition of $a'$. Second kind of conflicts, a pair of action nodes labeled $a$ and $a'$, where $a$ is a descendant of $a'$, and $a$ deletes a precondition of $a'$ that is not added by any action on the path from $a$ to $a'$. Third kind, an action node labeled $a$, where $a$ deletes a goal fact that is not added by any action on the path from $a$ to the respective root node.

If there are no conflicts in the FGT, then we call the task *interaction-free*. It is relatively easy to see that, without conflicts, for every non-redundant relaxed plan (for every non-redundant sub-tree of the FGT) there is an execution order that works in reality. So $h^+$ equals goal distance in interaction-free tasks.

In the illustrative example from Figure 10, the only conflict in the FGT is that between the nodes "mv D C" and "mv D E" – these nodes are allied, and "mv D C" deletes the precondition "at D" of "mv D E". Note that this conflict does indeed capture the reason why $h^+$ is *not* equal to goal distance in the example. In order to be able to move from D to E, one has to first move from D to C and get the 1 EUR. Doing the latter deletes the "at D" precondition of the former. But in the relaxation, after the move from D to C, one is located in both D and C at the same time, and so the relaxed plan needs one step less to achieve the goal (from all states where the move to C has yet to be done).

An example of a domain with interaction-free tasks is the graph-search domain mentioned earlier, where the tasks demand to find a path between two nodes in a directed graph, using the obvious "at"-predicate and "connected"-predicate based encoding. (Our illustrative example above becomes an instance of this domain if one removes the "1 EUR" constructs.) We can even come up with a purely syntactic criterion that captures this example domain.

**Proposition 4** *Let $(A, I, G)$ be a STRIPS planning task so that*

1. *$|G| \leq 1$,*

2. *for all $a \in A$: $|pre(a)| \leq 1$, and*

3. *for all $a \in A$: $del(a) \subseteq pre(a)$.*

*Then $(A, I, G)$ is interaction-free.*

**Proof:** Due to prerequisites 1 and 2, the AND nodes in the FGT all have at most one son. This implies that there are no allied action nodes. Together with prerequisite 3 and our termination rule 1, it implies that no action node can delete the goal fact, or the precondition fact of an ancestor node. □

Instances of the graph-search domain fulfill the prerequisites of Proposition 4 if the static "connected" facts are removed prior to planning. Note that the syntax identified by Proposition 4 is a subset of the syntax identified by Proposition 3, and thus in such tasks $h^{FF}$ is identical to $h^+$, and, since $h^+$ is identical to the real goal distance, plan existence can be decided in polynomial time. Intuitively, this is because the captured syntax can not express *more* than the graph-search domain: plans in a task qualifying for Proposition 4 correspond exactly to paths in the graph where the nodes are the facts, and the edges go from preconditions to add effects. The same is true for relaxed plans.

The instances of the Simple-Tsp domain are not interaction-free. There are conflicts in the FGT between pairs of actions achieving different "visited" goals. For example, say there are three locations to visit, $l_1$, $l_2$, and $l_3$. The action nodes "mv $l_1$ $l_2$" and "mv $l_1$ $l_3$" are allied since they achieve the goals "visited $l_2$" and "visited $l_3$" that both participate in the root AND node. But these actions mutually delete their precondition, "at $l_1$", so they constitute a conflict in the FGT. If they appear together in a relaxed plan, then that relaxed plan is not executable in reality (unless the relaxed plan happens to move back to $l_1$ in between). Observe, however, that after the execution of, for example, "mv $l_1$ $l_2$", one can replace "mv $l_1$ $l_3$" with "mv $l_2$ $l_3$" and so repair the conflict in the relaxed plan. All conflicts in Simple-Tsp FGTs behave this way.

In general, we say that a conflict between allied action nodes $a$ and $a'$ can be *repaired* if there is an action $a''$ such that $pre(a'') \subseteq (pre(a) \cup add(a)) \setminus del(a)$ (thus $a''$ can be executed after $a$), and $add(a'') \supseteq add(a')$ (thus $a''$ achieves what $a'$ should have achieved). Similar repairable cases can be identified for the two other kinds of conflicts. If all conflicts in the FGT can be repaired, then to any non-redundant relaxed plan there is a relaxed plan of the same length that is executable in reality, and so again $h^+$ equals goal distance. This is the case in the Simple-Tsp domain.

We made a preliminary implementation of the above FGT analysis techniques. The implementation correctly detects that in Simple-Tsp instances (as well as in graph-search instances), $h^+$ equals goal distance. In Simple-Tsp, with less than 18 locations the analysis takes only split seconds; with more than 18 locations, the runtime taken explodes fairly quickly.

## 6.3 A More Advanced Analysis

While the above results are encouraging, the technique's applicability – the $h^+$ topology it can detect – is clearly far too severely restricted. It turns out extremely difficult to find less restrictive implications from FGT structure to $h^+$ topology, i.e., sufficient criteria for weaker topological properties. The best we could come up with is a criterion that implies the non-existence of local minima under $h^+$, and that holds true in the Movie domain and in some extremely simple Logistics instances.

The idea behind the criterion is the following. To imply the non-existence of local minima under $h^+$, it suffices to know that, in every state $s$, there is a starting action $a$ of an optimal solution so that $h^+(Result(s, \langle a \rangle)) \leq h^+(s)$. Say we are considering a planning task where all actions are (at least) invertible. Let $s$ be a state and $a$ be the starting action of an optimal solution from $s$. If there is an optimal relaxed plan for $s$ that contains $a$, then we are done with the argument used in Lemma 3. Else, let $P^+$ be an optimal relaxed plan for $s$ that does not contain $a$. $P^+$ can be embedded in a sub-tree of the FGT. If $a$ does not delete any leaf nodes of that sub-tree – any facts that $P^+$ assumes to be true in the state of execution – then $P^+$ is a relaxed plan for $Result(s, \langle a \rangle)$ and we are done, too. The case left open is when $a$ does delete a leaf node of the sub-tree occupied by $P^+$. Observe that this does not matter if we have that there are no (or only repairable) conflicts in the sub-tree. Then, $P^+$ is executable in reality, so $P^+$ is an optimal plan for $s$, so the starting action of $P^+$ falls into the first case above and we are done again. We get the following sufficient criterion:

*There are no local minima under $h^+$ if for all actions $a$ it holds that $a$ is at least invertible, and for all non-redundant sub-trees of the FGT that do not contain $a$, either $a$ does not delete a leaf of the sub-tree, or the sub-tree does not contain any conflicts.*

To test this criterion, all one needs to do is to consider the (redundant) sub-tree of the FGT where the only branches left out are those that start in nodes labeled with $a$. If this sub-tree contains a conflict, and $a$ deletes some fact occurring in the sub-tree, then the criterion does not apply. Otherwise, if the test succeeds for all actions, it is proved that there are no local minima under $h^+$.

Reconsider the illustrative example from Figure 10, where as said above the only conflict in the FGT is that between the nodes "mv D C" and "mv D E". Any sub-tree that does not contain one of these nodes is conflict-free. So "mv D C" and "mv D E" do not violate the above criterion. Neither do "mv B D", "mv C D", and "mv A B" violate the criterion, since none of these actions deletes a fact occurring anywhere else but in its own precondition. However, for "mv B A" and "mv D B" the sub-tree looked at is the entire FGT including the conflict, and both these actions delete a fact that occurs in the FGT. So the criterion does not apply to our illustrative example. Note that "mv B A" and "mv D B" never start an optimal plan so really they could be left out of the considerations; but it is unclear how to detect this automatically, in a general way.

A remark on the side is in order here. If an action $a$ does not appear in the FGT, then, in difference to what one may think at first sight, this does not imply that $a$ does not appear in an optimal plan. Our FGT termination rules, while adequate for relaxed planning, are too restrictive for real planning. The following is an example. There are the facts $g_1$, $g_2$, and $p$. The goal is $\{g_1, g_2\}$, the current state is $\{g_1\}$. The actions are shown in Figure 11.

| name | | (pre, | add, | del) |
|---|---|---|---|---|
| **op**$p$ | $=$ | $(\{g_1\},$ | $\{p\},$ | $\emptyset)$ |
| **op**$g_2$ | $=$ | $(\emptyset,$ | $\{g_2\},$ | $\{\neg g_1\})$ |
| **op**$g_1$ | $=$ | $(\{p\},$ | $\{g_1\},$ | $\emptyset)$ |

Figure 11: Actions in an example task where the FGT does not contain an action (**op**$p$, namely) needed in reality.

The only optimal plan here is $\langle \mathbf{op}p, \mathbf{op}g_2, \mathbf{op}g_1 \rangle$: in order to be able to re-achieve $g_1$ after applying $\mathbf{op}g_2$, we must achieve $p$ first. However, $\mathbf{op}p$ does not appear in the FGT. The only location in the FGT where a node $N$ labeled with $\mathbf{op}p$ could be inserted is as a son of the precondition node $p$ of $\mathbf{op}g_1$, which is inserted as a son of $g_1$. But $N$ is pruned by termination rule 1, because $\mathbf{op}p$ has $g_1$ in its precondition, and $g_1$ appears on the path from the root node to $N$. Note that, indeed, $\mathbf{op}p$ is never part of a *relaxed* plan because achieving $p$ is only good for re-achieving $g_1$ if that is *deleted* by other actions necessary to reach the goals.

Our implementation of the criterion given above easily – within split seconds – proves the non-existence of local minima in Movie instances, regardless of the size of the instance.

The technique does not, however, work in any other domain we tried, except Logistics instances where there is only a single city, with only two locations in it, only a single truck, and only a single package to be transported. Note that this is even simpler than the small illustrative example used in Section 3, where two objects need to be transported.

It is an open question how better results can be achieved, i.e., how more state spaces can be recognized to not feature any local minima under $h^+$. Our feeling is that the backward chaining approach to domain analysis is promising. But, to be successful, the analysis technique should probably invest much more effort into analyzing the way in which the goals can be achieved, and with how many steps, rather than doing just the very crude FGT approximation. With more information available about how goals can be achieved, maybe it would be possible to discover non-trivial cases in which actions are respected by the relaxation.[27] As for detecting actions that have no relaxed-plan relevant delete effects, it is yet completely unclear to us how this could be accomplished.

## 7. Discussion

We have derived a formal background to an understanding of what classes of domains relaxed plan-based heuristic methods, the most wide-spread methods in the modern planning landscape at the time of writing, are well suited for. The formal approach taken is to identify characteristics of the local search topology – the heuristic cost surface – under the idealized heuristic function $h^+$, in a forward searching framework. For 30 commonly used benchmark domains including all competition examples, i.e., for basically all STRIPS and ADL benchmark domains that are used in the field at the time of writing, we proved what the relevant topological properties are. The results coincide well with the runtime behavior of FF. Indeed, empirical results suggest that the quality of $h^+$ is often preserved in FF's approximation of it.

The results are interesting in that they give a rare example of a successful theoretical analysis of the connections between typical-case problem structure, and search performance. From a more practical point of view, the results provide a clear picture of where the strengths and weaknesses of $h^+$ lie, and so form a good basis for embarking on improving the heuristic in the weak cases. Approaches of this kind have already appeared in the literature (Fox & Long, 2001; Gerevini et al., 2003). Most particularly, Fast-Downward's heuristic function (Helmert, 2004) is motivated by observations regarding unrecognized dead ends under $h^+$ in the Mystery domain, and large benches in transportation domains with non-trivial road maps.

Regarding the relevance of our topological results for forward search algorithms other than enforced hill-climbing, note that things like the non-existence of unrecognized dead ends or the non-existence of local minima are certainly useful for any heuristic search algorithm, albeit not in the form of a provable polynomiality result.[28] More generally, the relevance of the topological results for the performance of planners using other search paradigms, or enhanced heuristics, like LPG and Fast-Downward, is a matter needing further investigation. One thing that is certainly clear is that, in the easiest classes of the taxonomy, particularly in domains where in the state space there are *no* local minima under $h^+$, and benches can be escaped in a single step, any planner using an approximation of $h^+$ is likely to work quite well. Indeed that's what one observes in practice. The intuition of the author is that the topology of $h^+$ plays a large role for the efficiency of these planners more generally, i.e., also in other domains. Proving or disproving this is beyond the scope of this paper. In any case, our investigation provides a nice theoretical background with proved results in an idealized setting, and these results can be used as a starting point into investigations tailored to individual systems other than FF.

---

27. We remark that it is not easy to find even trivial syntactical restrictions under which actions are, in general, respected by the relaxation. For example, even when every fact is added by only a single action, one can construct cases of non-respected actions. One such case is the example from Figure 11, where **op**$p$ is not respected by the relaxation.

28. Except in the case where the heuristic function identifies the precise goal distances, which is the case for $h^+$ in 1 of the 30 domains, namely, the Simple-Tsp domain.

Our investigation considers solvable planning tasks only, which is well justified by the focus set in the international planning competitions. Turning the focus on unsolvable tasks, one realizes that much of our techniques and results become useless. In a search space with no solution, the only difference a heuristic function can make lies in the states with infinite heuristic value, i.e., in the states recognized as dead ends. Which means that the only interesting question remaining is for what kinds of dead end states there is no relaxed plan. What do the results herein tell us about this? In those domains where we identified unrecognized dead ends, the results tell us that relaxed plans are a too generous approximation.[29] In the other domains, things look more hopeful. Still, these results are relative to *solvable* instances. Whether or not $h^+$ will detect many of the dead end states in unsolvable tasks will depend on what reasons there can be for such states. The dead ends in unsolvable tasks may be caused by other reasons than those in solvable tasks, since the assumptions making the tasks solvable are not given. Note that many of the benchmarks (for example, Blocksworld and Logistics) do not have any unsolvable instances in their standard definition. To some extent, this makes the existence or non-existence of unrecognized dead ends a choice of the domain designer extending the domain definition. Exploring these issues in detail is a topic for future work.

Talking about future work, the biggest drawback of this research in its current form is, obviously, that it needs to be re-done for every single new planning domain. It would be very desirable, but turns out to be very hard, to come up with more generic – ideally, automatic – methods to determine the topological properties of a domain. We have outlined an attempt we made to develop such automatic methods, based on analyzing properties of fact generation trees. We presented some first promising results, but regarding the applicability to domains of the complexity one would like to be able to handle, our methods are yet far too weak. It is left for future research to answer the question if there are approaches to the topic that work better in practice. As said, our intuition is that there are such better approaches, based on more intelligent backchaining-style reasoning about how the goals can be achieved in a domain. But, at the time of writing, this is pure speculation.

Beside easening the burden of doing all the proofs by hand, the benefits of automatic domain analysis techniques would be twofold. First, an ambitious long-term vision in domain-independent planning is to have an arsenal of complementary heuristics, and combine these into a hybrid system that can automatically be configured to best suit a given arbitrary planning task. The contribution made towards this vision by the results at hand is a very clear picture of where the strengths of $h^+$ lie; to be able to automatically configure a hybrid system, one would need multiple heuristics with different strengths and weaknesses (i.e., heuristics that are of high quality in different classes of domains), as well as the ability to determine automatically what heuristic is likely to work best. (At least such an approach could be more cost-effective, beside being much more insightful, than just trying out all possible combinations of techniques.)

---

29. Unsurprisingly, seeing as deciding plan existence is **NP**-hard in, for example, Mystery, Mprime, Miconic-ADL, and Freecell (Helmert, 2003).

Another benefit from enhanced domain analysis techniques might lie in the ability to generate a high-quality admissible heuristic function for sequential planning. In many domains, optimal relaxed plans mostly consist of actions of which it is easy – for a human – to see that they (or one of a set of similar actions) must be contained in any optimal relaxed plan (for example, all the loading and unloading actions that can't be avoided in a transportation task). So the number of such actions in a state could provide a good lower bound on the value of $h^+$. Note that this phenomenon – actions that must be contained in every relaxed plan – is a stronger version of the notion of actions that are respected by the relaxation. A promising approach seems to be to try to detect the former as a sufficient approximation of the latter.

Since we observed that there are arbitrarily deep local minima under $h^+$ in Blocksworld-arm, but none in Blocksworld-no-arm, one might try to come up with encoding methods trying to model a domain in a way making it best suited for $h^+$. Since Blocksworld-no-arm is basically a version of Blocksworld-arm where all possible pairs of consecutive actions (pickup-stack, unstack-stack, unstack-putdown) were replaced with macro-actions, a good (but somewhat obvious) heuristic for modeling is probably to choose the domain granularity on as high a level of abstraction as possible. More insightful heuristics may be obtained when considering the $h^+$ topology in planning benchmarks enriched with automatically detected macro actions (Botea, Müller, & Schaeffer, 2004, 2005).

Apart from the above, the most important future direction is the adaption of the formal framework, and of the theoretical analysis methods, to the temporal and numeric settings dealt with in modern planning benchmarks and in modern planning systems. The needed adaptations are straightforward for the numeric framework used in Metric-FF (Hoffmann, 2003a). As for temporal planning, if the objective function estimated by the heuristic is the number of actions needed to complete the partial plan, then the adaptation of the framework is probably straightforward as well. If, however, *makespan* is estimated by the heuristic, then most of what is said in this article does not apply. At most, in such a setting our analysis techniques could be relevant if the search uses an estimation of remaining action steps as a secondary heuristic.

## Acknowledgments

I would like to thank Drew McDermott, Fahiem Bacchus, Maria Fox, and Derek Long for their responses to various questions concerning the definitions of/intentions behind the competition domains. I also thank the anonymous reviewers, whose comments helped to improve the paper.

## Appendix A. Proof Sketches

We list the proof sketches in sections concerning dead ends, local minima, and benches, in that order.

### A.1 Dead Ends

**Theorem 1** *The state space of any solvable instance of*

1. *Blocksworld-arm, Blocksworld-no-arm, Briefcaseworld, Depots, Driverlog, Ferry, Fridge, Gripper, Hanoi, or Logistics is undirected,*

2. *Grid, Miconic-SIMPLE, Miconic-STRIPS, Movie, Pipesworld, PSR, Satellite, Simple-Tsp, Tireworld, or Zenotravel is harmless,*

3. *Dining-Philosophers, Optical-Telegraph, Rovers, or Schedule is recognized under evaluation with $h^+$.*

Most of the proofs are simple applications of Lemma 1 or 2. As said, descriptions of the domains can be looked up in Appendix B.

**Proof Sketch:** [Theorem 1]

All actions in Blocksworld-arm, Blocksworld-no-arm, Driverlog, Ferry, Gripper, Hanoi, and Logistics instances are invertible, so we can apply Lemma 1 and are finished. The inverse actions are the obvious ones in all cases, like stacking/unstacking a block onto/from some other block, loading/unloading an object onto/from a vehicle, or moving from $l$ to $l'$/moving from $l'$ to $l$ (in the case of Driverlog, the latter can always be done as the underlying road map is bi-directional, c.f. Appendix B.8). In the Briefcaseworld, Depots, and Fridge domains, while the actions do not strictly obey the definition of being invertible (neither that of being at least invertible), they still invert each other in an obvious way, i.e., for every state $s$ and applicable action $a$ there is an action $\overline{a}$ so that $Result(s, \langle a, \overline{a} \rangle) = s$.

In Movie, actions getting snacks have irrelevant delete effects and static add effects, while rewinding the movie and resetting the counter are at least invertible. A Simple-Tsp action moving from $l$ to $l'$ is at least invertible by moving back. In Tireworld, to all working steps there is an inverse one, except to inflating a wheel. But that has irrelevant delete effects and static add effects. In Miconic-STRIPS, moving a lift is invertible, boarding a passenger is at least invertible, and departing a passenger has irrelevant delete effects and static add effects. In all the four domains, Lemma 2 can thus be applied. In the Miconic-SIMPLE and Grid domains, while the actions do not strictly adhere to the relevant definitions, similar arguments like Lemma 2 prove the non-existence of dead ends. In Miconic-SIMPLE, moving the lift is invertible. Letting passengers in or out of the lift can not be inverted, but those actions need to be applied at most once (similar to static add effects), and they do not interfere with anything else (similar to irrelevant deletes). In Grid, to all actions there is an inverse action, except opening a lock. The latter action excludes only other actions opening the same lock (similar to irrelevant deletes), and each lock needs to be opened at most once, as locks can not be closed (static add effects). In Zenotravel and Satellite, all facts can be re-achieved but sometimes one has to apply several actions to do so. In Zenotravel, after flying an airplane from $l$ to $l'$, to get back to $l'$ one might have to refuel the airplane on top of flying it back. In Satellite, after switching an instrument on, one might have to re-calibrate it, which can always be done but can involve several actions (turning the satellite into the right direction before applying the actual calibration action). In Pipesworld, any push action is inverted by the respective pop action, and vice versa. The state space is not undirected since the pushs/pops for non-unitary pipeline segments are split into two parts. In PSR, there are no dead end states since one can always reach a goal state by waiting, if necessary, then opening all breakers, then bringing the (non-breaker) devices into a goal position, then closing the needed breakers.

In Dining-Philosophers, dead ends arise only when a process (a philosopher) has ini-tiated an impossible reading or writing command (from/to an empty/a full queue) – the queue contents can then not be updated, and no more actions are applicable. (The derived predicate rules that determine if a process is blocked do not apply in this case, since they require that no read/write command has been initiated yet.) Obviously, with no applicable actions there is no relaxed plan either. In all other states, the goal can be reached by traversing individual process state transitions until all philosophers have one fork, and try to take up the other.

In Optical-Telegraph, dead ends arise in two kinds of situations. First, when a process has initiated an impossible reading or writing command, similarly as in Dining-Philosophers, there are no applicable actions and thus no relaxed plan. The second possibility is that the two processes in a pair may take different decisions of where to go next in their commu-nication sequence: one may decide to stop data exchange, while the other may decide to send or receive more data. In such a situation, at least one of the processes is in a state where it has two transitions available, has already activated one of these transitions, and might have already initiated the respective write/read command. The write/read command is impossible (since the other process took a different decision), and no more actions are applicable for that process. The derived predicate "blocking" rules do not apply to the process, because they never apply in process states with more than one available transition. So neither a real nor a relaxed plan exist for the state. From all other reachable states, the goal can be reached by traversing individual process state transitions until all pairs of communicating processes occupy one control channel, and try to write into the other.

In Rovers, there is a plan to a state if and only if, for all soil/rock samples and images that need to be taken, there is a rover that can do the job, and that can communicate the gathered data to a lander. The only chance to run into a dead end is to take a soil/rock sample with a rover that can not reach a lander (the soil/rock sample is available only once). But then, there is no relaxed plan to the state either.

In Schedule, any state $s$ with $gd(s) < \infty$ can be solved by applying, to each object $o$ in turn, a certain sequence of working steps. If the sequence can *not* be applied for some object $o$ then it follows that the preconditions of a needed action are not fulfilled, which must be the case because $o$ is not cold in $s$ (a "do-roll" action has been applied to $o$ previously, making $o$ hot). No operator can make $o$ cold again, i.e., no operator adds the respective fact. Thus there is no relaxed plan for $s$ either. □

Note that the worst cases in Theorem 1 can occur, i.e., in the domains whose instances are harmless, there can be directed state transitions, and in the domains whose instances are recognized, there can be dead ends. We remark that the dead ends in Dining-Philosophers and Optical-Telegraph are due to what seem to be bugs in the encoding of the queues (whose contents aren't always updated correctly) and of the blocked situations (whose rules for detection seem to be incomplete). Modifying the operators in a straightforward way to fix these (apparent) bugs, one gets dead-end free (harmless) state spaces.

The domains not mentioned in Theorem 1 are Airport, Assembly, Freecell, Miconic-ADL, Mprime, and Mystery. In all these domains, one can construct arbitrarily deep unrecognized dead ends. In Airport, unrecognized dead ends arise when two planes move towards each other on a line of segments, with no possibility of changing the direction. Such deadlock situations aren't recognized by relaxed planning since, in the relaxation, the free space left between the two planes remains free, and can be used to navigate the planes "across" each other. The dead end becomes arbitrarily deep when, independently of the deadlock situation, other planes can still be moved. We remark that, in reality – and in the IPC-4 example instances – deadlock situations like this rarely occur. Airplanes are only movable along standard paths that serve to avoid such deadlocks on the main connecting routes of the airport. The only places on the airport where deadlocks can occur, both in reality and in the IPC-4 example instances, are near the parking areas, where space can be dense, and airplanes need to move in both directions on the same airport segment. If no deadlocks can occur at all, i.e., if all planes can move to their target positions one after the other without hindering each other, then $h^+$ delivers the exact goal distance. This is presumably the reason why the heuristic planners performed very well in the IPC-4 Airport test suites. The performance would probably become worse if one were to use (unrealistic) instances with excessively many potential deadlock situations.

In Assembly, unrecognized dead ends can arise when several objects are "stuck" due to complex ordering constraints, which imply that any solution plan would need to go through a cyclic assembly pattern. The details are rather complicated, and the interested reader is referred to the TR (Hoffmann, 2003c). It can be proved that, unless the ordering constraints in an Assembly instance have the potential to yield a cyclic situation, there are no dead ends at all. In all but one of the IPC-1 competition instances, the ordering constraints do not have this potential. This helps to explain how FF can be so efficient in that test suite (it solves even the largest task within half a second search time, finding a plan with 112 steps).

In Freecell, unrecognized dead ends can arise, for example, when one is not cautious enough about moving cards into the free cells. A relaxed plan can still achieve the goal with a single free cell, using that cell as an intermediate store for all cards. In reality, however, moving a card into a free cell occupies space (by deleting the availability of the free cell), and can thus exclude possibilities of reaching the goal. Thus moving a card into a free cell can lead into an unrecognized dead end state. The unrecognized dead end can be arbitrarily deep when other cards can still be moved around independently of the deadlock situation.

In Miconic-ADL, unrecognized dead ends arise when a problem constraint is violated, but this violation goes unrecognized by the relaxed plan. An example is when two passengers $p_1$ and $p_2$ are in the lift, such that $p_1$ can only be transported downwards, $p_2$ has no access to $p_1$'s destination floor, and $p_2$'s destination floor is below $p_1$'s. The state is a dead end because one can not let $p_1$ get out first – $p_2$ has no access to the respective floor – but neither can one let $p_2$ get out first – afterwards, the lift would need to drive upwards, which it can't with $p_1$ on board. In the relaxation, one can stop at both destination floors "simultaneously" because the *at*-facts are not deleted. The unrecognized dead end becomes arbitrarily deep when several other passengers can be moved around before reaching $p_1$'s destination floor.

In Mystery, unrecognized dead ends arise when fuel is scarce, and a vehicle makes suboptimal moves. The relaxed plan can achieve the goal as long as all relevant locations are still accessible at least once. But that may not suffice in reality. The dead end becomes arbitrarily deep when additional objects can be transported independently of the problematic situation. Mprime behaves similarly. The only difference to the Mystery example is that, to avoid the possibility of transferring fuel items to the problematic locations, one must make sure that there is just enough fuel to enable the transportation of the additional objects.

## A.2 Local Minima

**Theorem 2** *Under $h^+$, the maximal local minimum exit distance in the state space of any solvable instance of*

1. *Blocksworld-no-arm, Briefcaseworld, Ferry, Fridge, Grid, Gripper, Hanoi, Logistics, Miconic-SIMPLE, Miconic-STRIPS, Movie, Simple-Tsp, or Tireworld is* 0,

2. *Zenotravel is at most* 2, *Satellite is at most* 4, *Schedule is at most* 5, *Dining-Philosophers is at most* 31.

We present the proof sketch to Theorem 2 in terms of three groups of domains with similar proofs. Note that the domains where the maximal local minimum exit distance is 0 are domains where there are no local minima at all. We first focus on the domains where Lemma 3, or slight extensions of it, can be applied.

**Proof Sketch:** [Theorem 2, Ferry, Fridge, Gripper, Logistics, Miconic-SIMPLE, Miconic-STRIPS, Movie, Simple-Tsp, Tireworld]

With Theorem 1, none of the listed domains contains dead ends. As said in the proof sketch to the theorem, all actions in the Ferry, Gripper, Logistics, Miconic-STRIPS, Movie, Simple-Tsp, and Tireworld domains are either at least invertible, or have irrelevant delete effects. With Lemma 3 it suffices to show that all actions are respected by the relaxation. In all cases, except the driving/flying actions in Logistics, it is very easy to see that any optimal starting action does something that can not be avoided in the relaxed plan. (For example, the relaxed plan can not avoid to load/unload objects onto/from vehicles, and it can not avoid missing working steps in Tireworld.) If an optimal starting action $a$ in Logistics drives a truck/flies an airplane to some location $l$, then some object must either be loaded or unloaded at $l$, so a relaxed plan from $s$ has no choice but to apply some action that moves a transportation vehicle (of $a$'s kind) there. All vehicles are equally good, except when there is a clever choice, i.e., a vehicle that already carries objects to be unloaded at $l$. But then, $a$ will move one of those vehicles just like an optimal relaxed plan will, and all such vehicles are equally good in the relaxation. (In Ferry, Gripper, and Miconic-STRIPS, there is only a single vehicle, which makes the moving actions in these domains easier to reason about.)

In the Fridge and Miconic-SIMPLE domains, the actions do not adhere strictly to the definitions of invertibility and irrelevant delete effects. But the proof to Theorem 1 has shown that they have similar semantics, i.e., they can either be inverted, or delete only facts that are no longer needed once they are applied. Furthermore, all actions in these domains are respected by the relaxation. In Fridge, missing working steps must also be done in the relaxed plan. In Miconic-SIMPLE, lift moves are trivially respected, and lift stops are respected since clever choices in reality coincide with clever choices in the relaxed plan. □

In the next four domains, there are no local minima either, but the proofs are more sophisticated and make use of rather individual properties of the respective domains. In all cases it is proved that there is a path to the goal on which $h^+$ does not increase.

**Proof Sketch:** [Theorem 2, Blocksworld-no-arm, Briefcaseworld, Grid, Hanoi]

With Theorem 1, none of these domains contains dead ends. In Blocksworld-no-arm, if an optimal starting action $a$ stacks a block into its goal position, then $a$ also starts an optimal relaxed plan (because there is no better thing to do than to achieve this goal immediately); in that relaxed plan, $a$ can be replaced with its inverse counterpart to form a relaxed plan for the successor state. If there is no such action $a$ in a state $s$, then one optimal plan starts by putting some block $b$ – that must be moved in order to access a block below it – from some other block $c$ onto the table, yielding the state $s'$. A relaxed plan for $s'$ can be constructed from a relaxed plan $P^+$ for $s$ by, taking account of various case distinctions, replacing the move actions regarding $b$ in $P^+$ with the same number of other such move actions. The case distinctions are about what kind of action $P^+$ uses to move $b$ away from $c$ – one such action $a'$ must be contained in $P^+$. If $a'$ moves $b$ to the table then we can replace $a'$ in $P^+$ with the action that moves $b$ back onto $c$, and are finished. Else, we must distinguish between the cases where $b$ is required to be on $c$ for the goal, or on some other block. In both cases, we can make successful use of the fact that $b$ can be moved from any position to any other position within a single action, enabling us to exchange actions in $P^+$ quite flexibly.

In Briefcaseworld, all actions can be inverted. Actions that put objects into the briefcase are trivially respected by the relaxation. In a state $s$ where an optimal plan starts with a take-out action, an optimal relaxed plan for $s$ can also be used for the successor state, since taking out an object does not delete important facts. In a state $s$ where an optimal plan starts with a move action from $l$ to $l'$, and $P^+$ is a relaxed plan for $s$, a relaxed plan for the successor state can be constructed by replacing moves from $l$ to $l''$, $l'' \neq l'$, in $P^+$, with moves from $l'$ to $l''$.

In Grid, a rather complex procedure can be applied to identify a flat path to a state with better $h^+$ value. In a state $s$, let $P^+$ be an optimal relaxed plan for $s$, and let $a$ be the first unlock action in $P^+$, or a putdown if there is no such unlock action. Identifying a flat path to a state $s'$ where $a$ can be applied suffices, because unlocking deletes only facts that are irrelevant once the lock is open, and the deletes of putting down a key are irrelevant if there are no more locks that must be opened. The selected action $a$ uses some key $k$ at a position $p$. $P^+$ contains a sequence of actions moving to $p$. Moving along the path defined by those actions does not increase $h^+$ since those actions are contained in the relaxed plan, and they can be inverted. If $k$ is already held in $s$, then we can now apply $a$. If the hand is empty in $s$, or some other key is held, then one can use $P^+$ to identify a flat path to a state where one *does* hold the appropriate key $k$. If the hand is empty, then $P^+$ must contain a sequence of actions moving to a location where $k$ can be picked up. If some other key is held, then $P^+$ must contain sequences of actions moving between locations where a series of keys are picked up and put down, where the key series ends with picking up $k$.

In Hanoi, it can be proved that the optimal relaxed solution length for any state is equal to the number of discs that are not yet in their final goal position – proceeding from the smallest to the largest disc, each respective goal can be achieved with a single action. No optimal plan moves a disc away from its final position, so $h^+$ does not increase on optimal solution paths. □

We finally consider those four domains where there are local minima, but one can always escape them within a constant number of steps. In all cases, we prove an upper bound $d$ on the distance of any non-dead end state $s$ to a state $s'$ with $h^+(s') < h^+(s)$. This immediately implies that $d - 1$ is an upper bound on the maximal local minimum exit distance (it also implies that $d - 1$ is an upper bound on the maximal bench exit distance; the results will be re-used in Appendix A.3).

In Dining-Philosophers, $h^+$ is only loosely connected to goal distance, and the bound, which holds even for the trivial heuristic function returning the number of yet un-blocked philosophers, follows from the rather constant and restrictive domain structure. In the other three domains, the proofs proceed as follows. For a reachable state $s$, we identify a constant number of steps that suffices to execute one action $a$ in the optimal relaxed plan for $s$, and, without deleting $a$'s relevant add effects, to re-achieve all relevant facts that were deleted by $a$. Then, a state $s'$ with $h^+(s') < h^+(s)$ is reached.

**Proof Sketch:** [Theorem 2, Dining-Philosophers, Satellite, Schedule, Zenotravel]

By Theorem 1, the dead ends in Dining-Philosophers are all recognized. In any non dead end state $s$, the shortest relaxed plan blocks all processes (philosophers) that are not yet blocked. For each individual process, at most 3 steps are needed – in the relaxation, to block a process it always suffices to activate a state transition, to initiate a read/write command, and to do the queue update. After the update, the queue is both empty and full, and the read/write is impossible in the sense that the "blocking" rules apply. (With this, a process can block itself in the relaxation, and the $h^+$ value is only fairly loosely correlated with the true goal distance.) Thus, to reach a state with lower $h^+$ value, obviously it always suffices to block one more process. We prove our upper bound by determining a constant bound on the number of steps needed to do that. Such a bound exists because, beside the fact that the philosopher processes are constant and can interfere only with their respective two neighbors at the table, the philosophers have a fixed order in which they try to pick up the forks: they always first try to pick up the fork to their right, then the fork to their left. This restricts the possible combinations of internal states of neighbored philosophers.

In more detail, a philosopher is blocked iff he tries to pick up a fork that is not on the table. For a philosopher $p$, we refer by $pL$ to $p$'s neighbor philosopher on the left side. A description of the 5 different states of each philosopher process is in Appendix B.7. Let $s$ be a non dead end state. Let $p$ be a philosopher that is not blocked in $s$ (if no such $p$ exists, then $s$ is a goal state and there is nothing to prove). We can prove the desired upper bound by an exhaustive case distinction over the states of $p$ and $pL$. For each state $i \in \{1, \ldots, 5\}$ of $p$, we consider each state $iL \in \{1, \ldots, 5\}$ of $pL$. If the combination of $i$ and $iL$ is not possible, then we do nothing. Else, we determine a number $k$ of process state transitions that leads to a state where either: $p$ is blocked and $pL$ is still blocked if it was blocked in $s$; or $pL$ is blocked and was not blocked in $s$. In a few cases, to do so we also have to make distinctions over the internal state of $pL$'s left neighbor $pLL$. The worst case, $k = 6$, occurs when $i = 3$, i.e., when $p$ holds both adjacent forks. Then, $pL$ has to be in either state $iL = 1$ or $iL = 4$ (which means, $pL$ can't hold the fork between $pL$ and $p$ since that is held by $p$). If $iL = 4$, then $pL$ is not blocked in $s$; $pL$ can put down its left fork, getting to state 1 where $pL$ is blocked since it waits to pick up its right fork, held by $p$. If $iL = 1$ then we have to distinguish two cases about the state of $pLL$. We have that $i = 3$ ($p$ holds both adjacent forks), and $iL = 1$ ($pL$ waits to pick up the fork between $p$ and $pL$); $pL$ is blocked. Case A, if the state of $pLL$ is 0, 2, or 3, then $pLL$ holds the fork between $pLL$ and $pL$. We go with $p$ from 3 to 4, from 4 to 1, and from 1 to 2, and we go with $pL$ from 1 to 2. Then, both $p$ and $pL$ are blocked since they wait to pick up the fork to their left. Case B, if the state of $pLL$ is 1 or 4, then the fork between $pLL$ and $pL$ is on the table, and $pLL$ is not blocked. We go with $pLL$ from 4 to 1 (if in 4), and from 1 to 2. After that, $pLL$ holds the fork between $pLL$ and $pL$; we are in case A and can apply that sequence, getting us to a state where $pLL$ is possibly blocked, and both $p$ and $pL$ are definitely blocked.

We always need at most 6 process state transitions to block one more philosopher. The process state transitions take 4 planning actions each, and so this makes 24 planning steps. Some more planning steps are needed due to the subtleties of the PDDL encoding. Subtlety A, a process may have already decided to go to a state, but not yet arrived there – i.e., the respective transition is activated and the read/write command is initiated, so that communication channel/queue is occupied but the transition is not yet complete. At most 2 steps are needed to reach the next internal state (update the queue and wrap up the transition). Subtlety B, to be blocked in a state a process must activate its outgoing transition. In the worst case described above, each of $p$, $pL$, and $pLL$ may require the 2 steps induced by subtlety A; both $p$ and $pL$ require the step induced by subtlety B. So all in all we get to (at most) 32 planning actions. In effect of the last action, one more process becomes blocked, so an upper bound on the exit distance is 31.

By Theorem 1, there are no dead ends in Satellite. Let $s$ be a reachable state. To determine an upper bound $d$ on the distance from $s$ to a state $s'$ with $h^+(s') < h^+(s)$, one can look at an optimal relaxed plan $P^+$ for $s$, and distinguish four cases regarding the existence of applicable actions of different types in $P^+$. For each action type, a constant number of steps suffices to re-achieve the deleted facts after application of the action. The worst case, $d = 5$, arises when a switch-on action is applied. Switching on an instrument deletes the instrument's calibration. To re-achieve this, one must turn the satellite and calibrate it. After another turn and taking an image, a state with lower $h^+$ value is reached.

By Theorem 1, the dead ends in Schedule are all recognized. Let $s$ be a non-dead end state. To determine an upper bound $d$ on the distance from $s$ to a state $s'$ with $h^+(s') < h^+(s)$, one can look at an optimal relaxed plan $P^+$ for $s$ and distinguish seven cases regarding the kinds of applicable actions that $P^+$ contains. The worst case, $d = 6$, arises when only a do-roll action is available (and applicable) in $P^+$. One then needs to apply a time-step, a do-lathe action to achieve the desired effects of do-roll, another time step, a do-polish or a do-grind action to re-achieve the previous surface condition, another time step, and a do-immersion-paint action to re-achieve the previous color.

By Theorem 1, there are no dead ends in Zenotravel. In a reachable state $s$, to determine the desired constant $d$, distinguishing two cases does the job. If the relaxed plan $P^+$ for $s$ contains an applicable boarding, departing, or refueling action, then applying that action leads into a state with lower $h^+$ value. Else, $P^+$ starts with a flying action, and a better state can be reached by executing the flight, refueling once, and boarding or departing a person. We get $d = 3$. □

Note that the proved bound for Dining-Philosophers holds even if we take the heuristic function to be the trivial one that returns the number of yet un-blocked philosophers. It is extremely cumbersome to figure out what *exactly* the worst-case exit distance is in Dining-Philosophers under $h^+$ – to do so, one has to consider all combinations of possible states of neighbored processes, and their possible developments over a lot of action steps, in a rather un-intuitive PDDL encoding made by an automated translation machinery. The highest exit distance we could actually construct in Dining-Philosophers was 15. We conjecture that this is a (tight) upper bound.

In Satellite, Schedule, and Zenotravel, the proved upper bounds are tight. In all of Dining-Philosophers, Satellite, Schedule, and Zenotravel, the bounds are valid for *any* non-dead end state $s$. So, beside a bound on the maximum exit distance, these results also provide a bound on the bench exit distance; we will re-use them below in Appendix A.3.

In Blocksworld-arm, Depots, Driverlog, Optical-Telegraph, Pipesworld, PSR, and Rovers, one can construct local minima with arbitrarily large exit distances. In Blocksworld-arm, an example situation is that where $n$ blocks $b_1, \ldots, b_n$ initially form a stack where $b_i$ is on $b_{i+1}$ and $b_n$ is on the table, and where the goal is to build the same stack on top of another block $b_{n+1}$, i.e., the goal is a stack $b_1, \ldots, b_n, b_{n+1}$. Reaching, from the initial state, a state with better $h^+$ value, involves disassembling the entire stack $b_1, \ldots, b_n$. During the disassembling process, $h^+$ increases. The same example can be used in Depots.

In Driverlog, local minima can arise due to the different road maps for trucks and drivers, for example, when it takes one step to drive from a location $l$ to another location $l'$, but $n$ steps to walk. In the relaxed plan, the driver can drive the truck to its goal while himself staying where he is, but in reality, the driver will have to walk all the way back.

As for Optical-Telegraph, this is treated most easily by reconsidering the Dining-Philosophers domain, for which we proved a constant upper bound above. The reason is that Optical-Telegraph is basically a more permissive version of Dining-Philosophers, where the philosophers can choose which fork to pick up first, and, if they hold both forks, which fork they want to put down again first. Consider the configuration depicted in Figure 12. That configuration is not reachable given the automata underlying Dining-Philosophers, but is reachable given the automata underlying Optical-Telegraph.

In Figure 12, Nietzsche holds both adjacent forks, while Kant holds none and tries to get access to the fork to his right. In between Nietzsche and Kant, there are arbitrarily many other philosophers that all hold one fork each, and are trying to access the other.

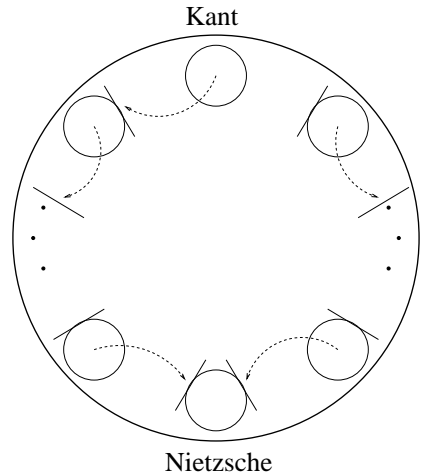

Figure 12: An unreachable situation in Dining-Philosophers, in which an unbounded local minimum under $h^+$ would arise. Arrows indicate "pickup"-requests.

The only non-blocked philosopher is Nietzsche, who can put down the forks again. In the PDDL encoding of this, in the world state where Nietzsche has just activated the transition putting down the right (or left) fork, the $h^+$ value is 2: in the relaxation, it suffices to initiate the write command, and to update the queue contents. After the write command was initiated, however, $h^+$ goes up to 3 because the transition has become non-activated; so the relaxed plan has to update the queue contents, wrap up the transition, then activate the (same) transition again. Reaching a state where the $h^+$ value is 1 involves propagating forks through the entire sequence of philosophers between Nietzsche and Kant, either on the right hand side, or on the left hand side. For example, say Nietzsche puts down both forks and then picks up the right fork. Then the philosopher to the left of Nietzsche can pick up his requested fork (or Nietzsche can pick it up which gets us back to where we started). In the resulting state, we are in the same situation as before, except that now the philosopher with the "Nietzsche-role" sits one more position to the left. After iterating the procedure around the left side of the table, Kant can pick up the requested fork, and request to get the other, giving us a goal state where all philosophers are blocked. The state with $h^+$ value 1 is the one where Kant has not yet activated the transition to request the other fork.

The configuration in Figure 12 is not reachable in the Dining-Philosophers domain as used in IPC-4, because, there, a philosopher can not pick up the fork on his left hand side first – as is done in Figure 12 by all the philosophers between Nietzsche and Kant on Nietzsche's left hand side. As said, in Optical-Telegraph the "philosophers" do have this freedom of choice, and so the situation is reachable. In more detail, as described in Appendix B.22, in Optical-Telegraph there are $n$ pairs of communicating processes. The pairs are arranged in a cycle, where between each pair there is a control channel. Internally, the two processes within each pair can go through a fairly long, heavily interactive, sequence of operations, implementing the possibility to exchange data between the two stations.

Before these operations can begin, each of the processes has to occupy (write into) one control channel. That is, one of the processes occupies a channel, then it waits for a signal from the other process, indicating that the second control channel was occupied as well. After the data exchange was terminated, the control channels get released (read) in an arbitrary order. The overall system is blocked iff all process pairs are in the state where they have occupied one control channel, and are waiting to occupy the other. Thus, the process pairs correspond exactly to philosophers that can choose which fork to pick up (put down) first, and Figure 12 provides an example with arbitrarily high exit distance from a local minimum state. Precisely, the local minimum state is the one where the "Nietzsche" process pair has just occupied both channels, and the process that blocked the second channel has just activated the transition sending the "occupied-the-other-one" signal: in that state, $h^+$ has value 2 (all processes except the active one are blocked).

In Pipesworld, consider the situation where several areas form a circle with unitary connections. In the local minimum state $s$, a single goal batch $g$ has to go into an area $a$; $g$ is currently in a segment $s$ adjacent to $a$, $a$ contains a batch $b$, all other areas are empty. The shortest plan is to push $b$ into the other segment (not $s$) adjacent to $a$, and propagate the batches around in the circle until $g$ can be pushed into $a$. The shortest relaxed plan for $s$ is, however, to push $b$ into $s$ and then push $g$ into $s$ from the other side – i.e., $g$ is used to push itself into the goal area. Reaching the nearest state with $h^+$ value 1 requires $n - 1$ steps when there are $n$ areas in the circle, and on the path the $h^+$ value increases. Note that this example uses neither tankage restrictions, nor interface restrictions, nor non-unitary pipeline segments.

In PSR, a deep local minimum is given when $n$ breakers each feed an individual goal line, in a way so that no breaker can feed any other breaker's goal line without that other breaker being also closed, and the breakers are all connected to some faulty line. All but one of the breakers are closed. The $h^+$ value of such a state is 1 (close the single open breaker) since the only unsatisfied goal condition (beside supplying the line fed by the open breaker) is the one postulating that no breaker is affected; that condition is a negated derived predicate, and thus ignored in the relaxation. The only applicable action in the state is to wait. After that, all breakers are open, and the shortest relaxed plan is to close them all, yielding the $h^+$ value $n$. Obviously, the nearest state with $h^+$ value 0 is at least $n$ steps away.[30]

In Rovers, local minima can arise because taking an image deletes the calibration of the camera. An example is this. There are $n$ waypoints $w_1, \ldots, w_n$ connected in a line (i.e., $w_{i-1}$ is connected to $w_i$), a lander at $w_1$, one rover has a camera $c$ that must be used to take two images at $w_1$, and $c$ can be calibrated (only) at $w_n$. When the rover is at $w_1$, and $c$ is calibrated, the relaxed plan is to take the two images and communicate the two data pieces. But after taking one image, one has to navigate all the way to $w_n$, calibrate $c$, and get back. Note that this example makes use of a road map with arbitrarily large *diameter*, where the diameter of a Rovers instance is the longest way any rover must travel in order to get from one waypoint to another. In general, the distance to a state with better $h^+$ value is bounded by $3d + 2$ where $d$ is the diameter of the instance (see the details in the TR). The road map diameter in the IPC-3 Rovers instances varies around 1 to 6.

---

30. We remark that this counter-example remains valid in the IPC-4 SIMPLE-ADL and STRIPS formulations of PSR, which use a different encoding of the derived predicates, not using a negation to formulate the goal that no breaker is affected.

As for the Airport, Assembly, Freecell, Miconic-ADL, Mprime, and Mystery domains, we have seen in Appendix A.1 that these contain unrecognized dead ends, so, by Proposition 1, the local minimum exit distance in these domains is unbounded. For Assembly, as the TR describes in detail, the *initial state* of an instance has a path to the goal on which $h^+$ decreases monotonically, unless there are complex interactions between the ordering constraints present in the instance. None of the IPC-1 instances features such complex interactions. Assuming that FF's search algorithm sticks to the monotonically decreasing paths, this gives another indication as to how the system can be so efficient in that example suite.

## A.3 Benches

**Theorem 3** *Under $h^+$, the maximal bench exit distance in the state space of any solvable instance of Simple-Tsp is 0, Ferry is at most 1, Gripper is at most 1, Logistics is at most 1, Miconic-SIMPLE is at most 1, Miconic-STRIPS is at most 1, Movie is at most 1, Zenotravel is at most 2, Satellite is at most 4, Schedule is at most 5, Tireworld is at most 6, and Dining-Philosophers is at most 31.*

As before, we subdivide the proof sketch to Theorem 3 into groups of domains with similar proofs. We first consider the transportation-type domains. In all of them, Lemma 4, or very similar proof arguments, can be applied.

**Proof Sketch:** [Theorem 3, Ferry, Gripper, Logistics, Miconic-SIMPLE, Miconic-STRIPS]

The proofs to Theorems 1 and 2 have shown that, in all these domains, the actions are respected by the relaxation, and, in all these domains except in Miconic-SIMPLE, the actions are either invertible, or have no relevant delete effects. To determine an upper bound $d$ on the exit distance from benches, we can thus apply Lemma 4. This requires us to show that, for any state $s$, there is an optimal plan in that the $d + 1$th action has no relaxed-plan relevant delete effects. In Miconic-SIMPLE, we have seen that the actions, while not adhering to the syntactic conditions of invertibility and (no) relevant delete effects, have similar semantics; so the same proof technique can be applied there.

In all the (transportation-type) domains under consideration, the argument is, roughly, that load-type and unload-type actions have no relaxed-plan relevant delete effects, while move-type actions need not be applied more than once in a row because all locations are immediately accessible from each other. This implies an upper bound of 1 on the maximal exit distance. Concretely, say $s$ is a reachable state in a Logistics instance, $a$ starts an optimal plan from $s$, $P^+$ is an optimal relaxed plan for $s$ that starts with $a$, and applying $a$ to $s$ yields the state $s'$. If $a$ is a loading (unloading) action, its only delete is the *at-* (*in-*) fact of the transported object; as the object is loaded from the respective location (unloaded from the respective vehicle) only once in the optimal relaxed plan $P^+$, $a$ has no relaxed-plan relevant delete effects, so $s$ is an exit. Otherwise, if $a$ drives or flies some vehicle $v$ from $l$ to $l'$, then $s'$ is an exit because an optimal plan for $s'$ starts by loading (unloading) some package to (from) $v$. For Miconic-STRIPS and Miconic-SIMPLE, the same arguments apply. In Ferry, the arguments also remain valid except that, if the optimal start action $a$ in the state $s$ boards a car, then this action also deletes the available free space on the ferry. But then, the relaxed plan $P^+$ for $s$ also contains actions that move the ferry to a location $l$, and that debark the car at $l$ (otherwise there would be no point in boarding the car). Placing these actions up front in $P^+$, and removing $a$, yields a relaxed plan for the state that results from applying $a$ in $s$. A similar argument can be applied to prove the claim for Gripper, where gripper hands can hold only one ball at a time. (Note that the argument for Ferry and Gripper uses a somewhat weaker notion than relaxed-plan relevant delete effects, where there *are* such effects, but they are undone by actions contained in the relaxed plan.) $\square$

Next come some non-transportation domains where also Lemma 4 can be applied.

**Proof Sketch:** [Theorem 3, Movie, Simple-Tsp, Tireworld]

The proofs to Theorems 1 and 2 have shown that in these domains all actions are respected by the relaxation, and either at least invertible, or have irrelevant delete effects. We apply Lemma 4 in all cases.

In Movie, all actions have no, and therefore no relaxed-plan relevant, delete effects, with the single exception of rewinding the movie (which deletes the counter being at zero). Obviously, no optimal plan rewinds the movie twice in a row. Thus, $d = 1$ is the desired upper bound.

In Simple-Tsp, $d = 0$ suffices. Say we are in a reachable state $s$ where one is at location $l$. An optimal plan starts with an action $a$ visiting a yet unvisited location $l'$. An optimal relaxed plan for $s$ is to start with $a$, then visit each remaining unvisited location $l''$ by a move from $l'$ to $l''$. The latter actions do not require preconditions deleted by $a$, and so $a$ – every action – has no relaxed-plan relevant delete effects.

In Tireworld, the lowest constant upper bound is $d = 6$. Some *non-final* working steps (like jacking up a hub with a flat wheel on) need to be undone later on, i.e., they have relaxed-plan relevant delete effects. Other *final* working steps (like jacking down the hub) need not be undone, i.e., they have no relaxed-plan relevant delete effects. The longest sequence of non-final working steps that any optimal plan does in a row is the following 6-step one: open the boot (it must be closed again), fetch the wrench and the jack (they must be put away again), loose the nuts on a hub that's got a flat wheel on (the nuts must be tightened again), jack up the respective hub (it must be jacked down again), and undo the nuts (they must be done up again). In the resulting state, one can remove the flat wheel, which needs not be undone. $\square$

For the remaining domains where Theorem 3 claims a constant upper bound on the maximal bench exit distance, we have seen in Appendix A.2 that there are upper bounds on the distance from any reachable state $s$ to a state $s'$ with $h^+(s') < h^+(s)$. These upper bounds trivially also imply upper bounds on the maximal bench exit distance.

**Proof Sketch:** [Theorem 3, Dining-Philosophers, Satellite, Schedule, Zenotravel]

Follows directly from the proof to Theorem 2. $\square$

For all of the above domains, except the last four, one can easily construct examples where the bench exit distance is equal to the proved upper bound. For Satellite, Schedule, and Zenotravel, it is an open question whether there are tighter bounds on the bench exit distance than on the local minimum exit distance; this does not seem particularly relevant, though. (For Dining-Philosophers, as said above it may be that not even the bound on the local minimum exit distance is tight.)

For the Blocksworld-no-arm, Briefcaseworld, Fridge, Grid, and Hanoi domains, Theorem 2 proves that there are no local minima. So there it is important to know whether it can be arbitrarily difficult to escape benches. The answer is "yes" in all cases. In Blocksworld-no-arm, the example is the same one that we already used in Blocksworld-arm and Depots (to show that there are no bounds on the local minimum exit distances). There are $n$ blocks $b_1, \ldots, b_n$ that initially form a stack where $b_i$ is on $b_{i+1}$ and $b_n$ is on the table, and the goal is to build the same stack on top of another block $b_{n+1}$, i.e., the goal is a stack $b_1, \ldots, b_n, b_{n+1}$. The shortest relaxed plan for the initial state is $n$ steps long (remove the stack on top of $b_n$, then move $b_n$ onto $b_{n+1}$). The nearest state with $h^+$ value $n-1$ is the one where $b_n$ has already been stacked onto $b_{n+1}$. That state is $n$ steps away from the initial state.

In Briefcaseworld, the bench exit distance becomes large when many objects must be taken out of the briefcase – in the relaxation, there is no point in taking objects out, since moving the briefcase does not delete any $at$-facts. Consider the state $s$ where $n$ objects $o_1, \ldots, o_n$ are inside the briefcase at a location $l$, and the goal is to have $o_1, \ldots, o_n$ at $l$ and the briefcase at another location $l'$. We have $h^+(s) = 1$: moving the briefcase to $l'$ suffices in the relaxation. But the nearest goal state, $h^+ = 0$, is $n+1$ steps away: one must take all the objects out before moving to $l'$.

In Fridge, if in a single fridge the compressor is held by $n$ screws, then the exit distance of the initial state is $n+1$. To reach a better state, one must: stop the fridge (which must be turned back on in the relaxed plan); unfasten the $n$ screws (which must be fastened again in the relaxed plan); and remove the broken compressor (which needs not be undone as it only deletes the fact that the broken compressor is attached to the fridge).[31]

In Grid, consider the instances where the robot is located on a $n \times 1$ grid (a line) without locked locations, the robot starts at the leftmost location, and shall transport a key from the rightmost location to the left end. The initial value of $h^+$ is $n+2$ (walk over to the key, pick it up, and put it down – the $at$-facts are not deleted), and the value does not get better until the robot has actually picked up the key.

In Hanoi, we have seen that $h^+$ is always equal to the number of discs that are not yet in their goal position. Thus the maximal bench exit distance grows exponentially with the number of discs. From the initial state in an instance with $n$ discs, it takes $2^{n-1}$ steps to move the first (i.e., the largest) disc into its goal position.

For the 9 domains where the local minimum exit distance can be arbitrarily large, it is not as relevant whether the bench exit distance is bounded or not. Escaping a bench might do the planner no better than ending up in a huge local minimum. We remark that, for example, in Driverlog, Rovers, Mprime, and Mystery, one can easily construct examples with large bench exit distances, by defining road maps with large diameters – i.e., by using basically the same example as used above in the Grid domain.

---

31. In fact, one can easily prove that $n+1$ is also an upper bound on the bench exit distance, in Fridge instances where compressors are held by $n$ screws (details are in the TR).

# Appendix B. Domain Descriptions

The following is a list of brief descriptions of the 30 investigated domains. We explain the overall idea behind each domain, the available operators, and what the initial states and goals are. In most cases the set of instances is obvious; restrictions, if any, are explained. We remark that, at some points, the domain semantics seem a bit odd (for example, in Zenotravel, the only difference between flying and zooming a plane is that zooming consumes more fuel). The odd points are, presumably, domain bugs that have been overlooked by the respective domain designers. We have not corrected these bugs as, after all, the investigation is meant to determine the properties of the benchmarks as they are used by the community. The domains are listed in alphabetical order.

## B.1 Airport

In the Airport domain, the planner has to safely navigate the ingoing and outgoing traffic, at a given point in time, across an airport. The main problem constraint is that planes must not endanger each other, which they do if they come too close to each other's running engines. The constraint is modeled by letting each plane "block" the segments that its engines currently endanger. Planes can not enter blocked areas. There are five operators. A plane can be moved from one airport segment to another, if the plane is facing the right direction, and no planes get endangered by the action. Similarly, a plane can be pushed back if that does not cause trouble. One can start up the engines of a plane, let the plane take off, or let the plane settle at a parking position. The initial state specifies the current positions and orientations of the planes, the goal specifies which planes are outbound (have to take off), and which are inbound and to what parking positions.

## B.2 Assembly

In the Assembly domain, a complex object must be constructed by assembling its parts together, obeying certain ordering constraints. The parts themselves might need to be assembled in the same way beforehand. Some parts are "transient", which means that they must be integrated only temporarily. There is a collection of machines, "resources", which might be needed by the working steps. There are four operators. An available resource can be committed to an object, deleting the resource's availability. Releasing a resource from an object is the inverse action. An available object $x$ can be assembled into an object $y$, if $x$ is either a part or a transient part of $y$, if all resources that $y$ requires are committed to $y$, and if all objects that have an assemble order before $x$ are already incorporated into $y$. In effect, $x$ is incorporated into $y$ but no longer available, and $y$ becomes available if all parts of $y$ except $x$ are already incorporated, and no transient part of $y$ is incorporated. An incorporated object $x$ can be removed from $y$, if all resources that $y$ requires are committed to $y$, and, given $x$ is a transient part of $y$ (a part of $y$), if all objects with a remove order (an assemble order) before $x$ are incorporated (not incorporated). In effect, $x$ is available but no longer incorporated, and $y$ becomes available if all parts of $y$ are incorporated, and all transient parts of $y$ except $x$ are not incorporated. In the instances, the part-of relation forms a tree where the goal is to make the root object of the tree available. Also, the assemble and remove order constraints are consistent (cycle-free). These restrictions hold true in the AIPS-1998 competition examples.

### B.3 Briefcaseworld

In Briefcaseworld, a number of portables must be transported, where the transportation is done via conditional effects of the move actions. There are three operators. Putting in a portable at a location can be done if the portable and the briefcase are at the respective location, and the portable is not yet inside. Taking a portable out can be done if it is inside. A move can be applied between two locations, and achieves, beside the *is-at*-fact for the briefcase, the respective *at*-facts for all portables that are inside (i.e., the portables inside are moved along by conditional effects). The goal is to have the briefcase, and a subset of the portables, at their goal locations.

### B.4 Blocksworld-no-arm

Blocksworld-no-arm is a variant of the widely known Blocksworld domain. There are three operators. One can move a block from the table onto another block. One can move a block from another block to the table. One can move a block from another block onto a third block. The initial state of an instance specifies the initial positions of the blocks, the goal state specifies a (consistent, i.e., cycle-free) set of *on* facts.

### B.5 Blocksworld-arm

The instances of Blocksworld-arm are the same as those of Blocksworld-no-arm. The difference is that blocks are moved via a single robot arm that can hold one block at a time. There are four operators. One can pickup a block that is on the table. One can put a block, that the arm is holding, down onto the table. One can unstack a block from some other block. Finally, one can stack a block, that the arm is holding, onto some other block.

### B.6 Depots

The Depots domain is a kind of mixture between Logistics and Blocksworld-arm. There is a set of locations, a set of trucks, a set of pallets, a set of hoists, and a set of crates. The trucks can transport crates between locations, the hoists can be used to stack crates onto other crates, or onto pallets. There are six operators, to move a truck between (different) locations, to load a crate that is held by a hoist onto a truck at a location, to unload a crate with a hoist from a truck at a location, to lift a crate with a hoist from a surface (a pallet or a crate) at a location, and to drop a crate that is held by a hoist onto a surface at a location. A hoist can hold only one crate at a time. The crates are initially arranged in arbitrary stacks, where the bottom crate in each stack is standing on a pallet. The goal is to arrange the crates in some other arbitrary stacks on (possibly) other pallets, which can involve transporting crates to other locations (as pallets can not be moved).

## B.7 Dining-Philosophers

Dining-Philosophers is an encoding of the well-known Dining-Philosophers problem, where the task for the planner is to find the deadlock situation that arises when every philosopher has taken up a single fork. The PDDL domain was created by an automatic translation from the automata-based Promela language. The automata are also referred to as *processes*. In Promela, each philosopher is a finite automaton/process that works as follows. From the start state, state 0, a transition puts the right fork onto the table (this is just an initialization step), getting him to state 1. Then there is a loop of four states. From state 1 to state 2, the philosopher takes up the right fork. From 2 to 3, he takes up the left fork, 3 to 4 he puts down the right fork, in state 4 he puts down the left fork and gets back to state 1. Each such process communicates with each of its neighbors through a communication channel, a queue, that either contains a fork, or is empty (if one of the adjacent philosophers is currently holding that fork).

In the PDDL encoding, each process state transition is broken down into four actions. The first action activates the chosen transition. The second action initiates a write or read command to the needed queue, deleting the activation of the transition and setting flags for queue update. The third action updates, if possible, the queue contents. An update is not possible if a write command shall be done to a full queue (a queue that already contains a fork), or if a read command shall be done to an empty queue. The fourth action wraps the process state transition up, re-setting all flags.

Derived predicates are used to model the conditions under which a process is blocked. The rules require that all outgoing transitions of the current state of the process are blocked. A transition is blocked if *it is activated*, and *would need to perform* an impossible queue write/read operation – in the sense that this impossible write/read operation has not yet been initiated.[32] After applying the planning action *initiating* the impossible write/read command, the blocking rules don't apply anymore and so the resulting state is a dead end in the planning task's state space (but not a blocking situation in the process network, according to the derived predicate rules modeling the blocking).

We remark that, in IPC-4, there was also a version of Dining-Philosophers that modeled process blocking via additional planning operators, not derived predicates. We chose to consider the other, above, domain version since it constitutes the more natural and concise formulation, and since planners at IPC-4 scaled further up in it than in the version without derived predicates.

## B.8 Driverlog

Driverlog is a variation of Logistics, where drivers are needed for the trucks, and where drivers and trucks can move along arbitrary (bi-directional) road maps. The road maps for drivers and trucks can be different. There are operators to load/unload an object onto/from a truck at a location, to board/disembark a driver onto/from a truck at a location, to walk a driver from a location to another one, and to drive a truck with a driver from a location to another one. The preconditions and effects of loading/unloading objects are the obvious ones. A driver can board a truck only if the truck is empty; in effect, the truck is no longer empty (as well as driven by the driver). Disembarking a driver is the inverse action. In order to walk a driver from $l$ to $l'$, there must be a *path* between $l$ and $l'$. In order to drive a truck from $l$ to $l'$, there has to be a *link* from $l$ to $l'$ (and there must be a driver on the truck). Paths and links form arbitrary (in particular, potentially different) graphs over the locations, the only restriction being that they are undirected, i.e., if a truck or driver can move from $l$ to $l'$ then it can also move back. This restriction is imposed on the Driverlog instances as can be generated with the IPC-3 generator.

---

32. Only one outgoing transition can be activated at any time, so a process can never become blocked in a state with more than one outgoing state transition. This appears to be a bug in the translation from Promela to PDDL – the more intuitive requirement would be that only the activated transition needs to be blocked, or that an outgoing transition does not need to be activated in order to be blocked. Note that, in Dining-Philosophers, every automaton state has just one outgoing transition.

## B.9 Ferry

In Ferry, a single ferry is used to transport cars between locations, one at a time. There are three operators. One can sail the ferry between two locations. One can board a car onto the ferry at a location, which deletes an *empty-ferry* fact (plus adding that the car is on the ferry and deleting that the car is at the location). One can debark a car from the ferry at a location, which achieves *empty-ferry* (plus adding that the car is at the location and deleting that the car is on the ferry). The goal is to have a subset of the cars at their goal locations.

## B.10 Freecell

The Freecell domain is a STRIPS formulation of the widely known solitaire card game that comes with Microsoft Windows. A number of cards from different suits are initially arranged in random stacks on a number of columns. The cards must be put home. For each suit of cards, there is a separate home column, on which the cards from that suit must be stacked in increasing order of card value. There is a number of free cells. The cards can be moved around according to certain rules. A card is *clear* if it has no other card on top of it. Any clear card can be put into a free cell (if it's not already there), each free cell holds only one card at a time. Any clear card can be moved onto an empty column. A clear card $c$ can be put home if the last card put home in the same suit was the one preceding $c$. If $c$ and $c'$ are clear cards from differently colored suits, then one can stack $c$ on top of $c'$ if $c'$ is not in a free cell, and $c$'s card value is one less than the card value of $c'$ (so stacks can only be built on columns, in decreasing order of card value, and in alternating colors). The goal is reached when the topmost cards of all suits have been put home.

## B.11 Fridge

In Fridge, one must replace the broken compressor in a fridge. To do this, one must remove the compressor; this involves unfastening the screws that hold the compressor, which in turn involves first switching the fridge off. The goal is to have the new compressor attached to the fridge, all screws fastened, and the fridge switched back on. The origin of this domain is a STRIPS formulation. We consider an adaptation that allows for an arbitrary number of fridges and screws, where each compressor is fastened by the same (arbitrary, at least one) number of screws. The adaptation involves an ADL precondition: a compressor can only be removed if *all* screws are unfastened. There are six operators. One can stop/start a fridge. One can unfasten/fasten a screw from/to a compressor attached to a fridge; to do so, the fridge needs to be turned off, the compressor needs to be attached, and the screw must fit the compressor. Finally, one can remove/attach a compressor from/to a fridge. Removing a compressor requires that the fridge is turned off, and that none of the screws that fit the compressor are fastened. In effect, the compressor is no longer attached to the fridge, and both the fridge and the compressor are free. Attaching a compressor requires that the fridge is turned off, and that the compressor fits the fridge. In effect, the compressor is attached, and the compressor and fridge are no longer free.

## B.12 Grid

In Grid, a robot must move along positions that are arranged in a grid-like reachability relation. The positions can be locked, and there are keys of different shapes to open them. The goal is to have some keys at their goal positions. There are five operators. One can move from position $p$ to position $p'$, which requires (apart from the obvious preconditions) that $p$ and $p'$ are connected, and that $p'$ is open (not locked). One can pick up a key at a position, which requires that the arm is empty (one can only hold one key at a time), and has as effects that one holds the key, that the arm is no longer empty, and that the key is no longer at the position. Putting a key down at a position is the inverse action. One can abbreviate the two previous actions by doing a pickup-and-lose of keys $k$ and $k'$ at a position; to do this, one must hold $k$, which is directly exchanged for $k'$, i.e., the effects are that one holds $k$ and that $k'$ is at the position. Finally, one can unlock a position $p'$ if one is at a position $p$ that is connected to $p'$, and holds a key that has the same shape as the locked position $p'$; the add effect is that $p'$ is open, the delete effect is that the position is no longer locked. The instances specify the initial locations of all keys, of all locked positions, and of the robot, as well as the shapes of the keys and the locked positions. The goal specifies positions for a subset of the keys. The robot always starts at an *open* position. This does make a significant difference: if the robot is allowed to start at a locked position, there can be local minima under $h^+$.[33] Otherwise there are none, c.f. Theorem 2. Intuitively, it makes more sense to let the robot be located in open positions only; the restriction also holds true in the published benchmark examples.

## B.13 Gripper

In Gripper, the task is to transport a number of balls from one location to another. There are three operators. One can move between locations. One can pick up a ball at a location with a hand; apart from the obvious preconditions this requires that the hand is empty; the effects are the obvious ones (the ball is in the hand and no longer in the room) plus that the hand is no longer empty. One can drop a ball at a location from a hand, which inverts the effects of the picking action. There are always exactly two locations, and two gripper hands. Instances thus differ only in terms of the number of balls. These severe restrictions hold true in the AIPS-1998 instances. We remark that adding more locations and/or hands does not affect the topological properties under $h^+$, in fact the proof arguments given in Theorems 1, 2, and 3 remain valid in this case.

---

33. Moving away from a locked initial position can lead to the need of applying several steps to re-open that position. The relaxed plan to the initial state does not realize this, since it ignores the delete on the initial *at*-fact.

## B.14 Hanoi

The Hanoi domain is a STRIPS encoding of the classical *Towers of Hanoi* problem. There are $n$ discs $d_1, \ldots, d_n$, and three pegs $p_1$, $p_2$, and $p_3$. There is a single operator that moves an object $x$ from an object $y$ onto an object $z$ (the operator parameters can be grounded with discs as well as pegs). The preconditions of the move are that $x$ is on $y$, $x$ is clear, $z$ is clear, and $x$ *is smaller than $z$*. The effects are that $x$ is on $z$ and $y$ is clear, while $x$ is no longer on $y$ and $z$ is no longer clear. The semantics of *Towers of Hanoi* are encoded via the *smaller* relation. This relation holds in the obvious way between the discs, and all discs are smaller than the pegs (the pegs are not smaller than anything so they can not be moved). The instances differ in terms of the number $n$ of discs that must be transferred from $p_1$ to $p_3$.

## B.15 Logistics

Logistics is the classical transportation domain, where objects must be transported within cities using trucks, and between different cities using airplanes. There are six operators, to drive a truck between two locations within a city, to fly an airplane between two airports, to load (unload) an object onto (from) a truck at a location, and to load (unload) an object onto (from) an airplane at an airport. The operators all have the obvious preconditions and effects (the most "complicated" operator is that moving a truck, whose precondition requires that both locations are within the city). There is always at least one city, and each city has a non-zero number of locations one of which is an airport. There is an arbitrary number of objects, and of airplanes (which are located at airports). The goal is to have a subset of the objects at their goal locations.

## B.16 Miconic-ADL

Miconic-ADL is an ADL formulation of a complex elevator control problem occurring in a real-world application of planning (Koehler & Schuster, 2000). A number of passengers are waiting at a number of floors to be transported with a lift, obeying a variety of constraints. There is always at least one floor, and an arbitrary number of passengers, each of which is given an origin and a destination floor. There are three operators. The lift can move up from floor $f$ to floor $f'$ if $f'$ is (transitively) above $f$, and vice versa for moving downwards. The lift can also stop at a floor. When it does so at floor $f$, by conditional effects of the stopping action all passengers waiting at $f$ are boarded, and all passengers wanting to get out at $f$ depart. The goal is to serve all passengers, i.e., to bring them to their destination floor. The constraints that must be obeyed are the following.

- In some cases, a passenger $p$ has no access to a floor $f$; the lift can then not stop at $f$ while $p$ is boarded.

- Some passengers are VIPs; as long as these are not all served, the lift can only stop at floors where a VIP is getting on or off.

- Some passengers must be transported non-stop, i.e., if they are boarded, the lift can make no intermediate stops before stopping at their destination floor.

- Some passengers can not travel alone, others can attend them; if one of the former kind is boarded, then so must be at least one of the latter kind.

- There are groups A and B of passengers such that it is not allowed to have people from both groups boarded simultaneously.

- Some passengers can only be transported in the direction of their own travel, i.e., if they need to go up (down), then, while they are boarded, the lift can not move downwards (upwards).

All of these constraints are formulated by means of complex first order preconditions of the operators.

### B.17 Miconic-SIMPLE

The Miconic-SIMPLE domain is the same as Miconic-ADL described above, except that there are no constraints at all.

### B.18 Miconic-STRIPS

The Miconic-STRIPS domain is almost the same as the Miconic-SIMPLE domain, see above. The only difference is that boarding and departing passengers is not done by conditional effects of a stopping operator, but explicitly by separate STRIPS operators. One can board a passenger at a floor. The precondition is that the (current) floor is the passenger's origin, and the only effect is that the passenger is boarded. One can let a passenger depart at a floor. The preconditions are that the (current) floor is the passenger's destination and that the passenger is boarded, and the effects are that the passenger is served but no longer boarded.

### B.19 Movie

In Movie, the task is to prepare for watching a movie. There are seven different operators. One can rewind the tape, which adds that the tape is rewound, and deletes that the counter is at zero. One can reset the counter, the only effect being that the counter is at zero. One can get five different kinds of snacks, the only (add) effect being that one has the respective snack. Instances differ only in terms of the number of items that there are of each sort of snacks. The goal is always to have one snack of each sort, to have the tape rewound, and to have the counter at zero.

### B.20 Mprime

Mprime is a transportation kind of domain, where objects must be transported between locations by means of vehicles, and vehicles use non-replenishable fuel. In an instance, there are a set $L$ of locations, a set $O$ of objects, and a set $V$ of vehicles. There also are sets $F$ and $S$ of fuel numbers and space numbers. Each location initially has a certain fuel

number – the number of fuel items available at the location – and each vehicle has a certain space number – the number of objects the vehicle can carry at a time. There are operators to move vehicles between locations, to load (unload) objects onto (from) vehicles, and to transfer fuel units between locations. A move from location $l$ to location $l'$ can only be made if $l$ and $l'$ are connected (where the connection relation is an arbitrary graph), and if there is *at least one fuel unit available at $l$* ($l$ has a fuel number that has a lower neighbor). In effect of the move, the respective vehicle is located at $l'$, and the amount of fuel at $l$ is decreased by one unit, i.e., $l$ is assigned the next lower fuel number. In a similar fashion, an object can only be loaded onto a vehicle if there is space for that, and in effect the available space decreases. Unloading the object frees the space again. The transfer operator can transfer one fuel unit from location $l$ to location $l'$, if $l$ and $l'$ are connected, and $l$ has at least two fuel units left. As the result of applying the operator, $l$'s fuel number decreases by one, while $l'$'s fuel number increases by one. Note that there is no way to re-gain fuel items (one can transfer them around but one can not obtain new ones). The goal is to transport a subset of the objects to their goal locations.

## B.21 Mystery

Mystery is exactly the same as the Mprime domain described above, except that there is no operator to transfer fuel items.

## B.22 Optical-Telegraph

Like the Dining-Philosophers domain described above in Appendix B.7, Optical-Telegraph is a PDDL compilation of a problem originally formulated in the automata-based Promela language. The mechanics of the PDDL compilation are the same as in Dining-Philosophers, using derived predicates to detect blocked situations. The problem involves $n$ pairs of communicating processes, each pair featuring an "up" and a "down" process. Such a pair can go through a fairly long, heavily interactive, sequence of operations, implementing the possibility to exchange data between the two stations. Before data is exchanged, various initializing steps must be taken, to ensure the processes are working synchronously. Most importantly, each of the process writes a token into a "control channel" (queue) at the beginning of the sequence, and reads the token out again at the end. This causes a deadlock situation because there are only $n$ control channels, each of which is accessed by two processes. More precisely, the process pairs are arranged in a cycle, where between each pair there is a control channel. The overall system is blocked iff all process pairs are in a state where they have occupied (written into) one control channel, and are waiting to occupy the other. In that sense, Optical-Telegraph can be viewed as a version of Dining-Philosophers where the internal states of the philosophers are more complicated. In particular, the "philosophers" (process pairs) here can choose in which order to "pick up the forks" (occupy the control channels). As it turns out, see Appendix A.2, the latter has an important impact on the topology under $h^+$.

We remark that, in IPC-4, there was also a version of Optical-Telegraph that modeled process blocking via additional planning operators, not derived predicates. We chose to consider the other, above, domain version since it constitutes the more natural and concise formulation, and since planners at IPC-4 scaled further up in it than in the version without derived predicates.

## B.23 Pipesworld

In Pipesworld, units of oil derivatives, called "batches", must be propagated through a pipeline network. The network consists of areas connected with pipe segments of different length. The pipes are completely filled with batches at all times, and if one pushes a batch in at the one end of a pipe, the last batch currently in that pipe comes out at the other end. There can be interface restrictions concerning the types of oil derivatives that are allowed to be adjacent to each other inside a pipe, and there can be tankage restrictions concerning the number of batches (of each derivative type) that can be stored at any point in time in the individual areas.

The only available planning operator is to push a batch into a pipe. In the IPC-4 encoding of the domain, which we look at here, for non-unitary pipe segments (pipes containing more than one batch) this operator is split into two parts, a start and a finish action (in order to reduce the number of operator parameters needed to correctly update the pipe contents). Also, pipe segments are encoded in a directed fashion, making it necessary to distinguish between (symmetrical) push and pop actions. The initial state specifies the current batch positions etc., the goal specifies what batches have to be brought to what areas.

## B.24 PSR

In the PSR domain, as used in IPC-4, the task is to re-supply a given set of lines in a faulty electricity network. The nodes of the network are "breakers", which feed electricity into the network, and "devices", which are just switches that can be used to change the network configuration. The edges in the network are the lines, each connecting two or three nodes. The breakers and devices can be open or closed. If they are open, then they disconnect the lines adjacent to them. If breakers are closed, then they feed electricity into the adjacent lines. Some of the lines are faulty. The goal is to ensure that none of the breakers is "affected", i.e., feeds electricity into a faulty line, through the transitive connections in the network. Also, the goal requires that each of the given set of lines is (transitively) fed with electricity from some breaker.

The transitive network semantics, determining if a breaker feeds electricity into some line, and if a breaker is affected, are modeled by means of various derived predicates (with recursive rule antecedents to enable the computation of transitive closure). There are three planning operators. One can open a device or breaker that is currently closed, and one can do the inverse closing action. Both actions require as a precondition that no breaker is currently affected. If the latter is untrue, i.e., if a breaker is currently affected, then the only available action is to wait. Its effect is to open all breakers that are affected.

We remark that, in IPC-4, there was also a different version of PSR, formulated in pure STRIPS without derived predicates. That version constitutes, however, a relatively superficial pre-compiled form of the domain (Hoffmann & Edelkamp, 2005; Edelkamp et al., 2005). It was included in IPC-4 only in order to provide the pure STRIPS planners with a domain formulation they could tackle (the pre-compilation was necessary in order to enable the formulation in pure STRIPS).

## B.25 Rovers

In Rovers, a number of rovers must navigate through a road map of waypoints, take rock and soil samples as well as images, and communicate the data to a number of landers. The nine available operators are the following. One can navigate a rover from one waypoint to another – to do this, the waypoints must be connected for the rover. One can sample soil/rock with a rover at a waypoint using a store – to do so, the rover must have the (empty) store and be equipped for soil/rock analysis, and there must be soil/rock to sample at the waypoint; in effect one has the soil/rock analysis, the store is full, and the soil/rock sample is no longer at the waypoint. One can empty a full store by dropping the store. One can calibrate a camera at a waypoint using an objective, and one can take an image of an objective in a mode with a camera at a waypoint. For both operators, the object must be visible from the waypoint, and the camera must be on board a rover that is equipped for imaging. To calibrate the camera, the object must be a calibration target for it; the only effect of the operator is the calibration of the camera. To take an image, the camera must be calibrated, and support the required mode. The effects are that one has the image data, and that the camera is no longer calibrated. Finally, there are three operators with which a rover can communicate soil/rock/image data to a lander. To do so, the lander's waypoint must be visible from that of the rover; the only effect is that the data is communicated. The instances are restricted in that the visibility and connectivity between waypoints are bi-directional – if waypoint $w$ is visible from waypoint $w'$ then the same holds true vice versa; if a rover can move from $w$ to $w'$ then it can also move back. Another restriction is that no camera is initially calibrated (this serves to make sure that, in a reachable state, each calibrated camera has at least one calibration target). Both restrictions are imposed on the Rovers instances as can be generated with the IPC-3 generator.

## B.26 Satellite

In Satellite, satellites need to take images in different directions, in certain modes, using appropriate instruments. There is a number of satellites, a number of directions, a number of instruments, and a number of modes. There are the following five operators. One can turn a satellite from a direction to another one; the preconditions and effects are the obvious ones, the action can be applied between any pair of directions (no connectivity constraints). One can switch on an instrument on board a satellite, if the satellite has power available; in effect, the instrument has power but is no longer calibrated, and the satellite has no more power available. One can switch off an instrument on board a satellite, if the instrument has power; in effect, the satellite has power available, but the instrument not anymore. One can calibrate an instrument on board a satellite in a direction, if the satellite points into the direction, the instrument has power, and the direction is a calibration target for the instrument. The only effect is the calibration of the camera. Finally, one can take an image with an instrument on board a satellite in a direction and a mode. To do so, the satellite must point into the direction, and the camera must support the mode, have power, and be calibrated; the only effect is that one has an image of the direction in the mode.

The goal is to have images of a number of direction/mode pairs; also, satellites can have a goal requirement to point into a specified direction. The initial states are such that each satellite (but no instrument) has power available, and no instrument is calibrated. The former restriction makes sure that each satellite has the power to run one instrument at a time; the latter restriction makes sure that, in a reachable state, each calibrated instrument has at least one calibration target. Both restrictions are imposed on the Satellite instances as can be generated with the IPC-3 generator.

### B.27 Schedule

In Schedule, a collection of objects must be processed on a number of machines, applying working steps that change an object's shape, surface condition, or color; one can also drill holes of varying widths in varying orientations. There are nine operators. Eight of these describe working steps for an object $o$ on a machine. Amongst other things, these operators' preconditions require that $o$ is not scheduled elsewhere and that the machine is not busy, and these operators' effect is that $o$ is scheduled, and that the machine is busy. The ninth operator does a time step, whose effect is that no object is scheduled, and no machine is busy, any longer. One can apply a do-roll action to an object $o$, which makes $o$ cylindrical and hot (no longer cold, see also below), while deleting any surface conditions, colors, and holes that $o$ might have. One can apply a do-lathe to $o$, making it cylindrical with a rough surface, and deleting any colors that $o$ might have been painted in before. One can apply a do-polish to $o$ if it is cold, giving it a polished surface. One can apply a do-grind to $o$, giving it a smooth surface with no colors. One can apply a do-punch to $o$, with width $w$ in orientation $o$, if $o$ is cold, resulting in $o$ having a hole in $w$ and $o$, and a rough surface. One can also apply a do-drill-press to $o$, if $o$ is cold, making a hole of width $w$ and orientation $o$ into $o$ (changing none of $o$'s properties except making the hole). If $o$ is cold, then one can also apply a do-spray-paint in color $c$, deleting all surface conditions that $o$ might have. Finally, one can apply a do-immersion-paint to $o$, changing none of $o$'s properties except the color. Note that there is no operator that can change $o$'s temperature, except do-roll which makes $o$ hot; after that, $o$ can not be made cold again (this is the reason why dead ends can arise, c.f. Theorem 1). Initially, all objects are cold, and have a shape and a surface condition specified. Some of the objects are also painted initially, and an object can have none or several holes. In the goal condition, some of the objects can be required to have cylindrical shape (the only shape that can be produced by the machines), some need a surface condition, some must be painted, and each object can be required to have an arbitrary number of holes.

### B.28 Simple-Tsp

Simple-Tsp is a trivial version of the TSP problem. There is a single operator to move between locations. This can be applied between any two (different) locations, and the effect (besides the obvious ones) is that the destination location is visited. The instances specify a number of locations that must all be visited, starting in one of them.

## B.29 Tireworld

In Tireworld, one must replace a number of flat tires. This involves a collection of objects that must be used in the appropriate working steps. Briefly summarized, the situation is as follows. There are thirteen operators. There is a boot that can be either opened or closed; it is initially closed and shall be so in the end. There are a pump, a wrench, and a jack which can be fetched or put away (from/into the boot); they are initially in the boot and shall be put back in the end. The spare wheels are initially not inflated, and can be inflated using the pump (the add effect is that the wheel is inflated, the delete effect is that it is no longer not-inflated). Each hub is fastened with nuts; these can be loosened or tightened, using the wrench, while the respective hub is on ground. The jack can be used to either jack up or jack down a hub. Once a hub is jacked up, one can undo the (loose) nuts, or do them up; if the nuts are undone, one can remove the respective wheel, or put on one. An optimal solution plan is this: open the boot; fetch the tools; inflate all spare wheels; loosen all nuts; in turn jack up each hub, undo the nuts, remove the flat wheel, put on the spare wheel, do up the nuts, and jack the hub down again; tighten all nuts; put away the tools; and close the boot.

## B.30 Zenotravel

Zenotravel is a transportation domain variant where the vehicles (called aircrafts) use fuel units that can be replenished using a refueling operator. There are a number of cities, a number of aircrafts, a number of persons, and a number of different possible fuel levels. The fuel levels encode natural numbers by a *next* predicate – $next(f,f')$ is true iff $f'$ is the next higher fuel level than $f$. The task is to transport a subset of the persons from their initial locations to their goal locations. There are the following five operators. One can board/debark a person onto/from an aircraft at a city; this has just the obvious preconditions and effects. One can fly an aircraft from a city to a different city, decreasing the aircraft's fuel level from $f$ to $f'$; $f$ must be the aircraft's current fuel level, and $f'$ must be the next lower level. One can also zoom the aircraft; this is exactly the same as flying it, except that zooming uses more fuel – the aircraft's fuel level is decreased by two units. Finally, one can refuel an aircraft at a city from fuel level $f$ to fuel level $f'$. The conditions are that $f$ is the aircraft's current fuel level, and that $f'$ is the next higher level. Thus aircrafts can be refueled at any city, and in steps of one unit.